\pdfoutput=1

\documentclass{article}

\usepackage[final,main]{neurips_2026}

\usepackage{microtype}
\usepackage{graphicx}
\usepackage{subcaption}
\usepackage{booktabs}
\usepackage{xurl}

\def\huggingface{\raisebox{-1.5pt}{\includegraphics[height=1.05em]{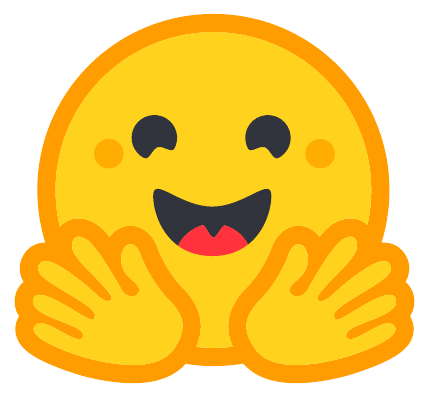}}}

\def\seongland{\raisebox{-1.5pt}{\includegraphics[height=1.05em]{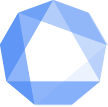}}}
\newcommand{\hflink}{https://huggingface.co/spaces/seonglae/ControlRL}
\newcommand{\articlelink}{https://seongland.com/article/crl}

\usepackage{fontawesome5}

\usepackage{amsmath}
\usepackage{amssymb}
\usepackage{mathtools}
\usepackage{amsthm}
\usepackage{bm}

\usepackage[utf8]{inputenc}
\usepackage[T1]{fontenc}
\usepackage{latexsym}
\usepackage{inconsolata}
\usepackage{float}
\usepackage{wrapfig}
\usepackage{amsfonts}
\usepackage{algorithm}
\usepackage{algorithmic}
\usepackage{nccmath}
\usepackage{nicefrac}

\usepackage{enumitem}

\usepackage{url}

\usepackage{xcolor}
\usepackage{multirow}
\usepackage{colortbl}

\usepackage{capt-of}

\usepackage{amsmath,amsfonts,bm}

\def\eqref#1{equation~\ref{#1}}

\def\1{\bm{1}}

\DeclareMathAlphabet{\mathsfit}{\encodingdefault}{\sfdefault}{m}{sl}
\SetMathAlphabet{\mathsfit}{bold}{\encodingdefault}{\sfdefault}{bx}{n}

\definecolor{bestrow}{rgb}{0.9, 1.0, 0.9}
\definecolor{secondrow}{rgb}{0.95, 0.95, 1.0}

\usepackage{xcolor}
\usepackage{hyperref}

\hypersetup{
    breaklinks=true,
    bookmarksopen=true,
    bookmarksnumbered=true,
    pdfstartview=FitH,
    pdftitle={Control Reinforcement Learning: Token-Level Mechanistic Analysis via Learned SAE Feature Steering},
    pdfauthor={Seonglae Cho, Zekun Wu, Adriano Koshiyama},
    pdfsubject={Machine Learning, Interpretability, Sparse Autoencoders},
    pdfkeywords={SAE, Steering, LLM, Reinforcement Learning, Feature Selection}
}

\definecolor{linkblue}{RGB}{0,0,0}
\definecolor{citegray}{RGB}{0,0,0}
\definecolor{urlblue}{RGB}{0,0,0}

\definecolor{mydarkblue}{rgb}{0,0.08,0.45}
\hypersetup{colorlinks=true, linkcolor=mydarkblue, citecolor=mydarkblue, filecolor=mydarkblue, urlcolor=mydarkblue}
\colorlet{linkblue}{mydarkblue}

\usepackage[capitalize,noabbrev]{cleveref}

\title{Control Reinforcement Learning: \\
Token-Level Mechanistic Analysis via Learned SAE Feature Steering}

\author{%
  Seonglae Cho \qquad Zekun Wu \qquad Adriano Koshiyama \\
  Holistic AI \qquad University College London \\
  \texttt{seonglae.cho@holisticai.com} \\
}

\begin{document}

\maketitle

\vspace{-0.32in}
\begin{center}
\begin{tabular}{rl}
\huggingface & \url{\hflink}\\
\seongland & \url{\articlelink}\\
\end{tabular}
\end{center}

\begin{abstract}
Sparse autoencoders (SAEs) decompose language model activations into interpretable features, primitives for mechanistic circuit analysis.
Yet existing methods identify task-specific circuits via attribution or apply steering along fixed directions, but none learns a policy from task reward for per-token feature amplification.
We cast token-level feature intervention as a single-layer, partially observed control problem over the layer-causal computation of the LLM and approximate it as an MDP.
Control Reinforcement Learning (CRL) trains a policy in this POMDP that selects an SAE feature to amplify at each generation step, yielding per-token intervention traces that expose which features drive task behavior.
Adaptive Feature Masking encourages diverse feature discovery while preserving single-feature attribution.
The framework provides branch point tracking (tokens where feature choice changes outcomes), critic trajectory analysis (separating policy from value estimation errors), and layer-wise comparison along the hierarchy of the residual stream.
Cross-task transfer is asymmetric (MMLU features harm GSM8K but not vice versa; HarmBench helps XSTest), consistent with partial task computation overlap rather than dataset-specific shortcuts: task-conditional attribution does not require transfer invariance.
On Gemma-2 2B and LLaMA-3.1 8B across MMLU, BBQ, GSM8K, HarmBench, and XSTest, CRL serves as an SAE quality diagnostic alongside accuracy gains.
Learned feature steering thus serves as a layer-specific diagnostic complementing SAE analysis based on activations.

\end{abstract}

\section{Introduction}

Circuits, internal computation subgraphs that produce specific model behaviors, have become a primary unit of analysis in mechanistic interpretability~\citep{olah2020zoom, elhage2021mathematical, wang2023interpretability}, with sparse autoencoders (SAEs) providing the feature-level decomposition~\citep{bricken2023monosemanticity, huben2024sparse}.
Existing methods identify task-specific feature subgraphs via attribution~\citep{lindsey2025biology, marks2025sparse, conmy2023towards} or apply steering along fixed directions~\citep{durmus2024steering, rimsky-etal-2024-steering}, but none learns a token- and state-conditional intervention policy from task reward, the adaptation required by the layer-causal autoregressive computation that produces each output (where the same feature may matter at one position and remain inert at another).

\begin{figure}[t]
  \centering
  \includegraphics[width=\textwidth]{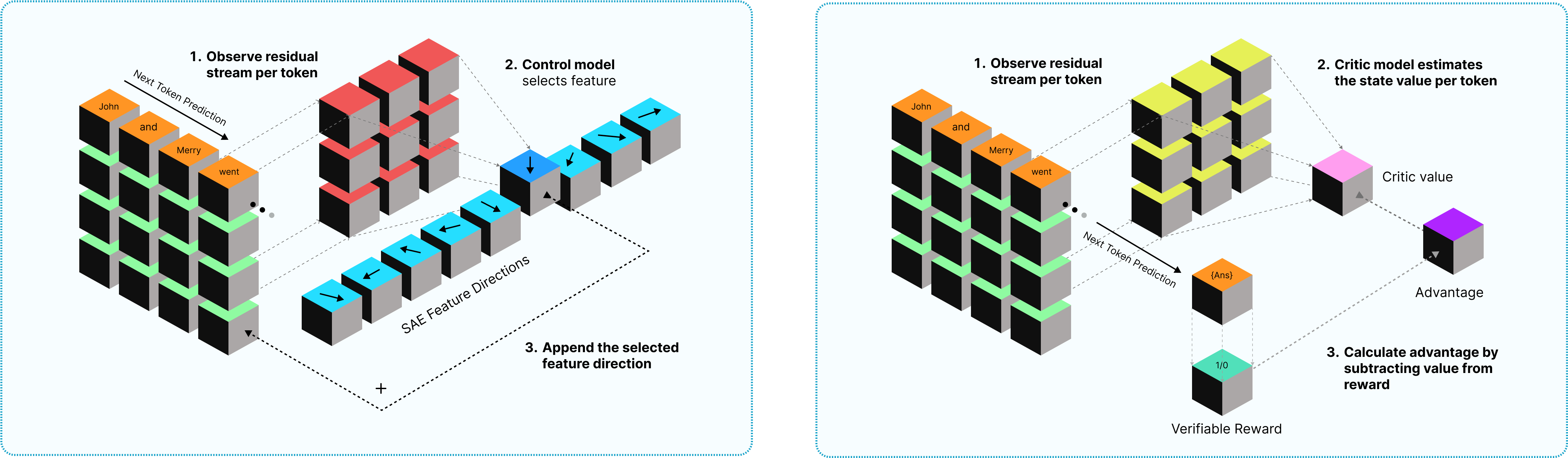}
  \vspace{-1em}
  \caption{CRL overview. The policy observes residual stream activations and selects an SAE feature to amplify at each token. Each intervention produces an interpretable log: which feature was selected and how it affected output at that position. The critic estimates state values for PPO optimization.}
  \vspace{-1em}
  \label{fig:crl_flowchart}
\end{figure}

We introduce \textbf{Control Reinforcement Learning} (CRL), a framework that frames token-level SAE intervention as control over the layer-causal graph of the LLM (single-layer residual stream as the partial observation, feature interventions as actions, forward pass as transition) and, approximating the partially observed problem as an MDP, trains a PPO policy that selects one SAE feature to amplify at each generation step.
Training with task-specific rewards yields per-token intervention traces showing which features, at which layers, drive behavior when amplified, with the training layer selecting the granularity of interpretability.
Adaptive Feature Masking (AFM) prevents policy collapse by dynamically restricting choices based on recent selections, preserving single-feature interpretability across the trace.

CRL provides four capabilities for analysis: \textbf{token-level intervention logging}, \textbf{branch point tracking} (identical prefixes yielding different outcomes localize where feature choice is causally responsible; \hyperref[subsec:branch_analysis]{\textcolor{linkblue}{Section~\ref*{subsec:branch_analysis}}}), \textbf{critic trajectory analysis} separating bottlenecks in policy selection from those in value estimation (\hyperref[subsec:critic_behavior_analysis]{\textcolor{linkblue}{Section~\ref*{subsec:critic_behavior_analysis}}}), and \textbf{layer-wise feature semantics} consistent with the hierarchy of the residual stream~\citep{tenney-etal-2019-bert}.

We evaluate CRL on Gemma 2 2B~\citep{gemmateam2024gemma2improvingopen} with Gemma Scope SAEs~\citep{lieberum-etal-2024-gemma} across MMLU, BBQ, GSM8K, HarmBench, and XSTest, with secondary validation on LLaMA-3.1 8B; cross-task transfer is asymmetric (MMLU features harm GSM8K but GSM8K features are inert on MMLU), consistent with partial circuit overlap rather than dataset-specific shortcuts.

\textbf{Contributions:}
\begin{enumerate}[leftmargin=*, itemsep=1pt, topsep=1pt, parsep=0pt]
\item A POMDP formulation of token-level SAE feature intervention, instantiated as CRL with single-feature/single-layer attribution.
\item Branch tracking, critic trajectory analysis, and AFM as diagnostic primitives for SAE-based mechanistic interpretability.
\item Empirical evidence that the diagnostics flag SAE coverage gaps and direct concrete next steps: richer critics for critic-bottleneck tasks, broader selection diversity for policy-bottleneck tasks.
\item Asymmetric cross-task transfer consistent with partial circuit overlap, not dataset-specific shortcuts.
\end{enumerate}

\section{Related Work}

\textbf{Sparse Autoencoders} address the superposition hypothesis~\citep{elhage2022superposition} by learning to decompose neural activations into interpretable, sparse features~\citep{huben2024sparse, bricken2023monosemanticity}.
Given an activation vector $\mathbf{x} \in \mathbb{R}^{d}$, an SAE learns an encoder $f_{\text{enc}}: \mathbb{R}^{d} \rightarrow \mathbb{R}^{d_{dict}}$ and decoder $f_{\text{dec}}: \mathbb{R}^{d_{dict}} \rightarrow \mathbb{R}^{d}$ where $d_{dict} \gg d$, such that:
\begin{align}
\mathbf{z} &= f_{\text{enc}}(\mathbf{x}) = \text{Activation}(\mathbf{W}_{\text{enc}} \mathbf{x} + \mathbf{b}_{\text{enc}}) \\
\hat{\mathbf{x}} &= f_{\text{dec}}(\mathbf{z}) = \mathbf{W}_{\text{dec}} \mathbf{z} + \mathbf{b}_{\text{dec}}
\end{align}
The training objective uses reconstruction loss and sparsity regularization: $\mathcal{L} = \|\mathbf{x} - \hat{\mathbf{x}}\|^2 + \lambda \|\mathbf{z}\|_1$.

\textbf{Activation Engineering} techniques work by making targeted perturbations to a model's activations \citep{DBLP:journals/corr/abs-2308-10248, hernandez2024inspecting, rimsky-etal-2024-steering}.
In Activation Engineering, steering vectors \citep{subramani-etal-2022-extracting, konen-etal-2024-style} enable control over the language model's behavior, offering more direct behavioral control compared to prompt engineering.
In this context, SAE features provide interpretability and can serve as steering vectors in clamp or addition operations \citep{chalnev2024improvingsteeringvectorstargeting}.
Selecting a suitable coefficient keeps the language model within its optimal "sweet spot" without disruption \citep{durmus2024steering}.
Typically, quantile-based adjustments or handcrafted coefficients are common methods for regulating a feature's coefficient \citep{choi2024automatic}.

\textbf{SAE-based Control} methods demonstrate that SAE features can serve as steering mechanisms for language model control.
However, current approaches face significant limitations: they require either contrastive datasets for feature selection or extensive activation storage for coefficient optimization.
Existing methods also have no feedback signal that adjusts steering during generation, limiting their effectiveness across diverse tasks.

\textbf{Reinforcement Learning for LLM} has been explored in various contexts~\citep{yu-etal-2017-learning, deepseekai2025deepseekr1}, primarily focusing on high-level policy optimization.
AutoSteer~\citep{ferrao2024autosteer} applies RL for interpretable model steering with a fixed feature set; CRL differs by selecting one of $\sim$16K SAE features per token through a learned token-conditional policy, producing per-token intervention traces rather than a global steering vector.
LLMs operate with dual sequences (token positions and layer-wise residual stream propagation), both of which admit Markov approximations, so per-token SAE feature selection can be cast as an RL problem with a task reward signal.

AxBench~\citep{wu2025axbench} shows simple baselines (difference-in-means, ReFT) often outperform SAE-based steering on raw accuracy; Arad et al.~\citep{arad2025saesgoodsteering} narrow that gap via output-score feature filtering without RL.
Concurrent SAE-steering work, including correlation-based selection~\citep{cho2025corrsteer}, per-prompt RL feature-modulation~\citep{ferrao2025fsrl}, and RL-trained reasoning steering vectors~\citep{sinii2025small}, can match or exceed CRL on raw steering accuracy at lower compute cost, but produces static or per-prompt outputs rather than token-conditional traces. CRL pays per-task PPO training to deliver primitives these methods do not produce: per-token (feature, layer, critic value) tuples that support branch-point localization, policy-vs-value bottleneck separation, and depth-selectable attribution; SAE quality bounds effectiveness (\hyperref[subsec:feature_diversity_and_impact_score]{\textcolor{linkblue}{Appendix~\ref*{subsec:feature_diversity_and_impact_score}}}).

\section{Method}
\label{sec:method}

\begin{figure}[t]
\centering
\includegraphics[width=\textwidth]{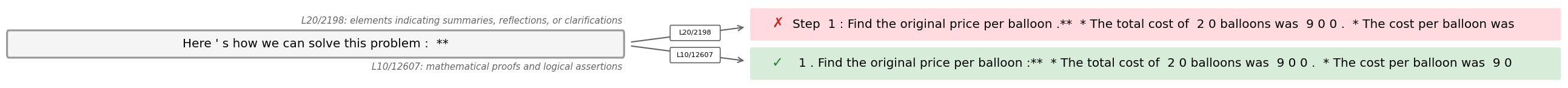}
\includegraphics[width=\textwidth]{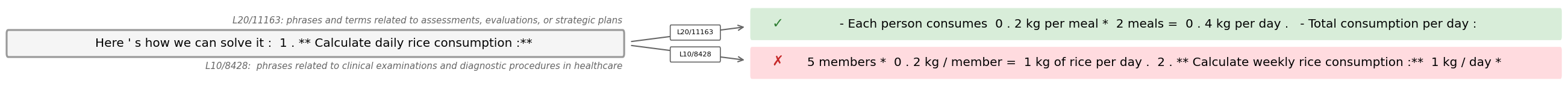}
\includegraphics[width=\textwidth]{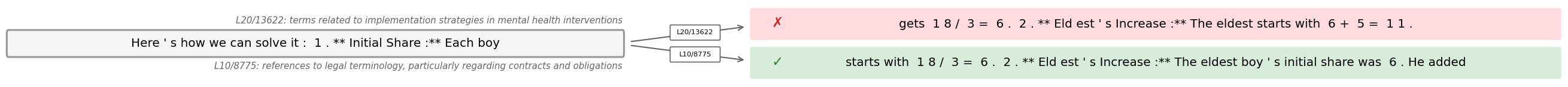}
\includegraphics[width=\textwidth]{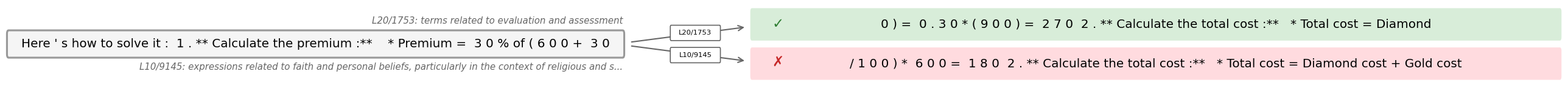}
\includegraphics[width=\textwidth]{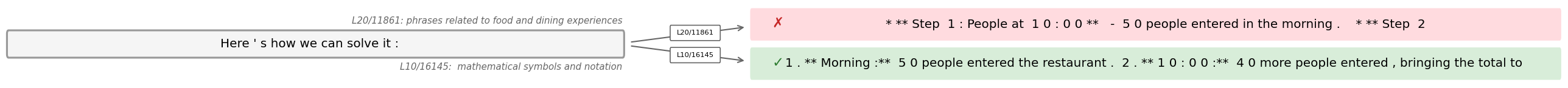}
\includegraphics[width=\textwidth]{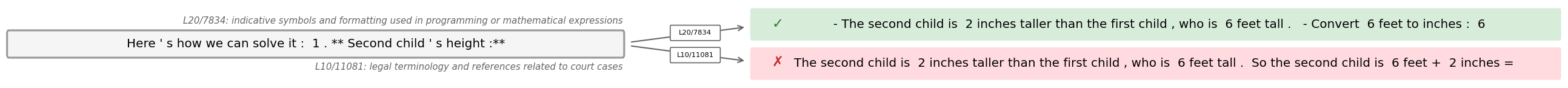}
\caption{Branch point analysis comparing layer 10 vs.\ layer 20 feature semantics. Each panel shows context and competing features from both layers, with correct features highlighted.}
\label{fig:branch_intro}
\end{figure}

\subsection{Problem Formulation: CRL as an MDP over SAE Features}
\label{subsec:problem_formulation}

\hyperref[fig:crl_flowchart]{\textcolor{linkblue}{Figure~\ref*{fig:crl_flowchart}}} illustrates the CRL framework.
We approximate the control of transformer representations as a Markov Decision Process (MDP) in which SAE features are manipulated to optimize task-specific rewards.
The underlying problem is naturally a Partially Observable MDP (POMDP)~\citep{zhong2023a, lim2023optimality}; we adopt the MDP formulation, treating the current residual stream activation as the state~\citep{yu-etal-2017-learning, gui2019long, xu2024dynamic}.

Let $\mathbf{x} \in \mathbb{R}^{d}$ denote the residual stream activations at layer $\ell$ for a target token position, where $d$ is the hidden dimension of the transformer model.
Given a pre-trained SAE with encoder $\mathbf{W}_{enc} \in \mathbb{R}^{d \times d_{dict}}$ and decoder $\mathbf{W}_{dec} \in \mathbb{R}^{d_{dict} \times d}$, the sparse feature activations are computed as $\mathbf{z}_t = \text{Act}(\mathbf{x}_t \mathbf{W}_{enc} + \mathbf{b}_{enc})$, where $\text{Act}(\cdot)$ denotes a generic pointwise activation used in the SAE encoder and $\mathbf{z} \in \mathbb{R}^{d_{dict}}$ represents the sparse feature activations with dictionary size $d_{dict}$.

The MDP is defined by the tuple $(\mathcal{S}, \mathcal{A}, \mathcal{P}, \mathcal{R})$ where:
\begin{itemize}[leftmargin=*,
                itemsep=2pt,
                topsep=0pt,
                parsep=0pt,
                partopsep=0pt]
\item \textbf{State Space} $\mathcal{S}$: The observation is $\mathbf{s} = \mathbf{x} \in \mathbb{R}^{d}$, the residual stream activation at layer $\ell$ for the current token position.
\item \textbf{Action Space} $\mathcal{A}$: Actions are binary vectors $\mathbf{a} \in \{0,1\}^{d_{dict}}$ where selected features (via argmax or top-k) are set to 1, reducing the exploration challenge in high-dimensional feature spaces.
\item \textbf{Transition Function} $\mathcal{P}$: Deterministic transition governed by the transformer's forward pass with steering applied.
\item \textbf{Reward Function} $\mathcal{R}$: Task-specific rewards $r$ based on final output quality evaluation.
\end{itemize}
Although the full problem is a POMDP, we empirically find that treating the current residual stream as a sufficient statistic is effective, making the MDP formulation a practical approximation.

\subsection{CRL-Token: Token-Level Interpretable Steering via SAE Features}
At each generation step $t$, CRL-Token performs token-level interventions by observing the current token's residual stream activation at layer $\ell$ and applying perturbations:
\begin{equation}
\tilde{\mathbf{x}}_t = \mathbf{x}_t + c \cdot \mathbf{a}_t \mathbf{W}_{dec},
\label{eq:steering}
\end{equation}
where $\mathbf{a}_t \in \{0,1\}^{d_{dict}}$ is a binary selection vector, $c$ is the steering coefficient, and $\tilde{\mathbf{x}}_t$ is the steered activation.
By the Markov property, the policy depends only on the current residual stream activation:
\[
\pi_\theta(\mathbf{a}_t \mid \mathbf{x}_{1:t}) = \pi_\theta(\mathbf{a}_t \mid \mathbf{x}_t),
\]
where $t$ is the generation step, starting from the first generated token after the input context.

\subsubsection{Policy Network for Feature Selection}
The policy network $\pi_\theta: \mathbb{R}^{d} \rightarrow \mathbb{R}^{d_{dict}}$ maps residual stream observations to SAE feature selection logits.
We implement this as a 2-layer MLP with Tanh activation (architecture details in \hyperref[subsec:terminology]{\textcolor{linkblue}{Appendix~\ref*{subsec:terminology}}}, training procedure in \hyperref[alg:crl_training]{\textcolor{linkblue}{Algorithm~\ref*{alg:crl_training}}}):

\begin{align}
\boldsymbol{\mu} &= \pi_\theta(\mathbf{s}) \label{eq:policy_logits} \\
\mathbf{a} &= \text{TopK}(\boldsymbol{\mu}, k) \label{eq:feature_selection}
\end{align}

The policy first computes feature selection logits, then selects the top-k features. In this study, we set $k=1$ to focus on the most relevant feature per token, effectively equivalent to ArgMax selection.
We set $k=1$ so each intervention attributes to a single feature, trading multi-feature interaction for clean per-token attribution.
We employ a softmax parameterization for the policy, converting logits $\boldsymbol{\mu} = \pi_\theta(\mathbf{s})$ into probabilities $p_j = \frac{\exp(\mu_j)}{\sum_{i=1}^{d_{dict}} \exp(\mu_i)}$.
During training, actions are \textbf{sampled} from this categorical distribution to ensure unbiased policy gradient estimation, while evaluation uses deterministic argmax selection for reproducibility.
This parameterization ensures differentiability and enables PPO~\citep{schulman2017proximal} to optimize the expected advantage using the log-probability of the selected feature: $\log \pi_\theta(\mathbf{a}|\mathbf{s}) = \log p_{j^\star}$.
By decoupling probability learning from coefficient determination, the model can learn which features to select without prematurely collapsing to a single feature.

\textbf{Source of partial observability.} The partiality comes primarily from observing a single layer $\ell$: the controllable state is the full multi-layer forward pass, but the policy sees only $\mathbf{x}_t \in \mathbb{R}^d$, one restricted projection of it.
Temperature=0 decoding makes the transition deterministic (removing token-sampling stochasticity) and injectivity of layer representations~\citep{nikolaou2025languagemodelsinjectiveinvertible} indicates a single layer retains most information about preceding tokens; what it does \emph{not} expose is the computation at other layers.
We therefore adopt the MDP approximation rather than claiming full observability, treating the single-layer residual stream as a practical sufficient statistic; the training-layer choice then also selects which slice of computation the policy acts on.

\subsection{Adaptive Feature Masking}
\label{subsec:afm}

CRL-Token introduces Adaptive Feature Masking (AFM), a mechanism designed to balance exploration and exploitation during multi-token generation.
AFM maintains a per-sample boolean mask over the SAE dictionary, where $\text{mask}_i = 1$ indicates feature $i$ is available for selection.
At the start of each sample, the mask is initialized to a small subset (e.g., frequently active features), and progressively expanded during generation:

\begin{equation}
\text{mask}_i^{(t+1)} = \text{mask}_i^{(t)} \lor \text{active}_i^{(t)}, \quad \mathbf{a}_t = \text{TopK}(\boldsymbol{\mu} \odot \text{mask}^{(t)}, k)
\end{equation}

where $\text{active}_i^{(t)} = [\mathbf{z}_{t, i} > 0]$ indicates whether feature $i$ was activated at step $t$, and the mask constrains policy selection by excluding unavailable features.
Policy selection ($\mathbf{a}_t$) determines which feature to amplify, while natural activations expand the mask for future steps.
This per-sample masking prevents policy collapse to features selected early while building on activated features within each trajectory.

\subsection{Critic Network for Value Estimation}
The critic $V_\phi: \mathbb{R}^{d} \rightarrow \mathbb{R}$ estimates the state value function $V_\phi(\mathbf{s}) = \mathbb{E}_{\pi_\theta}[r \mid \mathbf{s}]$, implemented as a multilayer perceptron (MLP).
Critic value distributions for single-token tasks (MMLU, BBQ) appear in \hyperref[fig:critic_single_token]{\textcolor{linkblue}{Figure~\ref*{fig:critic_single_token}}} and the GSM8K multi-token trajectory in \hyperref[fig:gsm8k_critic]{\textcolor{linkblue}{Figure~\ref*{fig:gsm8k_critic}}}.

\subsection{Optimization with PPO and Task-Specific Rewards}
The steering coefficient $c$ is determined adaptively by averaging activation magnitudes from correctly predicted training samples, eliminating the need for manual tuning across different layers and tasks.

We train both the policy and critic using Proximal Policy Optimization (PPO).
For feature selection via argmax/top-$k$, the PPO objectives are:
\begin{align}
L_{\text{policy}}(\theta) &= \mathbb{E}\!\left[\min\!\big(r_t(\theta) A_t,\; \text{clip}(r_t(\theta), 1-\epsilon, 1+\epsilon) A_t\big)\right], \\
L_{\text{critic}}(\phi) &= \mathbb{E}\!\left[(V_\phi(\mathbf{s}) - r)^2\right].
\end{align}
Here, $\pi_\theta(\mathbf{a}|\mathbf{s})$ is computed from the softmax probabilities described above,
$r_t(\theta) = \tfrac{\pi_\theta(\mathbf{a}|\mathbf{s})}{\pi_{\theta_{\text{old}}}(\mathbf{a}|\mathbf{s})}$ is the probability ratio
for the selected feature, and $A_t = r - V_\phi(\mathbf{s})$ is the advantage estimate.
PPO's explicit value function enables the critic trajectory analysis in \hyperref[subsec:critic_behavior_analysis]{\textcolor{linkblue}{Section~\ref*{subsec:critic_behavior_analysis}}}, which methods like DPO or GRPO cannot provide.

The reward $r$ is task-specific: binary correctness $r(\hat{y}, y^*)=\mathbb{1}[\hat{y}=y^*]$ for MMLU/MMLU-Pro/BBQ/GSM8K~\citep{shao2024deepseekmathpushinglimitsmathematical}, a RoBERTa-based refusal classifier for HarmBench (inverted for XSTest), and BERT-STS similarity for SimpleQA (\hyperref[subsec:task_rewards]{\textcolor{linkblue}{Appendix~\ref*{subsec:task_rewards}}}).

\subsection{Experimental Setup}
\label{subsec:experimental_setup}

We evaluate CRL using Gemma 2 2B-IT~\citep{gemmateam2024gemma2improvingopen} with Gemma Scope SAEs~\citep{lieberum-etal-2024-gemma} across benchmarks including knowledge (MMLU~\citep{hendrycks2021measuring}), reasoning (GSM8K~\citep{cobbe2021trainingverifierssolvemath}), bias (BBQ~\citep{parrish-etal-2022-bbq}), and safety (HarmBench~\citep{mazeika2024harmbench}, XSTest~\citep{rottger-etal-2024-xstest}) tasks.
The framework is trained using PPO with task-specific rewards to optimize feature selection.
The steering coefficient $c$ in Equation~\ref{eq:steering} is computed by averaging $\|\mathbf{a}_t \mathbf{W}_{dec}\|$ over correctly predicted training samples, adapting to layer-specific activation scales.
Complete experimental details are provided in \hyperref[sec:appendix]{\textcolor{linkblue}{Appendix~\ref*{sec:appendix}}}.

\section{Results}
\label{sec:results}

\subsection{Steering Effectiveness Across Tasks}

CRL-Token adaptively selects an SAE feature per token from the residual stream, producing per-token intervention traces that we analyze in \hyperref[subsec:feature_interpretability]{\textcolor{linkblue}{Section~\ref*{subsec:feature_interpretability}}}.
\hyperref[tab:gemma_results]{\textcolor{linkblue}{Table~\ref*{tab:gemma_results}}} reports the main Gemma 2 2B results under the \textbf{single-layer} CRL-Token configuration.

\begin{table}[t]
  \centering
  \caption{Performance results for Gemma 2 2B model across different tasks using \textbf{single-layer} CRL-Token (0-shot evaluation). The table shows task type, intervention layer, baseline accuracy (Before), CRL accuracy (After), and gain in percentage points. Results are mean $\pm$ std across three seeds.}
  \label{tab:gemma_results}
  \begin{tabular}{lccccc}
  \toprule
  \textbf{Task} & \textbf{Type} & \textbf{Layer} & \textbf{Before} & \textbf{After} & \textbf{Gain} \\
  \midrule
  MMLU & Multi-choice QA & 24 & 52.06$_{\pm 0.23}$ & 55.37$_{\pm 0.16}$ & +3.31 \\
  MMLU-Pro & Multi-choice QA & 25 & 30.30$_{\pm 0.00}$ & 30.49$_{\pm 0.07}$ & +0.19 \\
  BBQ Ambig & Bias QA (ambiguous) & 5 & 60.17$_{\pm 0.01}$ & 65.86$_{\pm 3.03}$ & +5.69 \\
  BBQ Disambig & Bias QA (disambiguated) & 5 & 84.38$_{\pm 0.52}$ & 84.85$_{\pm 0.09}$ & +0.47 \\
  SimpleQA & Short-form QA & 8 & 3.78$_{\pm 0.17}$ & 4.00$_{\pm 0.29}$ & +0.22 \\
  GSM8K & Math reasoning & 20 & 54.62$_{\pm 0.16}$ & 55.65$_{\pm 0.33}$ & +1.03 \\
  HarmBench & Adversarial safety & 21 & 41.46$_{\pm 9.05}$ & 49.12$_{\pm 1.59}$ & +7.66 \\
  XSTest & Over-refusal & 12 & 86.35$_{\pm 0.00}$ & 87.62$_{\pm 0.82}$ & +1.27 \\
  \bottomrule
  \end{tabular}
\end{table}

\noindent\textit{Multi-layer trade-off.}
Multi-layer CRL-Token reaches 87.14\% on HarmBench and 89.84\% on XSTest, and CRL-Layer (shared policy across layers) reaches 71.43\% / 86.98\% (\hyperref[subsec:layer_wise_steering]{\textcolor{linkblue}{Appendix~\ref*{subsec:layer_wise_steering}}}); \textbf{we keep single-layer CRL-Token as the mainline} since it preserves single-feature, single-layer attribution, the paper's interpretability contribution.
The 38\,pp HarmBench gap quantifies the steering-strength price of clean attribution.

SimpleQA's near-zero gain (+0.22) reflects the public SAE's limited factual-recall coverage (\hyperref[subsec:feature_diversity_and_impact_score]{\textcolor{linkblue}{Appendix~\ref*{subsec:feature_diversity_and_impact_score}}}, cf.~\citet{wu2025axbench}); the MMLU +3.31\,pp headline includes constrained-decoding format correction with a +0.36\,pp residual from learned features (\S\ref{subsec:baseline_comparisons}), and HarmBench's +7.66\,pp is evaluator-coupled (\S\ref{sec:limitations}).

\subsection{Baseline Comparisons and Intervention Validation}
\label{subsec:baseline_comparisons}

We validate CRL's effectiveness through constrained decoding comparison, complementarity analysis with fine-tuning, and comparison against alternative steering methods.

\textbf{Constrained Decoding Comparison.}
To isolate CRL's effect from output format correction, we compare against constrained decoding on MMLU.
Constrained decoding restricts outputs to valid answer tokens (A/B/C/D), eliminating format errors.
Under matched single-seed evaluation, base accuracy improves from 51.5\% (unconstrained) to 55.19\% with constrained decoding alone.
CRL achieves 55.55\%, demonstrating +0.36\% improvement beyond format correction.
This confirms that learned features contribute to performance gains beyond output regularization.

\textbf{Complementarity with Fine-tuning.}
CRL provides additive gains when applied on top of supervised fine-tuned models.
On MMLU, accuracy improves from 55.2\% (base) to 55.7\% (SFT) to 56.1\% (SFT + CRL), consistent with learned feature steering providing signals complementary to fine-tuning.
CRL can be applied on top of an SFT model without further weight updates.
The Gemma Scope SAEs transfer effectively across model variants, enabling CRL application without retraining the autoencoder.

\begin{wraptable}{r}{0.5\textwidth}
  \centering
  \small
  \caption{Ablation of feature selection methods. Random+AFM isolates masking; CRL w/o AFM isolates learned selection.}
  \label{tab:steering_baselines}
  \begin{tabular}{lcc}
  \toprule
  \textbf{Method} & \textbf{BBQ Ambig} & \textbf{HarmBench} \\
  \midrule
  No steering (Base) & 60.17 & 41.46 \\
  Random feature & 58.36 & 45.35 \\
  Random + AFM & 60.16 & 46.96 \\
  Most-active feature & 59.94 & 48.03 \\
  CRL w/o AFM & 62.55 & 48.67 \\
  CRL + AFM (Ours) & \textbf{65.86} & \textbf{49.12} \\
  \bottomrule
  \end{tabular}
\end{wraptable}
\textbf{Comparison with Alternative Steering.}
We ablate CRL's components by comparing against heuristic baselines and component-removed variants (\hyperref[tab:steering_baselines]{\textcolor{linkblue}{Table~\ref*{tab:steering_baselines}}}).
CRL without AFM achieves 62.55\% on BBQ Ambig (+2.39 over base) and 48.67\% on HarmBench (+7.21 over base), confirming that the learned policy provides most of the gain.
Adding AFM contributes further (+3.31 BBQ, +0.45 HarmBench) by encouraging diverse feature exploration.
Random+AFM improves over random alone (+1.80 BBQ, +1.61 HarmBench), isolating AFM's independent contribution.
These results confirm that learned feature selection is the primary driver; AFM and constrained decoding provide additional but smaller contributions.

\subsection{Sensitivity to Steering Layer and Coefficients}
\label{subsec:steering_coefficient_analysis}

Optimal layers and coefficients vary by task: later layers generally outperform earlier ones, and large coefficients in early layers degrade performance because residual-stream norms grow with depth~\citep{stefanhex2023residual} so the same scalar produces different effects per layer~\citep{durmus2024steering}.
We use a dataset-averaged static coefficient because three per-token dynamic alternatives (norm-scaled, activation-scaled, critic-gated) all degrade HarmBench (45.00\%, 41.79\%, 43.93\% vs.\ 49.75\% static) and GSM8K (41.39\%, 40.33\%, 41.47\% vs.\ 55.65\%): the policy learns a joint feature-coefficient coupling that inference-time coefficient changes disrupt.
Static-coefficient robustness, layer-wise sweeps, and BBQ ambiguous/disambiguated optima are detailed in \hyperref[sec:layerwise_analysis]{\textcolor{linkblue}{Appendix~\ref*{sec:layerwise_analysis}}} and \hyperref[subsec:runtime_overhead]{\textcolor{linkblue}{Appendix~\ref*{subsec:runtime_overhead}}}.

\subsection{Critic Network Analysis}
\label{subsec:critic_analysis}

\begin{wrapfigure}{r}{0.5\textwidth}
  \centering
  \vspace{-1em}
  \includegraphics[width=0.5\textwidth]{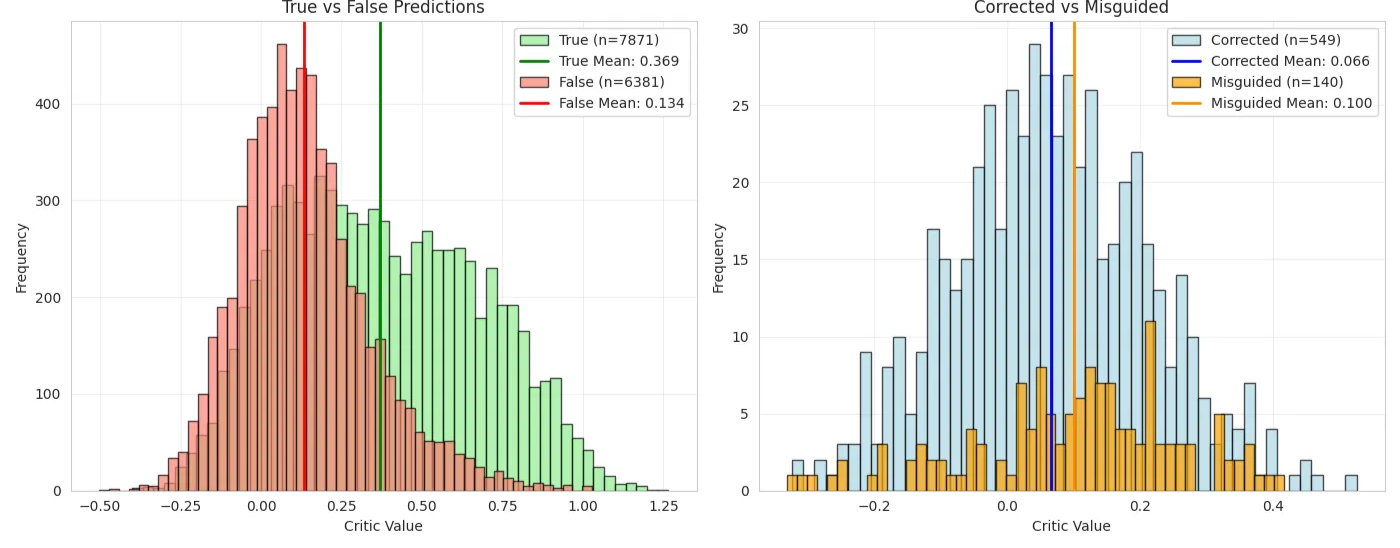} \\
  \includegraphics[width=0.5\textwidth]{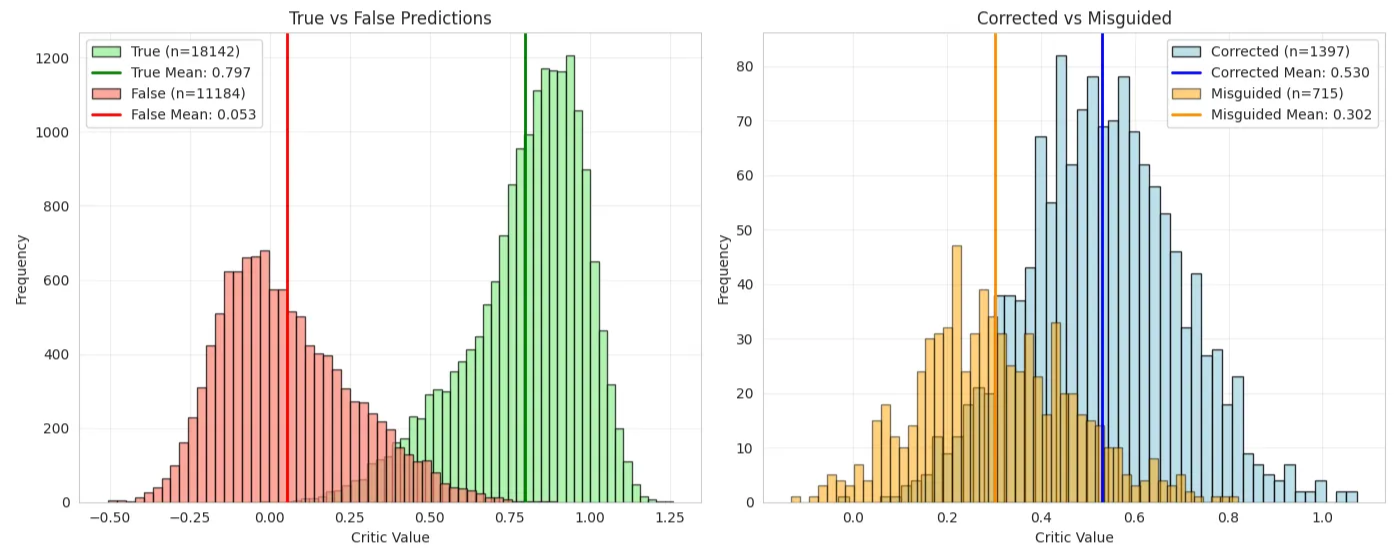}
  \caption{Single-token critic value distributions: MMLU (top), BBQ (bottom). Colors: green = unchanged-correct, red = unchanged-incorrect, blue = corrected, yellow = misguided (\hyperref[subsec:terminology]{\textcolor{linkblue}{Appendix~\ref*{subsec:terminology}}}).}
  \label{fig:critic_single_token}
  \vspace{-1em}
\end{wrapfigure}

The critic network reveals CRL's learning dynamics across different task types.
We analyze critic behavior patterns for both single-token and multi-token generation tasks.

\textbf{Single-token Tasks.}
These tasks are included for the critic-value diagnostic comparison, not because a one-token output requires multi-step sequential decision-making; they isolate the critic's discrimination from multi-token credit assignment.
On MMLU and BBQ, critic values show clear distributions between correct and incorrect samples.
MMLU exhibits a critic bottleneck: corrected and misguided samples (defined in \hyperref[subsec:terminology]{\textcolor{linkblue}{Appendix}}) remain nearly indistinguishable.
BBQ instead shows effective policy-critic coordination, with clear separation between samples.
Detailed single-token critic analysis is provided in \hyperref[subsec:critic_behavior_analysis]{\textcolor{linkblue}{Section~\ref*{subsec:critic_behavior_analysis}}}.

\begin{figure}[t]
  \centering
  \includegraphics[width=0.49\textwidth]{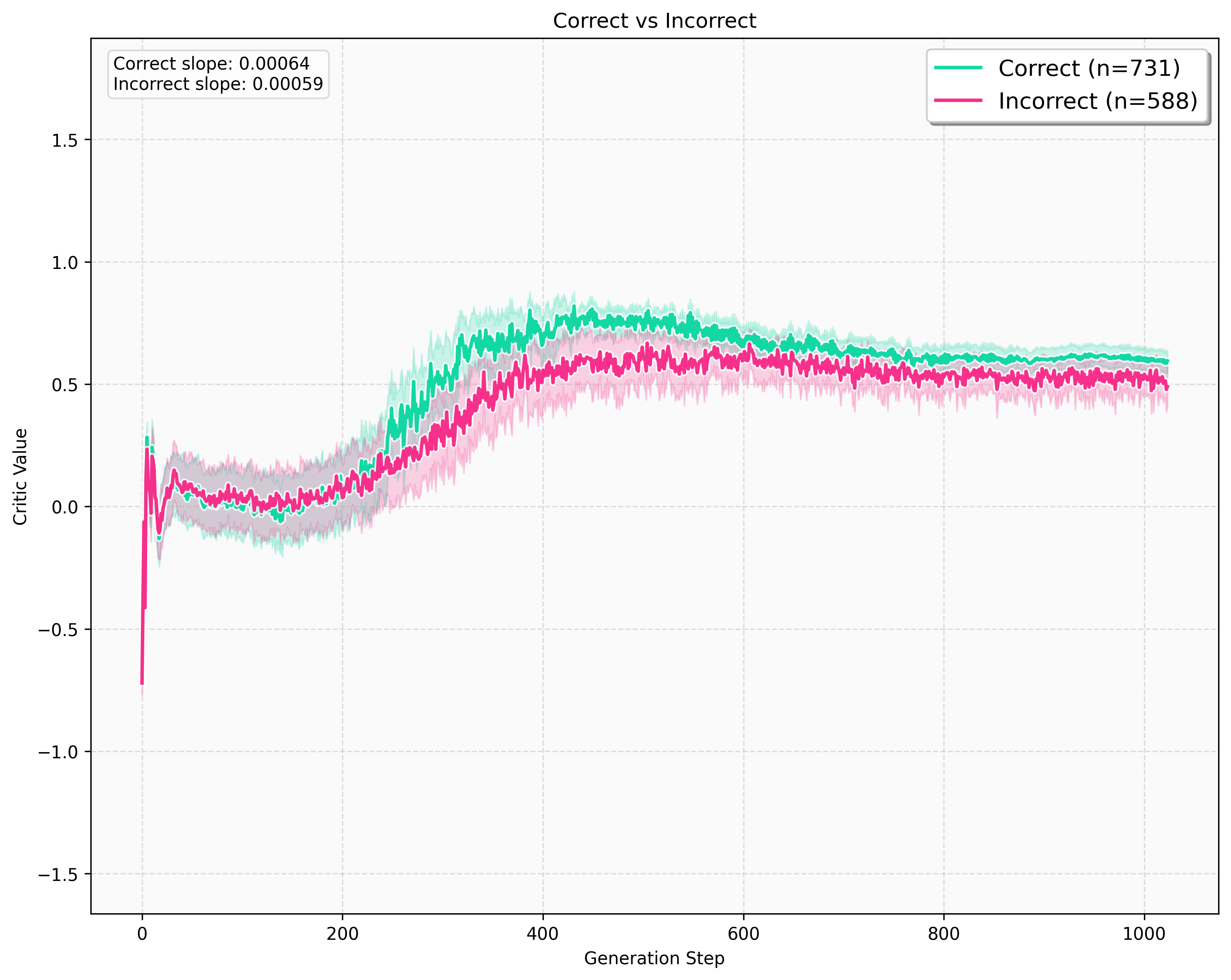}
  \includegraphics[width=0.49\textwidth]{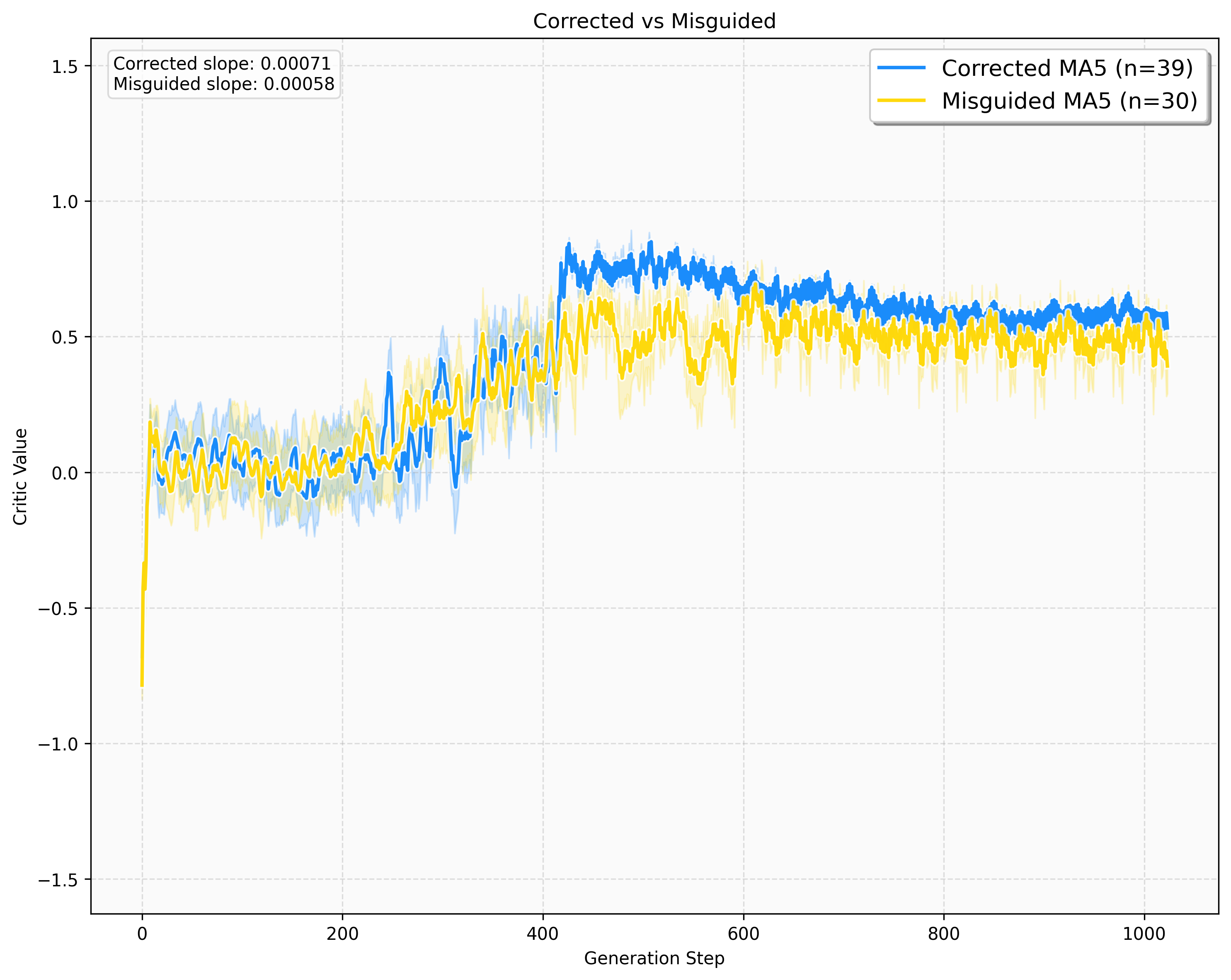}
  \caption{GSM8K critic value trajectories along token generation. Left: unchanged samples (green = correct, red = incorrect). Right: steering-induced changes (blue = corrected, yellow = misguided).}
  \label{fig:gsm8k_critic}
\end{figure}

\textbf{Multi-token Tasks.}
For multi-token tasks like GSM8K, the critic network's value function exhibits climbing patterns along token sequences, indicating effective learning of policy feature selection through AFM (\hyperref[subsec:afm]{\textcolor{linkblue}{Section~\ref*{subsec:afm}}}).
As illustrated in \hyperref[fig:gsm8k_critic]{\textcolor{linkblue}{Figure~\ref*{fig:gsm8k_critic}}}, critic value patterns fall into four groups.
Correct cases display higher gradients with divergence beginning around 200 tokens, while steering-induced cases show different trends with corrected answers exhibiting value increases at the 400-token point.
This indicates that CRL-Token's interventions become more effective in later generation stages.

CRL generalizes to LLaMA-3.1 8B with LLaMA Scope (32K features), with consistent improvements on MMLU (+0.92), MMLU-Pro (+1.03), BBQ Ambig (+1.07), and HarmBench (+5.36); the XSTest decline ($-$1.90) reflects an alignment-vs-opposition axis: reducing over-refusal steers \emph{against} the model's default safety features, which share representations with HarmBench's refusal policy (\hyperref[subsec:llama_results]{\textcolor{linkblue}{Appendix~\ref*{subsec:llama_results}}}).
Feature diversity (Shannon entropy $H$) and impact scores expose task-adaptive exploration: GSM8K spreads broadly ($H$=6.7, impact 0.36), MMLU-Pro is moderate ($H$=5.5, impact 0.63), XSTest intermediate ($H$=4.8, impact 0.36), and HarmBench concentrates on few high-impact refusal features ($H$=2.9, impact 0.81). Low diversity with high impact marks focused tasks where few features suffice; high diversity with moderate impact reflects distributed reasoning across long sequences, confirming the policy adapts to task complexity rather than collapsing to a fixed set (\hyperref[subsec:feature_diversity_and_impact_score]{\textcolor{linkblue}{Appendix~\ref*{subsec:feature_diversity_and_impact_score}}}).
Multi-token HarmBench/XSTest critic patterns appear in \hyperref[subsec:critic_value_analysis]{\textcolor{linkblue}{Appendix~\ref*{subsec:critic_value_analysis}}}.

\section{Discussion}

\subsection{Critic Bottlenecks: BBQ Coordinates, MMLU/XSTest Stagnate}
\label{subsec:critic_behavior_analysis}

Critic value distributions diagnose where improvement originates (\hyperref[fig:critic_single_token]{\textcolor{linkblue}{Figure~\ref*{fig:critic_single_token}}} for single-token tasks; \hyperref[fig:gsm8k_critic]{\textcolor{linkblue}{Figure~\ref*{fig:gsm8k_critic}}} for GSM8K multi-token).
For single-token tasks, MMLU corrected/misguided samples remain indistinguishable (critic bottleneck), while BBQ shows clear separation with corrected samples scoring higher (effective policy-critic coordination).
For multi-token tasks, GSM8K trajectories climb steadily under AFM-driven feature selection, HarmBench separates correct from incorrect samples, and XSTest shows inverted estimation patterns reflecting context-dependent reward challenges (policy bottleneck on HarmBench, critic bottleneck on XSTest; full analysis in \hyperref[subsec:critic_value_analysis]{\textcolor{linkblue}{Appendix~\ref*{subsec:critic_value_analysis}}}).

\begin{figure}[t]
\centering
\includegraphics[width=\textwidth]{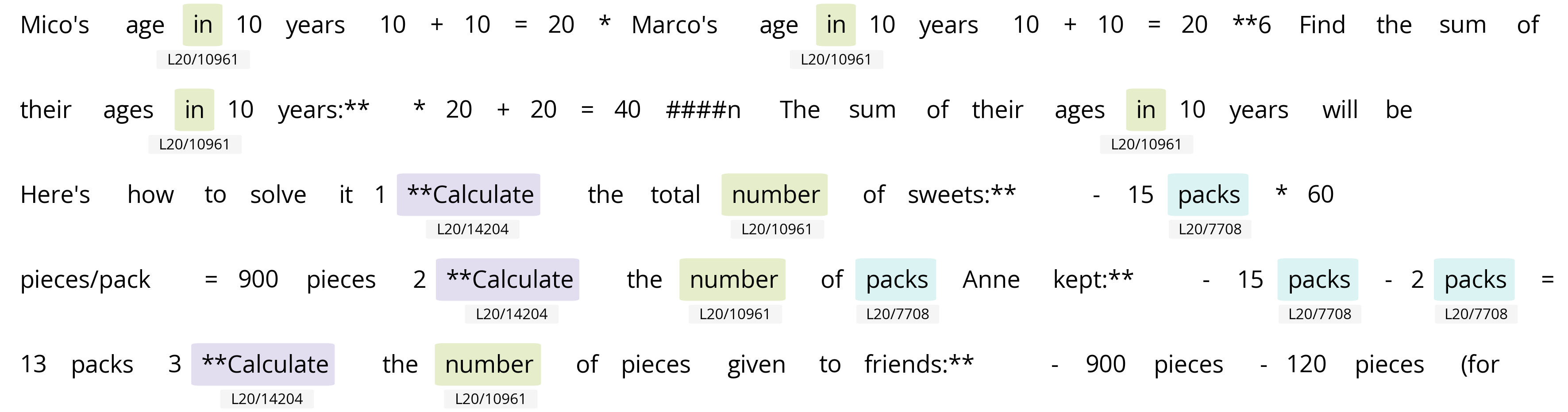} \\
\includegraphics[width=\textwidth]{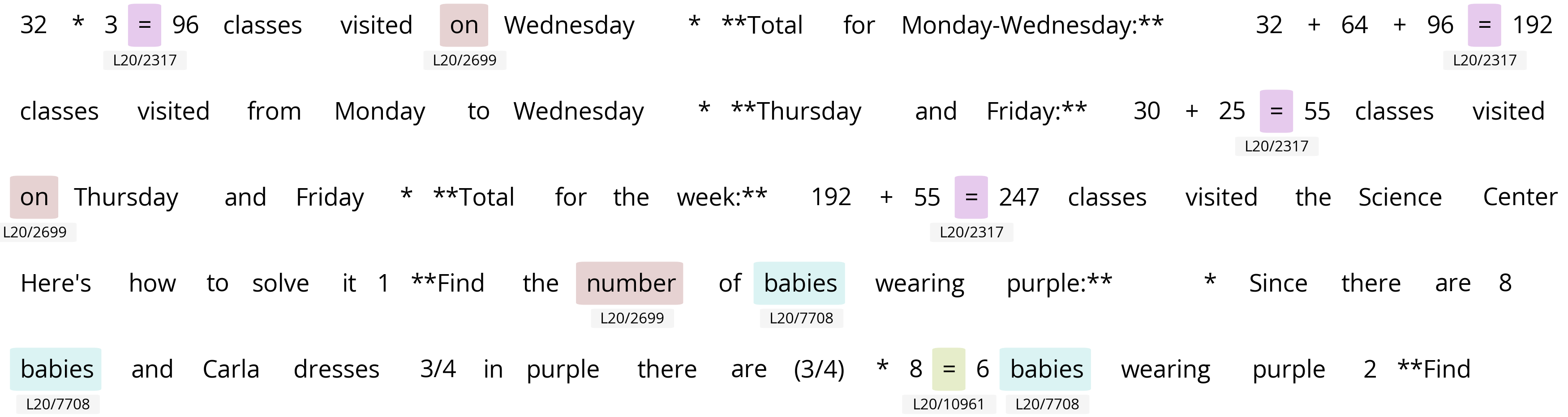}
\caption{GSM8K corrected cases. Top: Feature steering activates relevant numerical tokens. Bottom: Semantically coherent feature activations align with equality tokens.}
\label{fig:corrected_main}
\end{figure}

\subsection{Selected Features and Their Documented Semantics}
\label{subsec:feature_interpretability}

Feature selections track task semantics: the policy identifies features corresponding to reasoning, fact retrieval, and safety.
Successful steering shows semantic coherence (\hyperref[fig:corrected_main]{\textcolor{linkblue}{Figure~\ref*{fig:corrected_main}}}): \textit{statistical methods} activate on mathematical tokens strengthening reasoning chains, and \textit{mathematical operations} demonstrate lexical generalization (activating on units like ``babies'' and ``packs''), while failure cases reveal token-feature misalignment.
Complex reasoning tasks (GSM8K, MMLU-Pro) require broader feature engagement while focused tasks (HarmBench) benefit from concentrated activation (\hyperref[subsec:feature_diversity_and_impact_score]{\textcolor{linkblue}{Appendix~\ref*{subsec:feature_diversity_and_impact_score}}}); additional examples in \hyperref[subsec:additional_corrected_example]{\textcolor{linkblue}{Appendix~\ref*{subsec:additional_corrected_example}}}.

\subsection{Branch Analysis on GSM8K: Depth-Dependent Feature Semantics}
\label{subsec:branch_analysis}

Branch point analysis compares cases where the same problem context produces correct or incorrect outputs depending on which SAE features are selected.
Training separate policies on GSM8K at layers 10 and 20 reveals a clear depth-semantics gradient (representative cases in \hyperref[subsec:additional_branch_analysis]{\textcolor{linkblue}{Appendix~\ref*{subsec:additional_branch_analysis}}}).
In our cases, layer 10 features exhibit concrete, syntactic properties (notation patterns, surface lexical cues), while layer 20 captures abstract, task-relevant semantics (planning structure), consistent with the syntactic-to-semantic depth hierarchy reported in classical probing work~\citep{tenney-etal-2019-bert}.
Practitioners should select intervention depth by target behavior: early layers for syntactic/format control, later layers for semantic reasoning, confirmed by BBQ (optimal layer 5: syntactic) vs.\ MMLU (layer 24: domain knowledge); feature labels alone provide limited predictive value (\hyperref[subsec:additional_branch_analysis]{\textcolor{linkblue}{Appendix~\ref*{subsec:additional_branch_analysis}}}).

\subsection{CRL Logs Diagnose SAE Gaps and Direct Next Steps}
\label{subsec:sae_gaps}

CRL's per-token logs expose SAE quality issues activation-based analysis misses: missing concepts~\citep{engels2025decomposing} and polysemantic features propagate to intervention logs, and policy collapse despite AFM signals SAE deficiency~\citep{wu2025axbench}. Feature-selection entropy $H$ acts as a task-dependent diagnostic: critic bottlenecks (MMLU) call for richer critic features rather than reward redesign, policy bottlenecks (HarmBench) call for selection diversity rather than critic redesign (\hyperref[subsec:feature_diversity_and_impact_score]{\textcolor{linkblue}{Appendix~\ref*{subsec:feature_diversity_and_impact_score}}}). Applying the policy-broadening intervention improves HarmBench by 4.2\,pp over 3 seeds with no consistent benefit on the two critic-bottleneck tasks (\hyperref[subsec:remedy_test]{\textcolor{linkblue}{Appendix~\ref*{subsec:remedy_test}}}).

\subsection{Cross-Task Asymmetry as Directional Interference}
\label{subsec:cross_task_transfer}

To distinguish task-specific mechanisms from dataset-specific shortcuts, we evaluate policies trained on one task against other tasks (\hyperref[tab:cross_task]{\textcolor{linkblue}{Table~\ref*{tab:cross_task}}}).
All experiments use the training task's layer for steering.

\begin{wraptable}[11]{r}{0.5\textwidth}
  \centering
  \small
  \caption{Cross-task transfer. Each row applies the source task's policy at the source-task-optimal layer to the target task; ``vs.\ Base'' is the gap from no-steering at that same source-task layer (not the target-optimal baseline of Table~\ref*{tab:gemma_results}).}
  \label{tab:cross_task}
  \begin{tabular}{lccc}
  \toprule
  \textbf{Train $\to$ Eval} & \textbf{Acc.\ (\%)} & \textbf{vs.\ Base} \\
  \midrule
  MMLU $\to$ GSM8K & 31.01 & $-$10.31 \\
  GSM8K $\to$ MMLU & 52.24 & +0.18 \\
  HarmBench $\to$ XSTest & 86.35 & +1.27 \\
  BBQ $\to$ MMLU & 53.55 & +1.49 \\
  \bottomrule
  \end{tabular}
\end{wraptable}

Transfer is asymmetric: MMLU features harm GSM8K while GSM8K features stay inert on MMLU; HarmBench features slightly aid XSTest. Dataset-specific shortcuts would predict symmetric disruption, which is not observed. The pattern is consistent with shared-but-entangled circuit structure~\citep{lindsey2025biology} with directional interference (MMLU L24 vs.\ GSM8K L20). CRL targets task-conditional attribution, not transfer invariance (\hyperref[subsec:effectiveness_axes]{\textcolor{linkblue}{Appendix~\ref*{subsec:effectiveness_axes}}}).

\section{Limitations}
\label{sec:limitations}

\textbf{Intervention granularity.}
CRL applies one feature per layer per token for clean attribution. A multi-layer variant reaches 87.14\% HarmBench (vs.\ 49.12\% with a single layer) but loses clarity of attribution, and per-token dynamic coefficients underperformed the static dataset-averaged coefficient due to mismatch between training and inference (\S\ref{subsec:steering_coefficient_analysis}). Top-$k$ selection with explicit analysis of interactions is left for future work.

\textbf{SAE dependency.}
CRL inherits its SAE's coverage and quality bounds: missing dictionary concepts~\citep{engels2025decomposing} cannot be steered, polysemantic features may cause proxy selection, and automated feature labels~\citep{choi2024automatic} are noisy. The per-token logs make these failure modes inspectable; collapse to a narrow incoherent feature set despite AFM is a downstream signal of SAE deficiency~\citep{wu2025axbench}.

\textbf{Scalability and model coverage.}
Each task/layer pair requires a separate PPO run (0.5--5.5 A100 GPU-hours, 3.3--3.8\% extra parameters, 8--9\% inference overhead). Experiments use Gemma-2 2B with Gemma Scope (16K) and LLaMA-3.1 8B with LLaMA Scope (32K); scaling AFM to 100K+ dictionaries or to closed-weight frontier models requires efficient architectures and SAEs of matching quality that are not yet publicly available.

\textbf{Reward design and HarmBench evaluator coupling.}
Safety tasks use the same refusal classifier as both PPO reward and test-time evaluator; the +7.66\,pp HarmBench gain partly reflects optimization against the evaluator and is not directly comparable to leaderboards using automated jailbreak attacks. Held-out evaluators are needed to disentangle these factors (\hyperref[subsec:extended_limitations]{\textcolor{linkblue}{Appendix~\ref*{subsec:extended_limitations}}}).

\section{Conclusion}

Control Reinforcement Learning turns SAE feature steering into a mechanistic interpretability instrument by casting token-level intervention at a single layer as a control problem (approximated as an MDP) and training a PPO policy whose per-token selections are themselves the interpretability artifact.
The framework yields three diagnostic tools that observation-only attribution and fixed-direction steering do not jointly provide: branch tracking locates tokens where feature choice flips outcomes, critic trajectories separate bottlenecks in policy selection from those in value estimation, and the depth of the training layer selects the granularity of attribution (syntactic in early layers, semantic in later layers), while Adaptive Feature Masking keeps selections legible.

The diagnostics expose mechanism rather than headline accuracy (MMLU critic bottleneck, HarmBench policy bottleneck), and asymmetric cross-task transfer is consistent with partial circuit overlap~\citep{lindsey2025biology}; CRL complements activation-based SAE analysis as a layer-specific intervention diagnostic.

\begin{ack}
This work received no external or grant funding. It was carried out as an MSc project under University College London's Industry Exchange Network (IXN), in collaboration with Holistic AI. The authors declare no competing interests.
\end{ack}

\bibliographystyle{plainnat}
\bibliography{mechinterp}

\begin{thebibliography}{50}
\providecommand{\natexlab}[1]{#1}
\providecommand{\url}[1]{\texttt{#1}}
\expandafter\ifx\csname urlstyle\endcsname\relax
  \providecommand{\doi}[1]{doi: #1}\else
  \providecommand{\doi}{doi: \begingroup \urlstyle{rm}\Url}\fi

\bibitem[Arad et~al.(2025)Arad, Mueller, and
  Belinkov]{arad2025saesgoodsteering}
Dana Arad, Aaron Mueller, and Yonatan Belinkov.
\newblock Saes are good for steering -- if you select the right features, 2025.
\newblock URL \url{https://arxiv.org/abs/2505.20063}.

\bibitem[Bricken et~al.(2023)Bricken, Templeton, Batson, Chen, Jermyn, Conerly,
  Turner, Anil, Denison, Askell, Lasenby, Wu, Kravec, Schiefer, Maxwell,
  Joseph, Hatfield-Dodds, Tamkin, Nguyen, McLean, Burke, Hume, Carter,
  Henighan, and Olah]{bricken2023monosemanticity}
Trenton Bricken, Adly Templeton, Joshua Batson, Brian Chen, Adam Jermyn, Tom
  Conerly, Nick Turner, Cem Anil, Carson Denison, Amanda Askell, Robert
  Lasenby, Yifan Wu, Shauna Kravec, Nicholas Schiefer, Tim Maxwell, Nicholas
  Joseph, Zac Hatfield-Dodds, Alex Tamkin, Karina Nguyen, Brayden McLean,
  Josiah~E Burke, Tristan Hume, Shan Carter, Tom Henighan, and Christopher
  Olah.
\newblock Towards monosemanticity: Decomposing language models with dictionary
  learning.
\newblock \emph{Transformer Circuits Thread}, 2023.
\newblock
  https://transformer-circuits.pub/2023/monosemantic-features/index.html.

\bibitem[Chalnev et~al.(2024)Chalnev, Siu, and
  Conmy]{chalnev2024improvingsteeringvectorstargeting}
Sviatoslav Chalnev, Matthew Siu, and Arthur Conmy.
\newblock Improving steering vectors by targeting sparse autoencoder features,
  2024.
\newblock URL \url{https://arxiv.org/abs/2411.02193}.

\bibitem[Cho et~al.(2026)Cho, Wu, and Koshiyama]{cho2025corrsteer}
Seonglae Cho, Zekun Wu, and Adriano Koshiyama.
\newblock Corrsteer: Generation-time {LLM} steering via correlated sparse
  autoencoder features.
\newblock In \emph{Forty-third International Conference on Machine Learning},
  2026.

\bibitem[Choi et~al.(2024)Choi, Huang, Meng, Johnson, Steinhardt, and
  Schwettmann]{choi2024automatic}
Dami Choi, Vincent Huang, Kevin Meng, Daniel~D Johnson, Jacob Steinhardt, and
  Sarah Schwettmann.
\newblock Scaling automatic neuron description.
\newblock \url{https://transluce.org/neuron-descriptions}, October 2024.

\bibitem[Cobbe et~al.(2021)Cobbe, Kosaraju, Bavarian, Chen, Jun, Kaiser,
  Plappert, Tworek, Hilton, Nakano, Hesse, and
  Schulman]{cobbe2021trainingverifierssolvemath}
Karl Cobbe, Vineet Kosaraju, Mohammad Bavarian, Mark Chen, Heewoo Jun, Lukasz
  Kaiser, Matthias Plappert, Jerry Tworek, Jacob Hilton, Reiichiro Nakano,
  Christopher Hesse, and John Schulman.
\newblock Training verifiers to solve math word problems, 2021.
\newblock URL \url{https://arxiv.org/abs/2110.14168}.

\bibitem[Conmy et~al.(2023)Conmy, Mavor-Parker, Lynch, Heimersheim, and
  Garriga-Alonso]{conmy2023towards}
Arthur Conmy, Augustine~N. Mavor-Parker, Aengus Lynch, Stefan Heimersheim, and
  Adri{\`a} Garriga-Alonso.
\newblock Towards automated circuit discovery for mechanistic interpretability.
\newblock In \emph{Thirty-seventh Conference on Neural Information Processing
  Systems}, 2023.
\newblock URL \url{https://openreview.net/forum?id=89ia77nZ8u}.

\bibitem[DeepSeek-AI(2025)]{deepseekai2025deepseekr1}
DeepSeek-AI.
\newblock Deepseek-r1: Incentivizing reasoning capability in llms via
  reinforcement learning, 2025.
\newblock URL \url{https://arxiv.org/abs/2501.12948}.

\bibitem[Durmus et~al.(2024)Durmus, Tamkin, Clark, Wei, Marcus, Batson, Handa,
  Lovitt, Tong, McCain, Rausch, Huang, Bowman, Ritchie, Henighan, and
  Ganguli]{durmus2024steering}
Esin Durmus, Alex Tamkin, Jack Clark, Jerry Wei, Jonathan Marcus, Joshua
  Batson, Kunal Handa, Liane Lovitt, Meg Tong, Miles McCain, Oliver Rausch,
  Saffron Huang, Sam Bowman, Stuart Ritchie, Tom Henighan, and Deep Ganguli.
\newblock Evaluating feature steering: A case study in mitigating social
  biases.
\newblock \url{https://anthropic.com/research/evaluating-feature-steering},
  2024.
\newblock Anthropic Research.

\bibitem[Elhage et~al.(2021)Elhage, Nanda, Olsson, Henighan, Joseph, Mann,
  Askell, Bai, Chen, Conerly, DasSarma, Drain, Ganguli, Hatfield-Dodds,
  Hernandez, Jones, Kernion, Lovitt, Ndousse, Amodei, Brown, Clark, Kaplan,
  McCandlish, and Olah]{elhage2021mathematical}
Nelson Elhage, Neel Nanda, Catherine Olsson, Tom Henighan, Nicholas Joseph, Ben
  Mann, Amanda Askell, Yuntao Bai, Anna Chen, Tom Conerly, Nova DasSarma, Dawn
  Drain, Deep Ganguli, Zac Hatfield-Dodds, Danny Hernandez, Andy Jones, Jackson
  Kernion, Liane Lovitt, Kamal Ndousse, Dario Amodei, Tom Brown, Jack Clark,
  Jared Kaplan, Sam McCandlish, and Chris Olah.
\newblock A mathematical framework for transformer circuits.
\newblock \emph{Transformer Circuits Thread}, 2021.
\newblock https://transformer-circuits.pub/2021/framework/index.html.

\bibitem[Elhage et~al.(2022)Elhage, Hume, Olsson, Schiefer, Henighan, Kravec,
  Hatfield-Dodds, Lasenby, Drain, Chen, Grosse, McCandlish, Kaplan, Amodei,
  Wattenberg, and Olah]{elhage2022superposition}
Nelson Elhage, Tristan Hume, Catherine Olsson, Nicholas Schiefer, Tom Henighan,
  Shauna Kravec, Zac Hatfield-Dodds, Robert Lasenby, Dawn Drain, Carol Chen,
  Roger Grosse, Sam McCandlish, Jared Kaplan, Dario Amodei, Martin Wattenberg,
  and Christopher Olah.
\newblock Toy models of superposition.
\newblock \emph{Transformer Circuits Thread}, 2022.
\newblock URL
  \url{https://transformer-circuits.pub/2022/toy\_model/index.html}.

\bibitem[Elhoushi et~al.(2024)Elhoushi, Shrivastava, Liskovich, Hosmer, Wasti,
  Lai, Mahmoud, Acun, Agarwal, Roman, Aly, Chen, and Wu]{elhoushi2024}
Mostafa Elhoushi, Akshat Shrivastava, Diana Liskovich, Basil Hosmer, Bram
  Wasti, Liangzhen Lai, Anas Mahmoud, Bilge Acun, Saurabh Agarwal, Ahmed Roman,
  Ahmed~A Aly, Beidi Chen, and Carole-Jean Wu.
\newblock Layerskip: Enabling early exit inference and self-speculative
  decoding.
\newblock In \emph{Proceedings of the 62nd Annual Meeting of the Association
  for Computational Linguistics}, pages 12622--12642, 2024.
\newblock URL \url{https://doi.org/10.18653/v1/2024.acl-long.681}.

\bibitem[Engels et~al.(2025)Engels, Riggs, and Tegmark]{engels2025decomposing}
Joshua Engels, Logan Riggs, and Max Tegmark.
\newblock Decomposing the dark matter of sparse autoencoders.
\newblock \emph{Transactions on Machine Learning Research}, 2025.

\bibitem[Ferrao(2024)]{ferrao2024autosteer}
Jeremias~Lino Ferrao.
\newblock Autosteer: Weight-preserving reinforcement learning for interpretable
  model control.
\newblock https://apartresearch.com, 2024.
\newblock Research submission to the research sprint hosted by Apart.

\bibitem[Ferrao et~al.(2025)Ferrao, van~der Lende, Lichkovski, and
  Neo]{ferrao2025fsrl}
Jeremias~Lino Ferrao, Matthijs van~der Lende, Ilija Lichkovski, and Clement
  Neo.
\newblock The anatomy of alignment: Decomposing preference optimization by
  steering sparse features.
\newblock In \emph{Mechanistic Interpretability Workshop at NeurIPS 2025},
  2025.
\newblock URL \url{https://openreview.net/forum?id=CoaaltVlx7}.

\bibitem[Gui et~al.(2019)Gui, Zhang, Zhao, Lin, Peng, Gong, and
  Huang]{gui2019long}
Tao Gui, Qi~Zhang, Lujun Zhao, Yaosong Lin, Minlong Peng, Jingjing Gong, and
  Xuanjing Huang.
\newblock Long short-term memory with dynamic skip connections.
\newblock \emph{Proceedings of the AAAI Conference on Artificial Intelligence},
  33\penalty0 (01):\penalty0 6481--6488, Jul. 2019.
\newblock \doi{10.1609/aaai.v33i01.33016481}.
\newblock URL \url{https://ojs.aaai.org/index.php/AAAI/article/view/4613}.

\bibitem[He et~al.(2024)He, Shu, Ge, Chen, Wang, Zhou, Liu, Guo, Huang, Wu,
  Jiang, and Qiu]{he2024llamascopeextractingmillions}
Zhengfu He, Wentao Shu, Xuyang Ge, Lingjie Chen, Junxuan Wang, Yunhua Zhou,
  Frances Liu, Qipeng Guo, Xuanjing Huang, Zuxuan Wu, Yu-Gang Jiang, and Xipeng
  Qiu.
\newblock Llama scope: Extracting millions of features from llama-3.1-8b with
  sparse autoencoders, 2024.
\newblock URL \url{https://arxiv.org/abs/2410.20526}.

\bibitem[Heimersheim and Turner(2023)]{stefanhex2023residual}
Stefan Heimersheim and Alex Turner.
\newblock Residual stream norms grow exponentially over the forward pass, May
  2023.
\newblock URL
  \url{https://www.lesswrong.com/posts/8mizBCm3dyc432nK8/residual-stream-norms-grow-exponentially-over-the-forward}.
\newblock LessWrong, AI Alignment Forum.

\bibitem[Hendrycks et~al.(2021)Hendrycks, Burns, Basart, Zou, Mazeika, Song,
  and Steinhardt]{hendrycks2021measuring}
Dan Hendrycks, Collin Burns, Steven Basart, Andy Zou, Mantas Mazeika, Dawn
  Song, and Jacob Steinhardt.
\newblock Measuring massive multitask language understanding.
\newblock In \emph{International Conference on Learning Representations}, 2021.
\newblock URL \url{https://openreview.net/forum?id=d7KBjmI3GmQ}.

\bibitem[Hernandez et~al.(2024)Hernandez, Li, and
  Andreas]{hernandez2024inspecting}
Evan Hernandez, Belinda~Z. Li, and Jacob Andreas.
\newblock Inspecting and editing knowledge representations in language models.
\newblock In \emph{First Conference on Language Modeling}, 2024.
\newblock URL \url{https://openreview.net/forum?id=ADtL6fgNRv}.

\bibitem[Huben et~al.(2024)Huben, Cunningham, Smith, Ewart, and
  Sharkey]{huben2024sparse}
Robert Huben, Hoagy Cunningham, Logan~Riggs Smith, Aidan Ewart, and Lee
  Sharkey.
\newblock Sparse autoencoders find highly interpretable features in language
  models.
\newblock In \emph{The Twelfth International Conference on Learning
  Representations}, 2024.
\newblock URL \url{https://openreview.net/forum?id=F76bwRSLeK}.

\bibitem[Konen et~al.(2024)Konen, Jentzsch, Diallo, Sch{\"u}tt, Bensch,
  El~Baff, Opitz, and Hecking]{konen-etal-2024-style}
Kai Konen, Sophie Jentzsch, Diaoul{\'e} Diallo, Peer Sch{\"u}tt, Oliver Bensch,
  Roxanne El~Baff, Dominik Opitz, and Tobias Hecking.
\newblock Style vectors for steering generative large language models.
\newblock In Yvette Graham and Matthew Purver, editors, \emph{Findings of the
  Association for Computational Linguistics: EACL 2024}, pages 782--802, St.
  Julian{'}s, Malta, March 2024. Association for Computational Linguistics.
\newblock URL \url{https://aclanthology.org/2024.findings-eacl.52/}.

\bibitem[Lieberum et~al.(2024)Lieberum, Rajamanoharan, Conmy, Smith, Sonnerat,
  Varma, Kramar, Dragan, Shah, and Nanda]{lieberum-etal-2024-gemma}
Tom Lieberum, Senthooran Rajamanoharan, Arthur Conmy, Lewis Smith, Nicolas
  Sonnerat, Vikrant Varma, Janos Kramar, Anca Dragan, Rohin Shah, and Neel
  Nanda.
\newblock Gemma scope: Open sparse autoencoders everywhere all at once on gemma
  2.
\newblock In Yonatan Belinkov, Najoung Kim, Jaap Jumelet, Hosein Mohebbi, Aaron
  Mueller, and Hanjie Chen, editors, \emph{Proceedings of the 7th BlackboxNLP
  Workshop: Analyzing and Interpreting Neural Networks for NLP}, pages
  278--300, Miami, Florida, US, November 2024. Association for Computational
  Linguistics.
\newblock \doi{10.18653/v1/2024.blackboxnlp-1.19}.
\newblock URL \url{https://aclanthology.org/2024.blackboxnlp-1.19/}.

\bibitem[Lim et~al.(2023)Lim, Becker, Kochenderfer, Tomlin, and
  Sunberg]{lim2023optimality}
Michael Lim, Nathan Becker, Mykel~J Kochenderfer, Claire~J Tomlin, and
  Zachary~N Sunberg.
\newblock Optimality guarantees for particle belief approximation of pomdps.
\newblock \emph{Journal of Artificial Intelligence Research}, 77:\penalty0
  1591--1636, 2023.
\newblock \doi{10.1613/jair.1.14491}.
\newblock URL \url{https://jair.org/index.php/jair/article/view/14491}.

\bibitem[Lindsey et~al.(2025)Lindsey, Gurnee, Ameisen, Chen, Pearce, Turner,
  Citro, Abrahams, Carter, Hosmer, Marcus, Sklar, Templeton, Bricken,
  McDougall, Cunningham, Henighan, Jermyn, Jones, Persic, Qi, Thompson,
  Zimmerman, Rivoire, Conerly, Olah, and Batson]{lindsey2025biology}
Jack Lindsey, Wes Gurnee, Emmanuel Ameisen, Brian Chen, Adam Pearce,
  Nicholas~L. Turner, Craig Citro, David Abrahams, Shan Carter, Basil Hosmer,
  Jonathan Marcus, Michael Sklar, Adly Templeton, Trenton Bricken, Callum
  McDougall, Hoagy Cunningham, Thomas Henighan, Adam Jermyn, Andy Jones, Andrew
  Persic, Zhenyi Qi, T.~Ben Thompson, Sam Zimmerman, Kelley Rivoire, Thomas
  Conerly, Chris Olah, and Joshua Batson.
\newblock On the biology of a large language model.
\newblock \emph{Transformer Circuits Thread}, 2025.
\newblock URL
  \url{https://transformer-circuits.pub/2025/attribution-graphs/biology.html}.

\bibitem[Marks et~al.(2025)Marks, Rager, Michaud, Belinkov, Bau, and
  Mueller]{marks2025sparse}
Samuel Marks, Can Rager, Eric~J Michaud, Yonatan Belinkov, David Bau, and Aaron
  Mueller.
\newblock Sparse feature circuits: Discovering and editing interpretable causal
  graphs in language models.
\newblock In \emph{The Thirteenth International Conference on Learning
  Representations}, 2025.
\newblock URL \url{https://openreview.net/forum?id=I4e82CIDxv}.

\bibitem[Mazeika et~al.(2024)Mazeika, Phan, Yin, Zou, Wang, Mu, Sakhaee, Li,
  Basart, Li, Forsyth, and Hendrycks]{mazeika2024harmbench}
Mantas Mazeika, Long Phan, Xuwang Yin, Andy Zou, Zifan Wang, Norman Mu, Elham
  Sakhaee, Nathaniel Li, Steven Basart, Bo~Li, David Forsyth, and Dan
  Hendrycks.
\newblock Harmbench: A standardized evaluation framework for automated red
  teaming and robust refusal.
\newblock In \emph{Forty-first International Conference on Machine Learning},
  2024.
\newblock URL \url{https://openreview.net/forum?id=f3TUipYU3U}.

\bibitem[Nikolaou et~al.(2025)Nikolaou, Mencattini, Crisostomi, Santilli,
  Panagakis, and Rodolà]{nikolaou2025languagemodelsinjectiveinvertible}
Giorgos Nikolaou, Tommaso Mencattini, Donato Crisostomi, Andrea Santilli,
  Yannis Panagakis, and Emanuele Rodolà.
\newblock Language models are injective and hence invertible, 2025.
\newblock URL \url{https://arxiv.org/abs/2510.15511}.

\bibitem[Olah et~al.(2020)Olah, Cammarata, Schubert, Goh, Petrov, and
  Carter]{olah2020zoom}
Chris Olah, Nick Cammarata, Ludwig Schubert, Gabriel Goh, Michael Petrov, and
  Shan Carter.
\newblock Zoom in: An introduction to circuits.
\newblock \emph{Distill}, 2020.
\newblock \doi{10.23915/distill.00024.001}.
\newblock https://distill.pub/2020/circuits/zoom-in.

\bibitem[Parrish et~al.(2022)Parrish, Chen, Nangia, Padmakumar, Phang,
  Thompson, Htut, and Bowman]{parrish-etal-2022-bbq}
Alicia Parrish, Angelica Chen, Nikita Nangia, Vishakh Padmakumar, Jason Phang,
  Jana Thompson, Phu~Mon Htut, and Samuel Bowman.
\newblock {BBQ}: A hand-built bias benchmark for question answering.
\newblock In Smaranda Muresan, Preslav Nakov, and Aline Villavicencio, editors,
  \emph{Findings of the Association for Computational Linguistics: ACL 2022},
  pages 2086--2105, Dublin, Ireland, May 2022. Association for Computational
  Linguistics.
\newblock \doi{10.18653/v1/2022.findings-acl.165}.
\newblock URL \url{https://aclanthology.org/2022.findings-acl.165/}.

\bibitem[Rajamanoharan et~al.(2024)Rajamanoharan, Lieberum, Sonnerat, Conmy,
  Varma, Kramár, and
  Nanda]{rajamanoharan2024jumpingaheadimprovingreconstruction}
Senthooran Rajamanoharan, Tom Lieberum, Nicolas Sonnerat, Arthur Conmy, Vikrant
  Varma, János Kramár, and Neel Nanda.
\newblock Jumping ahead: Improving reconstruction fidelity with jumprelu sparse
  autoencoders, 2024.
\newblock URL \url{https://arxiv.org/abs/2407.14435}.

\bibitem[Raposo et~al.(2024)Raposo, Ritter, Richards, Lillicrap, Humphreys, and
  Santoro]{raposo2024mixtureofdepths}
David Raposo, Sam Ritter, Blake Richards, Timothy Lillicrap, Peter~Conway
  Humphreys, and Adam Santoro.
\newblock Mixture-of-depths: Dynamically allocating compute in
  transformer-based language models, 2024.
\newblock URL \url{https://arxiv.org/abs/2404.02258}.

\bibitem[Rimsky et~al.(2024)Rimsky, Gabrieli, Schulz, Tong, Hubinger, and
  Turner]{rimsky-etal-2024-steering}
Nina Rimsky, Nick Gabrieli, Julian Schulz, Meg Tong, Evan Hubinger, and
  Alexander Turner.
\newblock Steering llama 2 via contrastive activation addition.
\newblock In Lun-Wei Ku, Andre Martins, and Vivek Srikumar, editors,
  \emph{Proceedings of the 62nd Annual Meeting of the Association for
  Computational Linguistics (Volume 1: Long Papers)}, pages 15504--15522,
  Bangkok, Thailand, August 2024. Association for Computational Linguistics.
\newblock \doi{10.18653/v1/2024.acl-long.828}.
\newblock URL \url{https://aclanthology.org/2024.acl-long.828/}.

\bibitem[R{\"o}ttger et~al.(2024)R{\"o}ttger, Kirk, Vidgen, Attanasio, Bianchi,
  and Hovy]{rottger-etal-2024-xstest}
Paul R{\"o}ttger, Hannah Kirk, Bertie Vidgen, Giuseppe Attanasio, Federico
  Bianchi, and Dirk Hovy.
\newblock {XST}est: A test suite for identifying exaggerated safety behaviours
  in large language models.
\newblock In Kevin Duh, Helena Gomez, and Steven Bethard, editors,
  \emph{Proceedings of the 2024 Conference of the North American Chapter of the
  Association for Computational Linguistics: Human Language Technologies
  (Volume 1: Long Papers)}, pages 5377--5400, Mexico City, Mexico, June 2024.
  Association for Computational Linguistics.
\newblock \doi{10.18653/v1/2024.naacl-long.301}.
\newblock URL \url{https://aclanthology.org/2024.naacl-long.301/}.

\bibitem[Schulman et~al.(2017)Schulman, Wolski, Dhariwal, Radford, and
  Klimov]{schulman2017proximal}
John Schulman, Filip Wolski, Prafulla Dhariwal, Alec Radford, and Oleg Klimov.
\newblock Proximal policy optimization algorithms.
\newblock \url{https://openai.com/index/openai-baselines-ppo/}, 2017.
\newblock URL \url{https://arxiv.org/abs/1707.06347}.

\bibitem[Shao et~al.(2024)Shao, Wang, Zhu, Xu, Song, Bi, Zhang, Zhang, Li, Wu,
  and Guo]{shao2024deepseekmathpushinglimitsmathematical}
Zhihong Shao, Peiyi Wang, Qihao Zhu, Runxin Xu, Junxiao Song, Xiao Bi, Haowei
  Zhang, Mingchuan Zhang, Y.~K. Li, Y.~Wu, and Daya Guo.
\newblock Deepseekmath: Pushing the limits of mathematical reasoning in open
  language models, 2024.
\newblock URL \url{https://arxiv.org/abs/2402.03300}.

\bibitem[Sinii et~al.(2025)Sinii, Balagansky, Aksenov, Kurochkin, Laptev,
  Gorbatovski, Shaposhnikov, and Gavrilov]{sinii2025small}
Viacheslav Sinii, Nikita Balagansky, Yaroslav Aksenov, Vadim Kurochkin, Daniil
  Laptev, Alexey Gorbatovski, Boris Shaposhnikov, and Daniil Gavrilov.
\newblock Small vectors, big effects: A mechanistic study of {RL}-induced
  reasoning via steering vectors.
\newblock In \emph{Mechanistic Interpretability Workshop at NeurIPS 2025},
  2025.
\newblock URL \url{https://openreview.net/forum?id=M8WDG1TfBb}.

\bibitem[Subramani et~al.(2022)Subramani, Suresh, and
  Peters]{subramani-etal-2022-extracting}
Nishant Subramani, Nivedita Suresh, and Matthew Peters.
\newblock Extracting latent steering vectors from pretrained language models.
\newblock In Smaranda Muresan, Preslav Nakov, and Aline Villavicencio, editors,
  \emph{Findings of the Association for Computational Linguistics: ACL 2022},
  pages 566--581, Dublin, Ireland, May 2022. Association for Computational
  Linguistics.
\newblock \doi{10.18653/v1/2022.findings-acl.48}.
\newblock URL \url{https://aclanthology.org/2022.findings-acl.48/}.

\bibitem[Team et~al.(2024{\natexlab{a}})Team, Riviere, Pathak, Sessa, Hardin,
  Bhupatiraju, Hussenot, Mesnard, Shahriari, Ramé, Ferret, Liu, Tafti,
  Friesen, Casbon, Ramos, Kumar, Lan, Jerome, Tsitsulin, Vieillard, Stanczyk,
  Girgin, Momchev, Hoffman, Thakoor, Grill, Neyshabur, Bachem, Walton, Severyn,
  Parrish, Ahmad, Hutchison, Abdagic, Carl, Shen, Brock, Coenen, Laforge,
  Paterson, Bastian, Piot, Wu, Royal, Chen, Kumar, Perry, Welty,
  Choquette-Choo, Sinopalnikov, Weinberger, Vijaykumar, Rogozińska, Herbison,
  Bandy, Wang, Noland, Moreira, Senter, Eltyshev, Visin, Rasskin, Wei, Cameron,
  Martins, Hashemi, Klimczak-Plucińska, Batra, Dhand, Nardini, Mein, Zhou,
  Svensson, Stanway, Chan, Zhou, Carrasqueira, Iljazi, Becker, Fernandez, van
  Amersfoort, Gordon, Lipschultz, Newlan, yeong Ji, Mohamed, Badola, Black,
  Millican, McDonell, Nguyen, Sodhia, Greene, Sjoesund, Usui, Sifre, Heuermann,
  Lago, McNealus, Soares, Kilpatrick, Dixon, Martins, Reid, Singh, Iverson,
  Görner, Velloso, Wirth, Davidow, Miller, Rahtz, Watson, Risdal, Kazemi,
  Moynihan, Zhang, Kahng, Park, Rahman, Khatwani, Dao, Bardoliwalla,
  Devanathan, Dumai, Chauhan, Wahltinez, Botarda, Barnes, Barham, Michel, Jin,
  Georgiev, Culliton, Kuppala, Comanescu, Merhej, Jana, Rokni, Agarwal,
  Mullins, Saadat, Carthy, Cogan, Perrin, Arnold, Krause, Dai, Garg, Sheth,
  Ronstrom, Chan, Jordan, Yu, Eccles, Hennigan, Kocisky, Doshi, Jain, Yadav,
  Meshram, Dharmadhikari, Barkley, Wei, Ye, Han, Kwon, Xu, Shen, Gong, Wei,
  Cotruta, Kirk, Rao, Giang, Peran, Warkentin, Collins, Barral, Ghahramani,
  Hadsell, Sculley, Banks, Dragan, Petrov, Vinyals, Dean, Hassabis,
  Kavukcuoglu, Farabet, Buchatskaya, Borgeaud, Fiedel, Joulin, Kenealy,
  Dadashi, and Andreev]{gemmateam2024gemma2improvingopen}
Gemma Team, Morgane Riviere, Shreya Pathak, Pier~Giuseppe Sessa, Cassidy
  Hardin, Surya Bhupatiraju, Léonard Hussenot, Thomas Mesnard, Bobak
  Shahriari, Alexandre Ramé, Johan Ferret, Peter Liu, Pouya Tafti, Abe
  Friesen, Michelle Casbon, Sabela Ramos, Ravin Kumar, Charline~Le Lan, Sammy
  Jerome, Anton Tsitsulin, Nino Vieillard, Piotr Stanczyk, Sertan Girgin,
  Nikola Momchev, Matt Hoffman, Shantanu Thakoor, Jean-Bastien Grill, Behnam
  Neyshabur, Olivier Bachem, Alanna Walton, Aliaksei Severyn, Alicia Parrish,
  Aliya Ahmad, Allen Hutchison, Alvin Abdagic, Amanda Carl, Amy Shen, Andy
  Brock, Andy Coenen, Anthony Laforge, Antonia Paterson, Ben Bastian, Bilal
  Piot, Bo~Wu, Brandon Royal, Charlie Chen, Chintu Kumar, Chris Perry, Chris
  Welty, Christopher~A. Choquette-Choo, Danila Sinopalnikov, David Weinberger,
  Dimple Vijaykumar, Dominika Rogozińska, Dustin Herbison, Elisa Bandy, Emma
  Wang, Eric Noland, Erica Moreira, Evan Senter, Evgenii Eltyshev, Francesco
  Visin, Gabriel Rasskin, Gary Wei, Glenn Cameron, Gus Martins, Hadi Hashemi,
  Hanna Klimczak-Plucińska, Harleen Batra, Harsh Dhand, Ivan Nardini, Jacinda
  Mein, Jack Zhou, James Svensson, Jeff Stanway, Jetha Chan, Jin~Peng Zhou,
  Joana Carrasqueira, Joana Iljazi, Jocelyn Becker, Joe Fernandez, Joost van
  Amersfoort, Josh Gordon, Josh Lipschultz, Josh Newlan, Ju~yeong Ji, Kareem
  Mohamed, Kartikeya Badola, Kat Black, Katie Millican, Keelin McDonell, Kelvin
  Nguyen, Kiranbir Sodhia, Kish Greene, Lars~Lowe Sjoesund, Lauren Usui,
  Laurent Sifre, Lena Heuermann, Leticia Lago, Lilly McNealus, Livio~Baldini
  Soares, Logan Kilpatrick, Lucas Dixon, Luciano Martins, Machel Reid,
  Manvinder Singh, Mark Iverson, Martin Görner, Mat Velloso, Mateo Wirth, Matt
  Davidow, Matt Miller, Matthew Rahtz, Matthew Watson, Meg Risdal, Mehran
  Kazemi, Michael Moynihan, Ming Zhang, Minsuk Kahng, Minwoo Park, Mofi Rahman,
  Mohit Khatwani, Natalie Dao, Nenshad Bardoliwalla, Nesh Devanathan, Neta
  Dumai, Nilay Chauhan, Oscar Wahltinez, Pankil Botarda, Parker Barnes, Paul
  Barham, Paul Michel, Pengchong Jin, Petko Georgiev, Phil Culliton, Pradeep
  Kuppala, Ramona Comanescu, Ramona Merhej, Reena Jana, Reza~Ardeshir Rokni,
  Rishabh Agarwal, Ryan Mullins, Samaneh Saadat, Sara~Mc Carthy, Sarah Cogan,
  Sarah Perrin, Sébastien M.~R. Arnold, Sebastian Krause, Shengyang Dai,
  Shruti Garg, Shruti Sheth, Sue Ronstrom, Susan Chan, Timothy Jordan, Ting Yu,
  Tom Eccles, Tom Hennigan, Tomas Kocisky, Tulsee Doshi, Vihan Jain, Vikas
  Yadav, Vilobh Meshram, Vishal Dharmadhikari, Warren Barkley, Wei Wei, Wenming
  Ye, Woohyun Han, Woosuk Kwon, Xiang Xu, Zhe Shen, Zhitao Gong, Zichuan Wei,
  Victor Cotruta, Phoebe Kirk, Anand Rao, Minh Giang, Ludovic Peran, Tris
  Warkentin, Eli Collins, Joelle Barral, Zoubin Ghahramani, Raia Hadsell,
  D.~Sculley, Jeanine Banks, Anca Dragan, Slav Petrov, Oriol Vinyals, Jeff
  Dean, Demis Hassabis, Koray Kavukcuoglu, Clement Farabet, Elena Buchatskaya,
  Sebastian Borgeaud, Noah Fiedel, Armand Joulin, Kathleen Kenealy, Robert
  Dadashi, and Alek Andreev.
\newblock Gemma 2: Improving open language models at a practical size,
  2024{\natexlab{a}}.
\newblock URL \url{https://arxiv.org/abs/2408.00118}.

\bibitem[Team et~al.(2024{\natexlab{b}})Team, Grattafiori, Dubey, Jauhri,
  Pandey, Kadian, Al-Dahle, Letman, Mathur, Schelten, Vaughan, Yang, Fan,
  Goyal, Hartshorn, Yang, Mitra, Sravankumar, Korenev, Hinsvark, Rao, Zhang,
  Rodriguez, Gregerson, Spataru, Roziere, Biron, Tang, Chern, Caucheteux,
  Nayak, Bi, Marra, McConnell, Keller, Touret, Wu, Wong, Ferrer, Nikolaidis,
  Allonsius, Song, Pintz, Livshits, Wyatt, Esiobu, Choudhary, Mahajan,
  Garcia-Olano, Perino, Hupkes, Lakomkin, AlBadawy, Lobanova, Dinan, Smith,
  Radenovic, Guzmán, Zhang, Synnaeve, Lee, Anderson, Thattai, Nail, Mialon,
  Pang, Cucurell, Nguyen, Korevaar, Xu, Touvron, Zarov, Ibarra, Kloumann,
  Misra, Evtimov, Zhang, Copet, Lee, Geffert, Vranes, Park, Mahadeokar, Shah,
  van~der Linde, Billock, Hong, Lee, Fu, Chi, Huang, Liu, Wang, Yu, Bitton,
  Spisak, Park, Rocca, Johnstun, Saxe, Jia, Alwala, Prasad, Upasani, Plawiak,
  Li, Heafield, Stone, El-Arini, Iyer, Malik, Chiu, Bhalla, Lakhotia,
  Rantala-Yeary, van~der Maaten, Chen, Tan, Jenkins, Martin, Madaan, Malo,
  Blecher, Landzaat, de~Oliveira, Muzzi, Pasupuleti, Singh, Paluri, Kardas,
  Tsimpoukelli, Oldham, Rita, Pavlova, Kambadur, Lewis, Si, Singh, Hassan,
  Goyal, Torabi, Bashlykov, Bogoychev, Chatterji, Zhang, Duchenne, Çelebi,
  Alrassy, Zhang, Li, Vasic, Weng, Bhargava, Dubal, Krishnan, Koura, Xu, He,
  Dong, Srinivasan, Ganapathy, Calderer, Cabral, Stojnic, Raileanu, Maheswari,
  Girdhar, Patel, Sauvestre, Polidoro, Sumbaly, Taylor, Silva, Hou, Wang,
  Hosseini, Chennabasappa, Singh, Bell, Kim, Edunov, Nie, Narang, Raparthy,
  Shen, Wan, Bhosale, Zhang, Vandenhende, Batra, Whitman, Sootla, Collot,
  Gururangan, Borodinsky, Herman, Fowler, Sheasha, Georgiou, Scialom,
  Speckbacher, Mihaylov, Xiao, Karn, Goswami, Gupta, Ramanathan, Kerkez,
  Gonguet, Do, Vogeti, Albiero, Petrovic, Chu, Xiong, Fu, Meers, Martinet,
  Wang, Wang, Tan, Xia, Xie, Jia, Wang, Goldschlag, Gaur, Babaei, Wen, Song,
  Zhang, Li, Mao, Coudert, Yan, Chen, Papakipos, Singh, Srivastava, Jain,
  Kelsey, Shajnfeld, Gangidi, Victoria, Goldstand, Menon, Sharma, Boesenberg,
  Baevski, Feinstein, Kallet, Sangani, Teo, Yunus, Lupu, Alvarado, Caples, Gu,
  Ho, Poulton, Ryan, Ramchandani, Dong, Franco, Goyal, Saraf, Chowdhury,
  Gabriel, Bharambe, Eisenman, Yazdan, James, Maurer, Leonhardi, Huang, Loyd,
  Paola, Paranjape, Liu, Wu, Ni, Hancock, Wasti, Spence, Stojkovic, Gamido,
  Montalvo, Parker, Burton, Mejia, Liu, Wang, Kim, Zhou, Hu, Chu, Cai, Tindal,
  Feichtenhofer, Gao, Civin, Beaty, Kreymer, Li, Adkins, Xu, Testuggine, David,
  Parikh, Liskovich, Foss, Wang, Le, Holland, Dowling, Jamil, Montgomery,
  Presani, Hahn, Wood, Le, Brinkman, Arcaute, Dunbar, Smothers, Sun, Kreuk,
  Tian, Kokkinos, Ozgenel, Caggioni, Kanayet, Seide, Florez, Schwarz, Badeer,
  Swee, Halpern, Herman, Sizov, Guangyi, Zhang, Lakshminarayanan, Inan,
  Shojanazeri, Zou, Wang, Zha, Habeeb, Rudolph, Suk, Aspegren, Goldman, Zhan,
  Damlaj, Molybog, Tufanov, Leontiadis, Veliche, Gat, Weissman, Geboski, Kohli,
  Lam, Asher, Gaya, Marcus, Tang, Chan, Zhen, Reizenstein, Teboul, Zhong, Jin,
  Yang, Cummings, Carvill, Shepard, McPhie, Torres, Ginsburg, Wang, Wu, U,
  Saxena, Khandelwal, Zand, Matosich, Veeraraghavan, Michelena, Li, Jagadeesh,
  Huang, Chawla, Huang, Chen, Garg, A, Silva, Bell, Zhang, Guo, Yu, Moshkovich,
  Wehrstedt, Khabsa, Avalani, Bhatt, Mankus, Hasson, Lennie, Reso, Groshev,
  Naumov, Lathi, Keneally, Liu, Seltzer, Valko, Restrepo, Patel, Vyatskov,
  Samvelyan, Clark, Macey, Wang, Hermoso, Metanat, Rastegari, Bansal,
  Santhanam, Parks, White, Bawa, Singhal, Egebo, Usunier, Mehta, Laptev, Dong,
  Cheng, Chernoguz, Hart, Salpekar, Kalinli, Kent, Parekh, Saab, Balaji,
  Rittner, Bontrager, Roux, Dollar, Zvyagina, Ratanchandani, Yuvraj, Liang,
  Alao, Rodriguez, Ayub, Murthy, Nayani, Mitra, Parthasarathy, Li, Hogan,
  Battey, Wang, Howes, Rinott, Mehta, Siby, Bondu, Datta, Chugh, Hunt, Dhillon,
  Sidorov, Pan, Mahajan, Verma, Yamamoto, Ramaswamy, Lindsay, Lindsay, Feng,
  Lin, Zha, Patil, Shankar, Zhang, Zhang, Wang, Agarwal, Sajuyigbe, Chintala,
  Max, Chen, Kehoe, Satterfield, Govindaprasad, Gupta, Deng, Cho, Virk,
  Subramanian, Choudhury, Goldman, Remez, Glaser, Best, Koehler, Robinson, Li,
  Zhang, Matthews, Chou, Shaked, Vontimitta, Ajayi, Montanez, Mohan, Kumar,
  Mangla, Ionescu, Poenaru, Mihailescu, Ivanov, Li, Wang, Jiang, Bouaziz,
  Constable, Tang, Wu, Wang, Wu, Gao, Kleinman, Chen, Hu, Jia, Qi, Li, Zhang,
  Zhang, Adi, Nam, Yu, Wang, Zhao, Hao, Qian, Li, He, Rait, DeVito, Rosnbrick,
  Wen, Yang, Zhao, and Ma]{grattafiori2024llama3herdmodels}
Llama Team, Aaron Grattafiori, Abhimanyu Dubey, Abhinav Jauhri, Abhinav Pandey,
  Abhishek Kadian, Ahmad Al-Dahle, Aiesha Letman, Akhil Mathur, Alan Schelten,
  Alex Vaughan, Amy Yang, Angela Fan, Anirudh Goyal, Anthony Hartshorn, Aobo
  Yang, Archi Mitra, Archie Sravankumar, Artem Korenev, Arthur Hinsvark, Arun
  Rao, Aston Zhang, Aurelien Rodriguez, Austen Gregerson, Ava Spataru, Baptiste
  Roziere, Bethany Biron, Binh Tang, Bobbie Chern, Charlotte Caucheteux, Chaya
  Nayak, Chloe Bi, Chris Marra, Chris McConnell, Christian Keller, Christophe
  Touret, Chunyang Wu, Corinne Wong, Cristian~Canton Ferrer, Cyrus Nikolaidis,
  Damien Allonsius, Daniel Song, Danielle Pintz, Danny Livshits, Danny Wyatt,
  David Esiobu, Dhruv Choudhary, Dhruv Mahajan, Diego Garcia-Olano, Diego
  Perino, Dieuwke Hupkes, Egor Lakomkin, Ehab AlBadawy, Elina Lobanova, Emily
  Dinan, Eric~Michael Smith, Filip Radenovic, Francisco Guzmán, Frank Zhang,
  Gabriel Synnaeve, Gabrielle Lee, Georgia~Lewis Anderson, Govind Thattai,
  Graeme Nail, Gregoire Mialon, Guan Pang, Guillem Cucurell, Hailey Nguyen,
  Hannah Korevaar, Hu~Xu, Hugo Touvron, Iliyan Zarov, Imanol~Arrieta Ibarra,
  Isabel Kloumann, Ishan Misra, Ivan Evtimov, Jack Zhang, Jade Copet, Jaewon
  Lee, Jan Geffert, Jana Vranes, Jason Park, Jay Mahadeokar, Jeet Shah, Jelmer
  van~der Linde, Jennifer Billock, Jenny Hong, Jenya Lee, Jeremy Fu, Jianfeng
  Chi, Jianyu Huang, Jiawen Liu, Jie Wang, Jiecao Yu, Joanna Bitton, Joe
  Spisak, Jongsoo Park, Joseph Rocca, Joshua Johnstun, Joshua Saxe, Junteng
  Jia, Kalyan~Vasuden Alwala, Karthik Prasad, Kartikeya Upasani, Kate Plawiak,
  Ke~Li, Kenneth Heafield, Kevin Stone, Khalid El-Arini, Krithika Iyer, Kshitiz
  Malik, Kuenley Chiu, Kunal Bhalla, Kushal Lakhotia, Lauren Rantala-Yeary,
  Laurens van~der Maaten, Lawrence Chen, Liang Tan, Liz Jenkins, Louis Martin,
  Lovish Madaan, Lubo Malo, Lukas Blecher, Lukas Landzaat, Luke de~Oliveira,
  Madeline Muzzi, Mahesh Pasupuleti, Mannat Singh, Manohar Paluri, Marcin
  Kardas, Maria Tsimpoukelli, Mathew Oldham, Mathieu Rita, Maya Pavlova,
  Melanie Kambadur, Mike Lewis, Min Si, Mitesh~Kumar Singh, Mona Hassan, Naman
  Goyal, Narjes Torabi, Nikolay Bashlykov, Nikolay Bogoychev, Niladri
  Chatterji, Ning Zhang, Olivier Duchenne, Onur Çelebi, Patrick Alrassy,
  Pengchuan Zhang, Pengwei Li, Petar Vasic, Peter Weng, Prajjwal Bhargava,
  Pratik Dubal, Praveen Krishnan, Punit~Singh Koura, Puxin Xu, Qing He,
  Qingxiao Dong, Ragavan Srinivasan, Raj Ganapathy, Ramon Calderer,
  Ricardo~Silveira Cabral, Robert Stojnic, Roberta Raileanu, Rohan Maheswari,
  Rohit Girdhar, Rohit Patel, Romain Sauvestre, Ronnie Polidoro, Roshan
  Sumbaly, Ross Taylor, Ruan Silva, Rui Hou, Rui Wang, Saghar Hosseini, Sahana
  Chennabasappa, Sanjay Singh, Sean Bell, Seohyun~Sonia Kim, Sergey Edunov,
  Shaoliang Nie, Sharan Narang, Sharath Raparthy, Sheng Shen, Shengye Wan,
  Shruti Bhosale, Shun Zhang, Simon Vandenhende, Soumya Batra, Spencer Whitman,
  Sten Sootla, Stephane Collot, Suchin Gururangan, Sydney Borodinsky, Tamar
  Herman, Tara Fowler, Tarek Sheasha, Thomas Georgiou, Thomas Scialom, Tobias
  Speckbacher, Todor Mihaylov, Tong Xiao, Ujjwal Karn, Vedanuj Goswami, Vibhor
  Gupta, Vignesh Ramanathan, Viktor Kerkez, Vincent Gonguet, Virginie Do, Vish
  Vogeti, Vítor Albiero, Vladan Petrovic, Weiwei Chu, Wenhan Xiong, Wenyin Fu,
  Whitney Meers, Xavier Martinet, Xiaodong Wang, Xiaofang Wang, Xiaoqing~Ellen
  Tan, Xide Xia, Xinfeng Xie, Xuchao Jia, Xuewei Wang, Yaelle Goldschlag,
  Yashesh Gaur, Yasmine Babaei, Yi~Wen, Yiwen Song, Yuchen Zhang, Yue Li,
  Yuning Mao, Zacharie~Delpierre Coudert, Zheng Yan, Zhengxing Chen, Zoe
  Papakipos, Aaditya Singh, Aayushi Srivastava, Abha Jain, Adam Kelsey, Adam
  Shajnfeld, Adithya Gangidi, Adolfo Victoria, Ahuva Goldstand, Ajay Menon,
  Ajay Sharma, Alex Boesenberg, Alexei Baevski, Allie Feinstein, Amanda Kallet,
  Amit Sangani, Amos Teo, Anam Yunus, Andrei Lupu, Andres Alvarado, Andrew
  Caples, Andrew Gu, Andrew Ho, Andrew Poulton, Andrew Ryan, Ankit Ramchandani,
  Annie Dong, Annie Franco, Anuj Goyal, Aparajita Saraf, Arkabandhu Chowdhury,
  Ashley Gabriel, Ashwin Bharambe, Assaf Eisenman, Azadeh Yazdan, Beau James,
  Ben Maurer, Benjamin Leonhardi, Bernie Huang, Beth Loyd, Beto~De Paola,
  Bhargavi Paranjape, Bing Liu, Bo~Wu, Boyu Ni, Braden Hancock, Bram Wasti,
  Brandon Spence, Brani Stojkovic, Brian Gamido, Britt Montalvo, Carl Parker,
  Carly Burton, Catalina Mejia, Ce~Liu, Changhan Wang, Changkyu Kim, Chao Zhou,
  Chester Hu, Ching-Hsiang Chu, Chris Cai, Chris Tindal, Christoph
  Feichtenhofer, Cynthia Gao, Damon Civin, Dana Beaty, Daniel Kreymer, Daniel
  Li, David Adkins, David Xu, Davide Testuggine, Delia David, Devi Parikh,
  Diana Liskovich, Didem Foss, Dingkang Wang, Duc Le, Dustin Holland, Edward
  Dowling, Eissa Jamil, Elaine Montgomery, Eleonora Presani, Emily Hahn, Emily
  Wood, Eric-Tuan Le, Erik Brinkman, Esteban Arcaute, Evan Dunbar, Evan
  Smothers, Fei Sun, Felix Kreuk, Feng Tian, Filippos Kokkinos, Firat Ozgenel,
  Francesco Caggioni, Frank Kanayet, Frank Seide, Gabriela~Medina Florez,
  Gabriella Schwarz, Gada Badeer, Georgia Swee, Gil Halpern, Grant Herman,
  Grigory Sizov, Guangyi, Zhang, Guna Lakshminarayanan, Hakan Inan, Hamid
  Shojanazeri, Han Zou, Hannah Wang, Hanwen Zha, Haroun Habeeb, Harrison
  Rudolph, Helen Suk, Henry Aspegren, Hunter Goldman, Hongyuan Zhan, Ibrahim
  Damlaj, Igor Molybog, Igor Tufanov, Ilias Leontiadis, Irina-Elena Veliche,
  Itai Gat, Jake Weissman, James Geboski, James Kohli, Janice Lam, Japhet
  Asher, Jean-Baptiste Gaya, Jeff Marcus, Jeff Tang, Jennifer Chan, Jenny Zhen,
  Jeremy Reizenstein, Jeremy Teboul, Jessica Zhong, Jian Jin, Jingyi Yang, Joe
  Cummings, Jon Carvill, Jon Shepard, Jonathan McPhie, Jonathan Torres, Josh
  Ginsburg, Junjie Wang, Kai Wu, Kam~Hou U, Karan Saxena, Kartikay Khandelwal,
  Katayoun Zand, Kathy Matosich, Kaushik Veeraraghavan, Kelly Michelena, Keqian
  Li, Kiran Jagadeesh, Kun Huang, Kunal Chawla, Kyle Huang, Lailin Chen,
  Lakshya Garg, Lavender A, Leandro Silva, Lee Bell, Lei Zhang, Liangpeng Guo,
  Licheng Yu, Liron Moshkovich, Luca Wehrstedt, Madian Khabsa, Manav Avalani,
  Manish Bhatt, Martynas Mankus, Matan Hasson, Matthew Lennie, Matthias Reso,
  Maxim Groshev, Maxim Naumov, Maya Lathi, Meghan Keneally, Miao Liu,
  Michael~L. Seltzer, Michal Valko, Michelle Restrepo, Mihir Patel, Mik
  Vyatskov, Mikayel Samvelyan, Mike Clark, Mike Macey, Mike Wang, Miquel~Jubert
  Hermoso, Mo~Metanat, Mohammad Rastegari, Munish Bansal, Nandhini Santhanam,
  Natascha Parks, Natasha White, Navyata Bawa, Nayan Singhal, Nick Egebo,
  Nicolas Usunier, Nikhil Mehta, Nikolay~Pavlovich Laptev, Ning Dong, Norman
  Cheng, Oleg Chernoguz, Olivia Hart, Omkar Salpekar, Ozlem Kalinli, Parkin
  Kent, Parth Parekh, Paul Saab, Pavan Balaji, Pedro Rittner, Philip Bontrager,
  Pierre Roux, Piotr Dollar, Polina Zvyagina, Prashant Ratanchandani, Pritish
  Yuvraj, Qian Liang, Rachad Alao, Rachel Rodriguez, Rafi Ayub, Raghotham
  Murthy, Raghu Nayani, Rahul Mitra, Rangaprabhu Parthasarathy, Raymond Li,
  Rebekkah Hogan, Robin Battey, Rocky Wang, Russ Howes, Ruty Rinott, Sachin
  Mehta, Sachin Siby, Sai~Jayesh Bondu, Samyak Datta, Sara Chugh, Sara Hunt,
  Sargun Dhillon, Sasha Sidorov, Satadru Pan, Saurabh Mahajan, Saurabh Verma,
  Seiji Yamamoto, Sharadh Ramaswamy, Shaun Lindsay, Shaun Lindsay, Sheng Feng,
  Shenghao Lin, Shengxin~Cindy Zha, Shishir Patil, Shiva Shankar, Shuqiang
  Zhang, Shuqiang Zhang, Sinong Wang, Sneha Agarwal, Soji Sajuyigbe, Soumith
  Chintala, Stephanie Max, Stephen Chen, Steve Kehoe, Steve Satterfield,
  Sudarshan Govindaprasad, Sumit Gupta, Summer Deng, Sungmin Cho, Sunny Virk,
  Suraj Subramanian, Sy~Choudhury, Sydney Goldman, Tal Remez, Tamar Glaser,
  Tamara Best, Thilo Koehler, Thomas Robinson, Tianhe Li, Tianjun Zhang, Tim
  Matthews, Timothy Chou, Tzook Shaked, Varun Vontimitta, Victoria Ajayi,
  Victoria Montanez, Vijai Mohan, Vinay~Satish Kumar, Vishal Mangla, Vlad
  Ionescu, Vlad Poenaru, Vlad~Tiberiu Mihailescu, Vladimir Ivanov, Wei Li,
  Wenchen Wang, Wenwen Jiang, Wes Bouaziz, Will Constable, Xiaocheng Tang,
  Xiaojian Wu, Xiaolan Wang, Xilun Wu, Xinbo Gao, Yaniv Kleinman, Yanjun Chen,
  Ye~Hu, Ye~Jia, Ye~Qi, Yenda Li, Yilin Zhang, Ying Zhang, Yossi Adi, Youngjin
  Nam, Yu, Wang, Yu~Zhao, Yuchen Hao, Yundi Qian, Yunlu Li, Yuzi He, Zach Rait,
  Zachary DeVito, Zef Rosnbrick, Zhaoduo Wen, Zhenyu Yang, Zhiwei Zhao, and
  Zhiyu Ma.
\newblock The llama 3 herd of models, 2024{\natexlab{b}}.
\newblock URL \url{https://arxiv.org/abs/2407.21783}.

\bibitem[Tenney et~al.(2019)Tenney, Das, and Pavlick]{tenney-etal-2019-bert}
Ian Tenney, Dipanjan Das, and Ellie Pavlick.
\newblock {BERT} rediscovers the classical {NLP} pipeline.
\newblock In Anna Korhonen, David Traum, and Llu{\'i}s M{\`a}rquez, editors,
  \emph{Proceedings of the 57th Annual Meeting of the Association for
  Computational Linguistics}, pages 4593--4601, Florence, Italy, July 2019.
  Association for Computational Linguistics.
\newblock \doi{10.18653/v1/P19-1452}.
\newblock URL \url{https://aclanthology.org/P19-1452/}.

\bibitem[Turner et~al.(2023)Turner, Thiergart, Udell, Leech, Mini, and
  MacDiarmid]{DBLP:journals/corr/abs-2308-10248}
Alexander~Matt Turner, Lisa Thiergart, David Udell, Gavin Leech, Ulisse Mini,
  and Monte MacDiarmid.
\newblock Activation addition: Steering language models without optimization.
\newblock \emph{CoRR}, abs/2308.10248, 2023.
\newblock URL \url{https://doi.org/10.48550/arXiv.2308.10248}.

\bibitem[Wang et~al.(2023)Wang, Variengien, Conmy, Shlegeris, and
  Steinhardt]{wang2023interpretability}
Kevin~Ro Wang, Alexandre Variengien, Arthur Conmy, Buck Shlegeris, and Jacob
  Steinhardt.
\newblock Interpretability in the wild: a circuit for indirect object
  identification in {GPT}-2 small.
\newblock In \emph{The Eleventh International Conference on Learning
  Representations}, 2023.
\newblock URL \url{https://openreview.net/forum?id=NpsVSN6o4ul}.

\bibitem[Wang et~al.(2024)Wang, Ma, Zhang, Ni, Chandra, Guo, Ren, Arulraj, He,
  Jiang, Li, Ku, Wang, Zhuang, Fan, Yue, and Chen]{wang2024mmlupro}
Yubo Wang, Xueguang Ma, Ge~Zhang, Yuansheng Ni, Abhranil Chandra, Shiguang Guo,
  Weiming Ren, Aaran Arulraj, Xuan He, Ziyan Jiang, Tianle Li, Max Ku, Kai
  Wang, Alex Zhuang, Rongqi Fan, Xiang Yue, and Wenhu Chen.
\newblock {MMLU}-pro: A more robust and challenging multi-task language
  understanding benchmark.
\newblock In \emph{The Thirty-eight Conference on Neural Information Processing
  Systems Datasets and Benchmarks Track}, 2024.
\newblock URL \url{https://openreview.net/forum?id=y10DM6R2r3}.

\bibitem[Wei et~al.(2024)Wei, Karina, Chung, Jiao, Papay, Glaese, Schulman, and
  Fedus]{wei2024measuringshortformfactualitylarge}
Jason Wei, Nguyen Karina, Hyung~Won Chung, Yunxin~Joy Jiao, Spencer Papay,
  Amelia Glaese, John Schulman, and William Fedus.
\newblock Measuring short-form factuality in large language models, 2024.
\newblock URL \url{https://openai.com/index/introducing-simpleqa/}.

\bibitem[Wu et~al.(2025)Wu, Arora, Geiger, Wang, Huang, Jurafsky, Manning, and
  Potts]{wu2025axbench}
Zhengxuan Wu, Aryaman Arora, Atticus Geiger, Zheng Wang, Jing Huang, Dan
  Jurafsky, Christopher~D Manning, and Christopher Potts.
\newblock {A}x{B}ench: Steering {LLM}s? {E}ven simple baselines outperform
  sparse autoencoders.
\newblock In \emph{Proceedings of the 42nd International Conference on Machine
  Learning}, volume 267 of \emph{Proceedings of Machine Learning Research},
  pages 67035--67080. PMLR, 2025.
\newblock URL \url{https://proceedings.mlr.press/v267/wu25a.html}.

\bibitem[Xu and Jin(2024)]{xu2024dynamic}
Wei Xu and Xiaodong Jin.
\newblock Dynamic layer skipping for large language models on natural language
  understanding tasks and machine translation using reinforcement learning.
\newblock \emph{Frontiers in Computing and Intelligent Systems}, 9\penalty0
  (3), 2024.
\newblock \doi{10.54097/wy0g8m89}.
\newblock Uses dynamic layer skipping with adapters and RL to accelerate
  inference of LLMs; achieves SOTA performance on NLU and MT tasks.

\bibitem[Ye et~al.(2021)Ye, Lin, Huang, and Sun]{ye-etal-2021-tr}
Deming Ye, Yankai Lin, Yufei Huang, and Maosong Sun.
\newblock Tr-bert: Dynamic token reduction for accelerating bert inference.
\newblock In Kristina Toutanova, Anna Rumshisky, Luke Zettlemoyer, Dilek
  Hakkani-Tur, Iz~Beltagy, Steven Bethard, Ryan Cotterell, Tanmoy Chakraborty,
  and Yichao Zhou, editors, \emph{Proceedings of the 2021 Conference of the
  North American Chapter of the Association for Computational Linguistics:
  Human Language Technologies}, pages 5798--5809, Online, June 2021.
  Association for Computational Linguistics.
\newblock \doi{10.18653/v1/2021.naacl-main.463}.
\newblock URL \url{https://aclanthology.org/2021.naacl-main.463/}.

\bibitem[Yu et~al.(2017)Yu, Lee, and Le]{yu-etal-2017-learning}
Adams~Wei Yu, Hongrae Lee, and Quoc Le.
\newblock Learning to skim text.
\newblock In Regina Barzilay and Min-Yen Kan, editors, \emph{Proceedings of the
  55th Annual Meeting of the Association for Computational Linguistics (Volume
  1: Long Papers)}, pages 1880--1890, Vancouver, Canada, July 2017. Association
  for Computational Linguistics.
\newblock \doi{10.18653/v1/P17-1172}.
\newblock URL \url{https://aclanthology.org/P17-1172/}.

\bibitem[Zhong and Zhang(2023)]{zhong2023a}
Han Zhong and Tong Zhang.
\newblock A theoretical analysis of optimistic proximal policy optimization in
  linear markov decision processes.
\newblock In \emph{Thirty-seventh Conference on Neural Information Processing
  Systems}, 2023.
\newblock URL \url{https://openreview.net/forum?id=1bTG4sJ7tN}.

\end{thebibliography}

\appendix
\newpage

\section{Appendix}
\label{sec:appendix}

\subsection{Key Terminology and Implementation Details}
\label{subsec:terminology}

\begin{itemize}[leftmargin=*,itemsep=3pt,topsep=0pt,parsep=0pt,partopsep=0pt]
\item \textbf{Corrected samples}: Samples where CRL steering changes incorrect predictions to correct ones
\item \textbf{Misguided samples}: Samples where CRL steering changes correct predictions to incorrect ones
\item \textbf{Policy network $\pi_\theta$}: 2-layer MLP (Tanh activation) that maps residual stream activations to probabilities of selecting SAE features (hidden dimension matching LLM)
\item \textbf{Critic network $V_\phi$}: 2-layer MLP (Tanh activation) that estimates state value function for PPO training (hidden dimension matching LLM)
\item \textbf{Steering coefficient $c$}: Scalar multiplier controlling intervention magnitude in $\tilde{x}_t = x_t + c \cdot a_t W_{dec}$
\item \textbf{Constrained decoding}: Generation restricted to valid answer tokens (e.g., A/B/C/D for MMLU)
\item \textbf{Unconstrained decoding}: Generation allowing any token output, including invalid responses
\end{itemize}

\subsection{Algorithm 1: CRL Training Procedure}

Training uses learning rate 1e-5, PPO clip parameter $\epsilon = 0.2$, and gradient clipping at norm 1.0.
Our training protocol follows these steps: generate steered responses, compute task-specific rewards per sequence, aggregate them into per-token PPO losses, and update the policy and critic networks by a backward pass.
After training, we evaluate the policy network on the test set and analyze critic values and the features used contributing to altered generations.

\begin{algorithm}[h]
\caption{CRL Training Procedure}
\label{alg:crl_training}
\begin{algorithmic}[1]
\REQUIRE Base model $M$, SAE ($W_{enc}$, $W_{dec}$), task dataset $D$
\ENSURE Trained policy $\pi$, value function $V$
\STATE Initialize policy $\pi(a|s)$ and value function $V(s)$
\FOR{each episode}
    \STATE Sample batch $B$ from dataset $D$
    \FOR{each token position $t$}
        \STATE Extract residual activation $x_t$
        \STATE Select feature $a_t \sim \pi(\cdot|x_t)$
        \STATE Apply intervention: $\tilde{x}_t = x_t + c \cdot a_t \cdot W_{dec}$
        \STATE Compute reward $r_t$ from task outcome
    \ENDFOR
    \STATE Update $\pi$ and $V$ using PPO with rewards $\{r_t\}$
\ENDFOR
\RETURN trained $\pi$, $V$
\end{algorithmic}
\end{algorithm}

\subsection{Task-Specific Rewards}
\label{subsec:task_rewards}

We use the following reward structures for each task:

\textbf{Benchmarks with ground-truth answers:}
\begin{itemize}
\item \textbf{MMLU \& MMLU-Pro:} Binary reward: +1 for correct multiple-choice answer, 0 for incorrect
\item \textbf{GSM8K:} Binary reward: +1 for exact numerical match, 0 otherwise  
\item \textbf{BBQ (Ambig \& Disambig):} Binary reward: +1 for unbiased answer, 0 for biased answer
\end{itemize}

\textbf{Benchmarks requiring evaluation models:}
\begin{itemize}
\item \textbf{HarmBench \& XSTest:} Binary reward: +1 for refusing the harmful request, 0 for compliance, scored by a RoBERTa-based refusal classifier\footnote{\url{https://huggingface.co/protectai/distilroberta-base-rejection-v1}} (inverted for XSTest over-refusal).
\item \textbf{SimpleQA:} Binary reward based on semantic similarity. The BERT STS model\footnote{\url{https://huggingface.co/datasets/HuggingFaceH4/stsb_multi_mt}} matches generated answers against expected responses.
\end{itemize}

\subsection{Experimental Setup}

\textbf{Models and SAEs.} We conduct experiments using Gemma 2 2B-IT~\citep{gemmateam2024gemma2improvingopen} with pre-trained SAEs from Gemma Scope~\citep{lieberum-etal-2024-gemma} (16K features) across layers 1-26, and LLaMA-3.1 8B~\citep{grattafiori2024llama3herdmodels} with LLaMA Scope SAEs~\citep{he2024llamascopeextractingmillions} (32K features).
Both SAE families employ JumpReLU activation~\citep{rajamanoharan2024jumpingaheadimprovingreconstruction} and were the only releases providing SAEs across all transformer layers at the time of writing.
SAEs are transferable across fine-tuned models with low reconstruction loss.
\textbf{Training Protocol.} Training uses PPO with batch size 8, evaluating on 500 validation samples every 100 training steps.
Selection tasks use 1 token generation, while HarmBench/XSTest use 32 tokens and GSM8K uses 1024 tokens.
For datasets with fewer than 4000 samples, we repeat training data to reach 4000 samples.
Layer selection for \hyperref[tab:gemma_results]{\textcolor{linkblue}{Table~\ref*{tab:gemma_results}}} uses a held-out validation set to select optimal layers and coefficients; final evaluation is performed on the separate test set to ensure no data leakage.
\textbf{Cross-seed Stability.} Results are reported as mean $\pm$ std across three seeds. Standard deviations remain low, with CRL notably reducing HarmBench variance from $\pm$9.05 to $\pm$1.59, indicating stable feature selection.

\textbf{Evaluation Benchmarks.} We evaluate CRL across five categories:
\textbf{Knowledge}: MMLU~\citep{hendrycks2021measuring}, MMLU-Pro~\citep{wang2024mmlupro}.
\textbf{Reasoning}: GSM8K~\citep{cobbe2021trainingverifierssolvemath}.
\textbf{Bias}: BBQ~\citep{parrish-etal-2022-bbq}.
\textbf{Factuality}: SimpleQA~\citep{wei2024measuringshortformfactualitylarge}.
\textbf{Safety}: HarmBench~\citep{mazeika2024harmbench}, XSTest~\citep{rottger-etal-2024-xstest}.

\subsection{CRL-Layer: Layer-wise Steering Mechanism}
\label{subsec:layer_wise_steering}

For layer-wise steering, we extend the framework to operate across multiple transformer layers simultaneously using a single token position.
The shared Markov Decision Process operates on layer-specific variables where all components are parameterized with layer index $\ell$.

Let $\mathbf{x}^{(\ell)} \in \mathbb{R}^{d}$ denote the residual stream activation at layer $\ell$ for a single target token position. Given layer-specific SAE components with encoder $\mathbf{W}_{enc}^{(\ell)} \in \mathbb{R}^{d \times d_{dict}}$ and decoder $\mathbf{W}_{dec}^{(\ell)} \in \mathbb{R}^{d_{dict} \times d}$, the steering mechanism applies perturbations across all layers:

\begin{equation}
\tilde{\mathbf{x}}^{(\ell)} = \mathbf{x}^{(\ell)} + \mathbf{a}^{(\ell)} \mathbf{W}_{dec}^{(\ell)}
\end{equation}

where $\mathbf{a}^{(\ell)} \in \{0,1\}^{d_{dict}}$ is the layer-specific action vector. The shared MDP coordinates decisions across layers through a joint policy:

\begin{equation}
\pi_\theta(\mathbf{a}^{(1:L)} | \mathbf{x}^{(1:L)}) = \prod_{\ell=1}^{L} \pi_\theta^{(\ell)}(\mathbf{a}^{(\ell)} | \mathbf{x}^{(\ell)})
\end{equation}

where $L$ is the total number of layers and each layer-specific policy $\pi_\theta^{(\ell)}$ shares parameters while adapting to layer-specific representations.

\textbf{CRL-Layer Results.}

CRL-Layer's practical advantage lies in sharing policy and critic networks across layers.
This approach reduces computational resources compared to extending CRL-Token to multiple layers without cross-layer compatibility.
Although representation spaces for each layer's residual stream differ, residual connections enforce shared vector spaces, and existing layer-wise reuse approaches~\citep{ye-etal-2021-tr, elhoushi2024, raposo2024mixtureofdepths} support network sharing across layers.

\begin{table}[ht!]
  \centering
  \small
  \caption{Performance comparison between CRL variants on Gemma 2 2B. Results use single-seed 0-shot evaluation with layer-shared policy; BBQ and HarmBench baselines differ from Table~\ref{tab:gemma_results} due to 0-shot vs 1-shot evaluation and single-seed variance.}
  \label{tab:method_comparison_crl}
  \begin{tabular}{lcccccc}
  \toprule
  \textbf{Method} & \textbf{MMLU} & \textbf{MMLU-Pro} & \textbf{BBQ Ambig} & \textbf{BBQ Disambig} & \textbf{XSTest} & \textbf{HarmBench} \\
  \midrule
  Baseline & 52.23 & 30.30 & 59.10 & 75.42 & 86.35 & 44.64 \\
  CRL-Layer & 55.00 & 29.18 & 62.93 & 75.68 & 86.98 & 71.43 \\
  CRL-Token & 55.55 & 30.44 & 61.88 & 76.73 & 86.98 & 50.25 \\
  \bottomrule
  \end{tabular}
\end{table}

\hyperref[tab:method_comparison_crl]{\textcolor{linkblue}{Table~\ref*{tab:method_comparison_crl}}} presents a comparison between CRL-Layer and CRL-Token across multiple benchmarks.
CRL-Layer achieves improvements over baseline across most tasks except MMLU-Pro, though it generally underperforms CRL-Token on complex tasks, with the exception of refusal tasks.
This pattern reveals that while layer-wise sharing provides clear computational advantages, it introduces limitations for complex tasks requiring layer-specific weights.
Notably, CRL-Layer achieves the highest performance on HarmBench, outperforming CRL-Token, suggesting that layer-shared refusal patterns benefit from cross-layer feature consistency.

\subsection{LLaMA Results}
\label{subsec:llama_results}

We extend our evaluation to LLaMA-3.1 8B model with LLaMA Scope SAEs to demonstrate the generalizability of our approach across different model architectures.

\begin{table}[ht!]
  \centering
  \small
  \caption{Performance results for LLaMA-3.1 8B model across different tasks using CRL-Token approach. MMLU and MMLU-Pro use 0-shot evaluation, BBQ tasks use 1-shot evaluation, SimpleQA uses 0-shot evaluation, XSTest and HarmBench use refusal rate evaluation.}
  \label{tab:llama_results}
  \begin{tabular}{lccccc}
  \toprule
  \textbf{Task} & \textbf{Type} & \textbf{Layer} & \textbf{Before} & \textbf{After} & \textbf{Gain} \\
  \midrule
  MMLU & knowledge & 20 & 61.41 & 62.33 & +0.92 \\
  MMLU-Pro & knowledge & 26 & 32.13 & 33.16 & +1.03 \\
  BBQ Ambig & bias & 7 & 83.97 & 85.04 & +1.07 \\
  BBQ Disambig & bias & 30 & 90.07 & 90.16 & +0.09 \\
  SimpleQA & factuality & 25 & 0.43 & 0.96 & +0.53 \\
  XSTest & safety & 23 & 61.27 & 59.37 & -1.90 \\
  HarmBench & safety & 5 & 0.71 & 6.07 & +5.36 \\
  \bottomrule
  \end{tabular}
\end{table}

\hyperref[tab:llama_results]{\textcolor{linkblue}{Table~\ref*{tab:llama_results}}} demonstrates that CRL-Token generalizes across different model families, with mixed results across tasks.
While most tasks show improvements (knowledge, bias, factuality), XSTest exhibits a decline (-1.90), suggesting task-specific sensitivity to the steering approach.
HarmBench shows improvement (+5.36), consistent with the Gemma 2 2B results, demonstrating safety steering across architectures.
The mixed results confirm that our approach transfers across different model architectures while highlighting the importance of task-specific optimization, maintaining the interpretable feature-based control mechanism.
\subsection{Critic Value Analysis}
\label{subsec:critic_value_analysis}

Single-token (MMLU, BBQ) critic visualizations are in \hyperref[fig:critic_single_token]{\textcolor{linkblue}{Figure~\ref*{fig:critic_single_token}}} and the GSM8K multi-token trajectory in \hyperref[fig:gsm8k_critic]{\textcolor{linkblue}{Figure~\ref*{fig:gsm8k_critic}}} (main text).
This appendix focuses on multi-token critic patterns for HarmBench and XSTest.

The four sample categories (correct/incorrect, corrected/misguided) are analyzed for HarmBench and XSTest tasks to understand critic behavior patterns in multi-token generation scenarios.

\begin{figure}[h]
  \centering
  \includegraphics[width=1\textwidth]{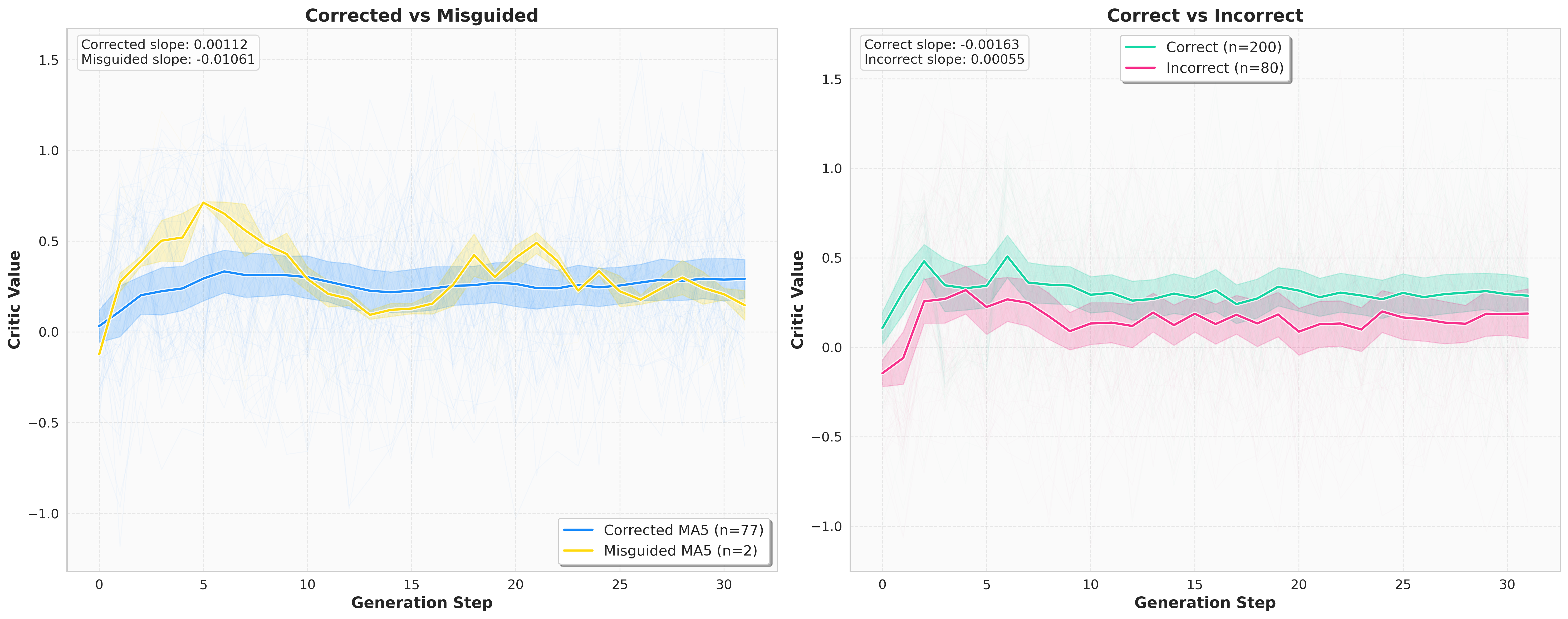}
  \caption{Critic value trajectories along token generation for HarmBench task. The critic successfully distinguishes correct from incorrect samples throughout generation.}
  \label{fig:harmbench_critic}
\end{figure}

Our analysis focuses on linear regression gradient slopes and the visible gaps between sample categories.
Empirically, we observed that critic value gaps between correct and incorrect samples become more pronounced in later layers.
We observe two distinct patterns in critic trajectories analogous to those in single-token generation tasks.
These patterns reflect either errors in critic value estimation (manifesting as bias gaps) or limitations in the policy's critic value utility (affecting gradient dynamics).

\begin{figure}[h]
  \centering
  \includegraphics[width=1\textwidth]{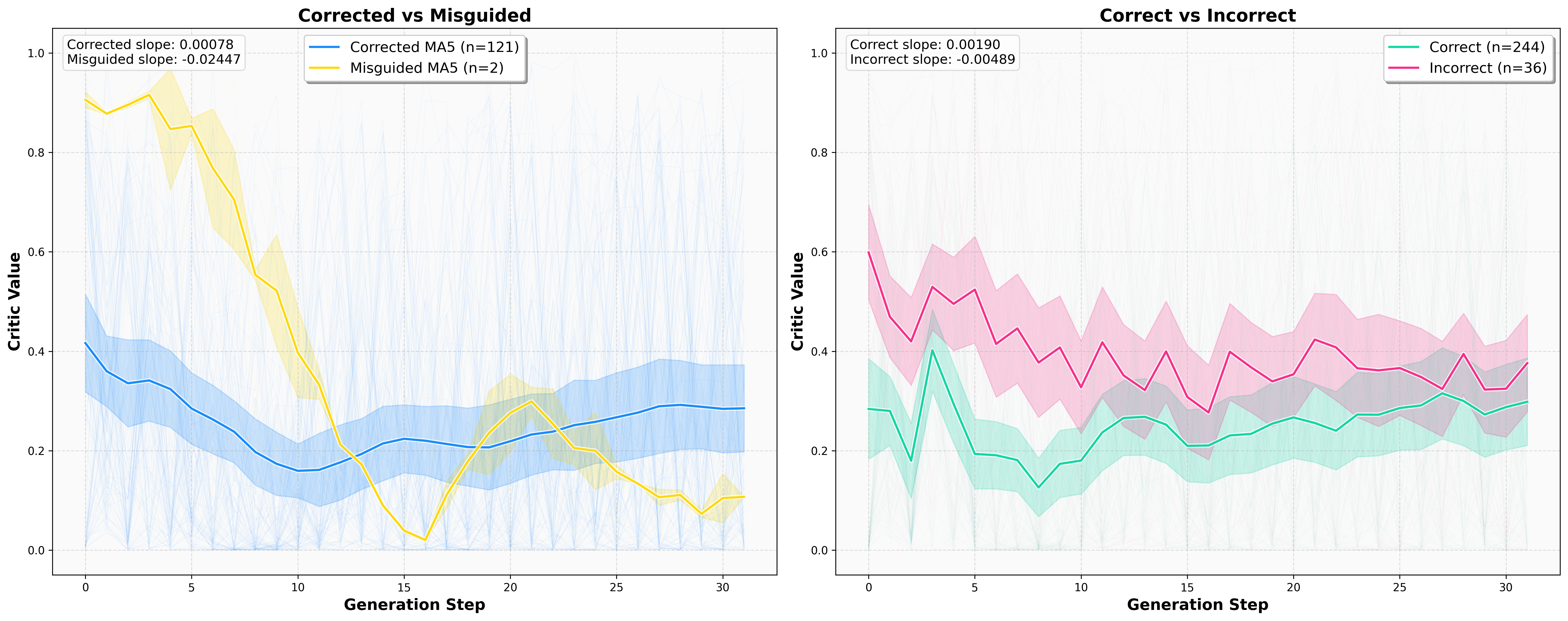}
  \caption{Critic value trajectories along token generation for XSTest task. XSTest shows inverted estimation patterns compared to HarmBench, indicating critic bottleneck for context-dependent benign requests.}
  \label{fig:xstest_critic}
\end{figure}

For HarmBench (\hyperref[fig:harmbench_critic]{\textcolor{linkblue}{Figure~\ref*{fig:harmbench_critic}}}), the critic successfully distinguishes between correct and incorrect samples, whereas XSTest (\hyperref[fig:xstest_critic]{\textcolor{linkblue}{Figure~\ref*{fig:xstest_critic}}}) exhibits inverted estimation patterns.
However, while HarmBench shows clear gaps at sequence boundaries with minimal gradient differences, XSTest demonstrates estimation errors in gap measurements but maintains clear superiority for correct samples in slope analysis across both correct/incorrect and corrected/misguided categories.

\subsection{Hallucination Mitigation Analysis}

\begin{figure}[ht]
  \centering
  \includegraphics[width=0.6\textwidth]{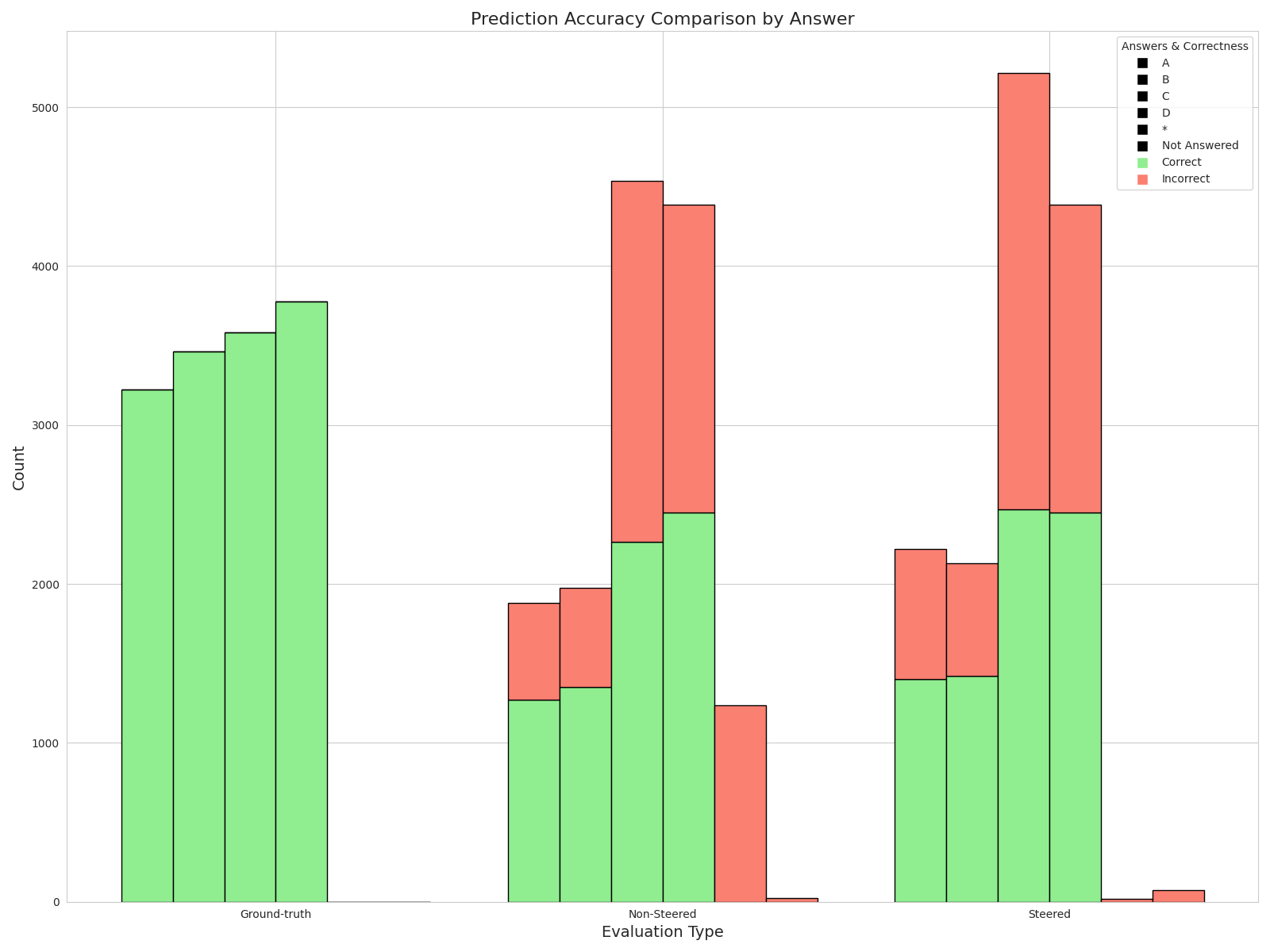}
  \caption{Comparison of invalid responses on MMLU. Baseline generates responses like ``*'' or whitespace instead of valid options (A, B, C, D). CRL-Token steering eliminates these invalid outputs, contributing to accuracy gains beyond format correction.}
  \label{fig:hallucination_answers}
\end{figure}

For single-token generation tasks, particularly MMLU without constrained decoding, a substantial portion of performance improvement stems from hallucination mitigation.
Steering with CRL-Token effectively eliminates responses that fall outside the provided answer options (A, B, C, D).
The baseline model frequently generates invalid responses such as "*" or whitespace instead of selecting from the given options.
This behavior strongly affects performance metrics, as these responses are automatically marked incorrect regardless of the underlying reasoning quality.
Our CRL-Token approach addresses this issue by learning to constrain outputs to valid answer choices while maintaining the model's reasoning capabilities.
The analysis reveals that a portion of performance gains in MMLU tasks can be attributed to this hallucination mitigation effect, with the remainder coming from improved feature selection and reasoning enhancement.
This finding highlights the importance of output format consistency in evaluation benchmarks and demonstrates the effectiveness of CRL in learning task-specific constraints.

\subsection{Feature Selection Evaluation}
\label{subsec:feature_selection_analysis}

We analyze the policy network's feature selection patterns to understand which SAE features contribute most to performance improvements.
For each feature $i$, we compute:

\subsubsection{Impact Score}

\begin{align}
n_i &= \sum_{t} \mathbf{1}\{i \in S_t\}, \quad N = \sum_{j} n_j, \\
p_i &= \frac{n_i}{N}, \\
c_i &= \sum_{t} \mathbf{1}\{i \in S_t\}\, \mathbf{1}\{\text{improved}_t\}, \quad m_i = \sum_{t} \mathbf{1}\{i \in S_t\}\, \mathbf{1}\{\text{degraded}_t\}, \\
\text{Impact}_i &= \frac{(c_i + m_i) \cdot \log(n_i + \varepsilon)}{n_i \cdot \log(n_i + \varepsilon)} = \frac{c_i + m_i}{n_i}, \quad (\varepsilon > 0)
\end{align}
where $S_t$ is the set of selected features at step $t$ (e.g., top-$k$), and $\text{improved}_t$/$\text{degraded}_t$ denote whether steering improved/degraded the outcome versus baseline.

\subsubsection{Feature Diversity}
Additionally, we compute feature diversity to understand the policy's exploration behavior:

\begin{equation}
\text{Feature Diversity} = -\sum_{i=1}^{d_{dict}} p_i \log p_i
\end{equation}

where $p_i$ is the probability of selecting feature $i$.
Higher entropy indicates more diverse feature usage across the SAE dictionary.
Together, these metrics quantify how diverse the trained policy's feature selection is and how much the selected features actually influence behavior change.

\subsection{Feature Diversity and Impact Score}
\label{subsec:feature_diversity_and_impact_score}

\begin{figure}[H]
  \centering
  \includegraphics[width=0.325\textwidth]{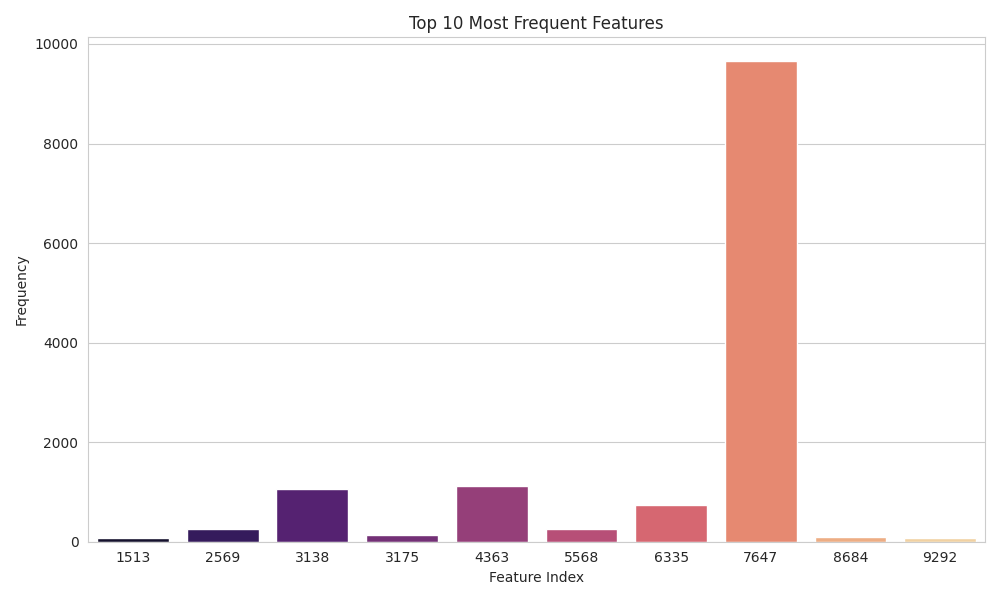}
  \includegraphics[width=0.325\textwidth]{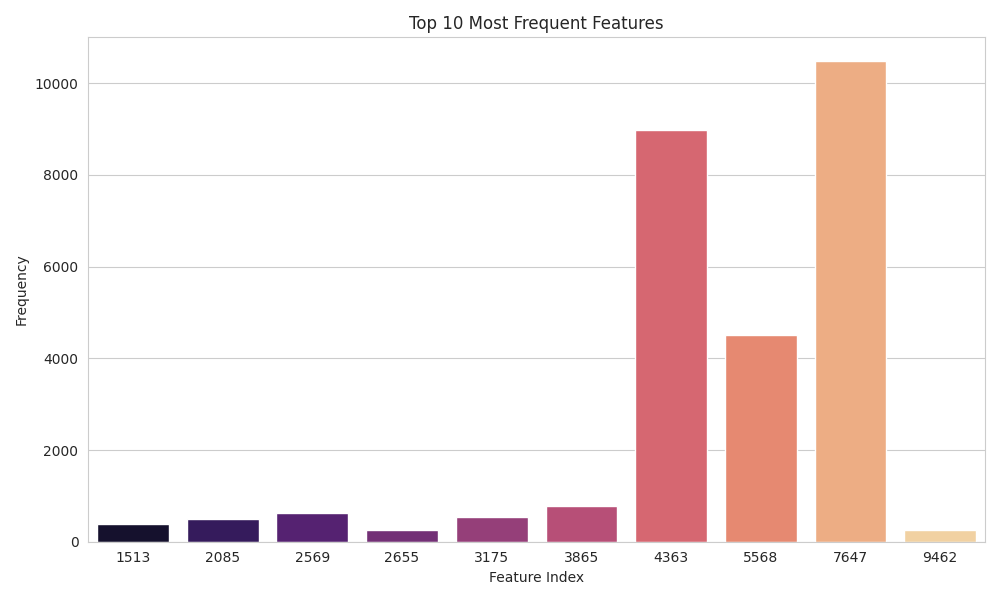}
  \includegraphics[width=0.325\textwidth]{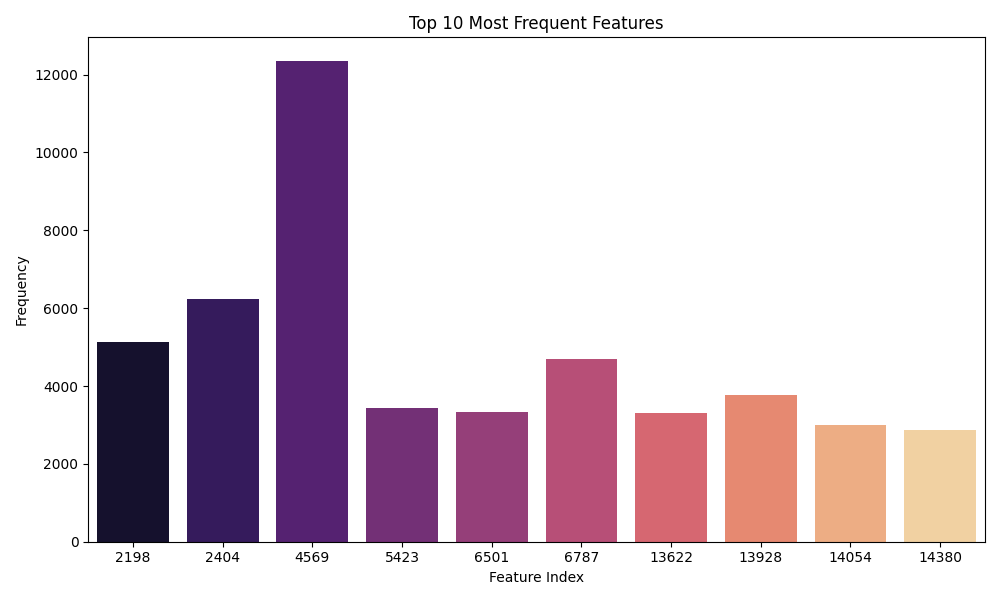}
  \caption{Top activated features for MMLU (left) and BBQ Disambig (center) and GSM8K (right) tasks, revealing semantic patterns in feature selection for reasoning and bias mitigation.}
  \label{fig:top_features_comparison}
  \end{figure}

\begin{figure}[t]
\centering
\includegraphics[width=0.85\textwidth]{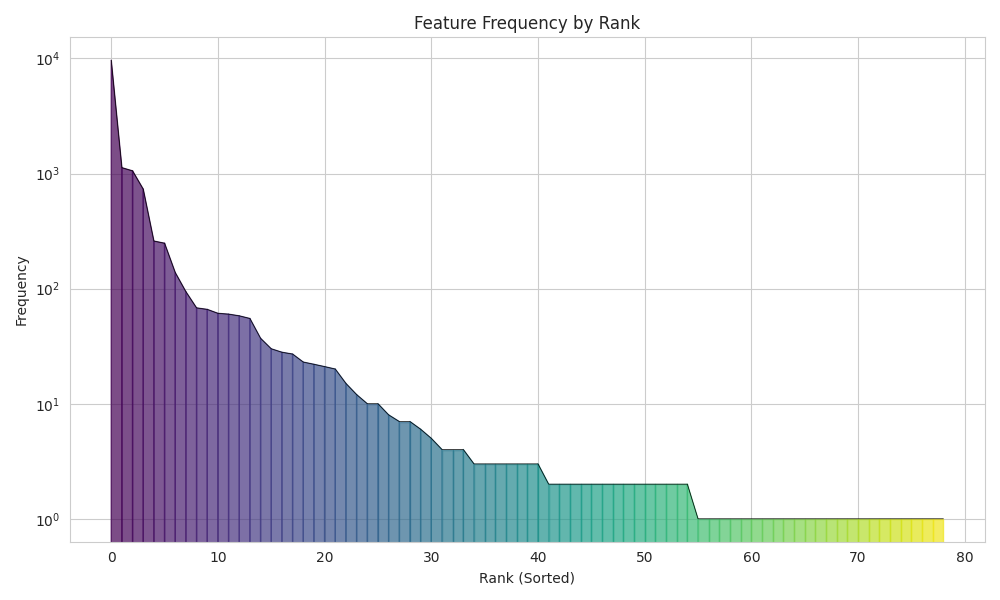}
\caption{Feature usage hierarchy in Gemma 2 2B MMLU task, demonstrating that the policy selectively identifies specific features rather than random selection.}
\label{fig:feature_frequency}
\end{figure}

Feature diversity remains consistent across different hyperparameters within the same task, though it decreases with increased policy layer depth.
As shown in \hyperref[fig:top_features_comparison]{\textcolor{linkblue}{Figure~\ref*{fig:top_features_comparison}}}, tasks requiring longer token generation horizons (GSM8K: 6.695) or higher complexity (MMLU-Pro: 5.476) demonstrate elevated feature diversity compared to shorter response tasks (HarmBench: 2.938), with XSTest (4.757) showing intermediate diversity levels.
Impact scores exhibit an inverse relationship with feature diversity: tasks with lower diversity show higher concentrated impact scores (HarmBench: 0.808), while high-diversity tasks demonstrate more distributed feature contributions (XSTest: 0.357, MMLU-Pro: 0.625).
This pattern suggests that complex reasoning tasks require broader feature engagement, whereas focused tasks benefit from concentrated feature activation.
\hyperref[fig:feature_frequency]{\textcolor{linkblue}{Figure~\ref*{fig:feature_frequency}}} demonstrates that activation frequency exhibits rank-dependent decay with task-specific distribution patterns reflecting underlying diversity characteristics.

\textbf{Four-step Failure Diagnosis Protocol.}
Combining feature diversity with per-token logs yields a practical SAE failure-diagnosis protocol:
(1)~detect unexpectedly low $H$ relative to task complexity;
(2)~localize by checking whether alternative layers produce higher diversity and accuracy;
(3)~recover by switching to a better-covered layer;
(4)~verify that diversity, accuracy, and feature coherence improve.
For tasks requiring distributed reasoning, unusually low diversity flags that the SAE may lack adequate feature coverage.
Known SAE limitations~\citep{engels2025decomposing}, including polysemantic features and missing dictionary concepts (``dark matter''), propagate to CRL's intervention logs.
As AxBench~\citep{wu2025axbench} demonstrates that simple baselines can outperform SAE-based steering, SAE quality directly bounds CRL's effectiveness; however, CRL makes these limitations observable: if the policy collapses to few semantically incoherent features despite AFM, this signals SAE deficiency rather than a CRL failure.

\subsection{Layer-wise Analysis}
\label{sec:layerwise_analysis}

\begin{figure}[t]
  \centering
  \includegraphics[width=0.95\textwidth]{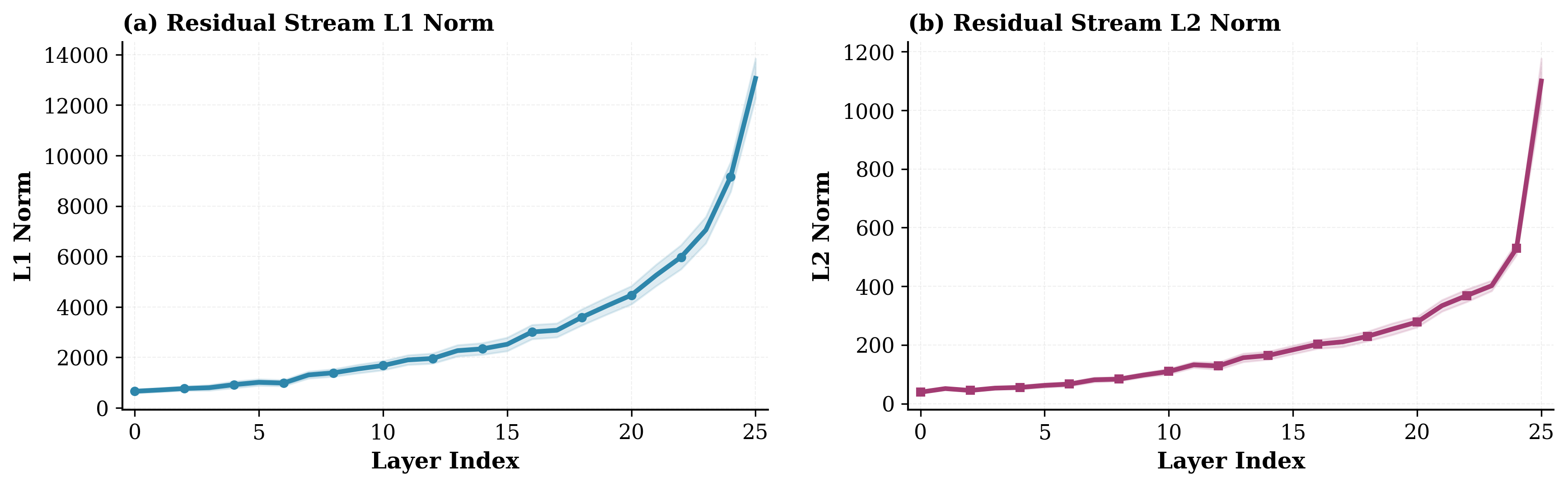}
  \caption{Residual stream norm growth across layers for Gemma 2 2B on MMLU. Norm increases with depth, explaining why identical steering coefficients produce different effects at different layers and motivating our adaptive coefficient strategy.}
  \label{fig:norm}
\end{figure}

\hyperref[fig:norm]{\textcolor{linkblue}{Figure~\ref*{fig:norm}}} shows residual stream norm growth across layers for Gemma 2 2B model on the MMLU task.
This pattern is consistent with prior observations that residual stream norms increase with network depth~\citep{stefanhex2023residual}.

\begin{figure}[h]
\centering
\includegraphics[width=0.49\textwidth]{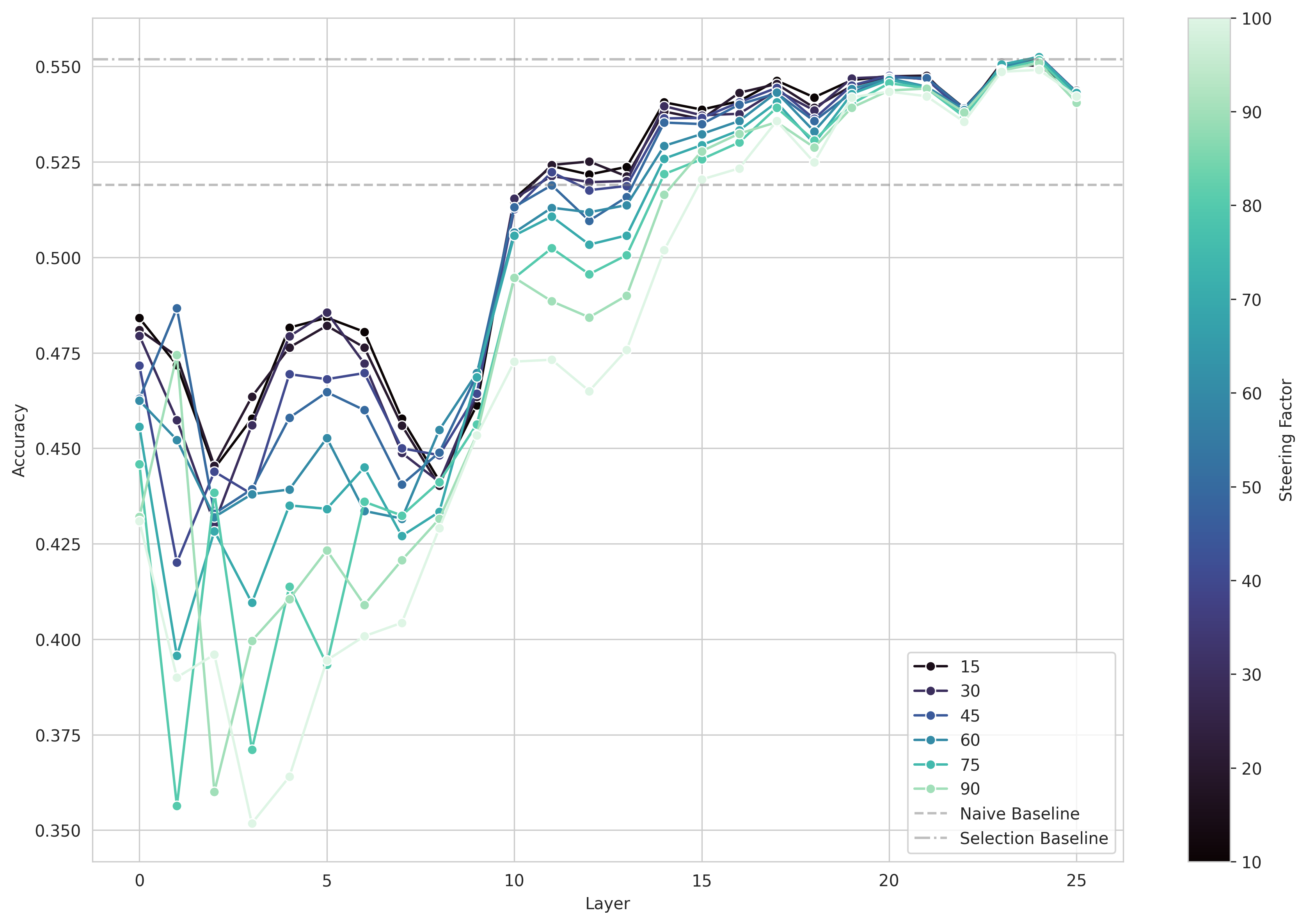}
\includegraphics[width=0.49\textwidth]{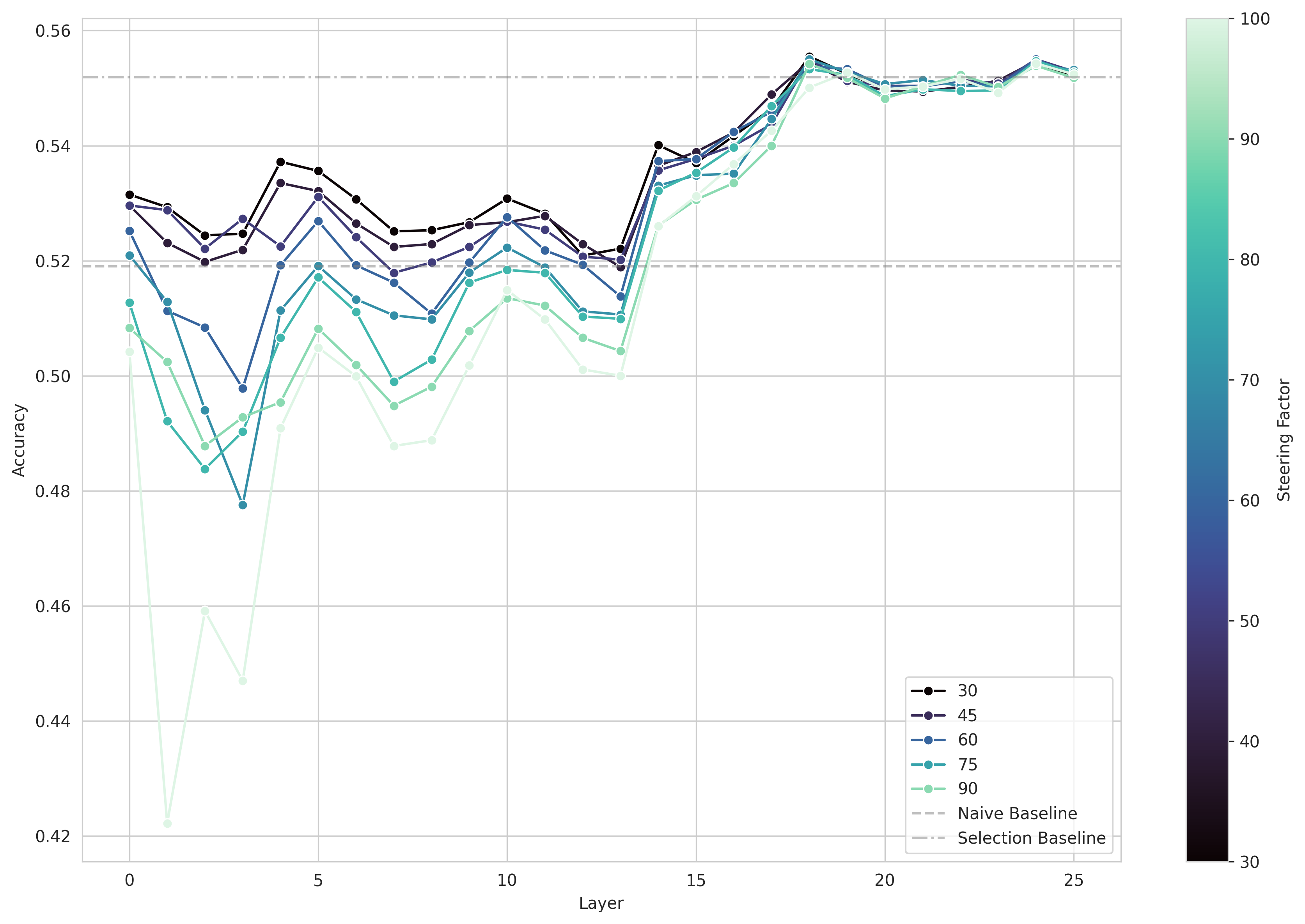}
\caption{MMLU accuracy per fixed coefficient across layers for Gemma 2 2B, showing optimal intervention points for unconstrained (left) and constrained (right) decoding. Later layers generally outperform earlier layers, with large coefficients degrading early-layer performance.}
\label{fig:gemma_mmlu_layers}
\end{figure}

\begin{figure}[h]
\centering
\includegraphics[width=0.49\textwidth]{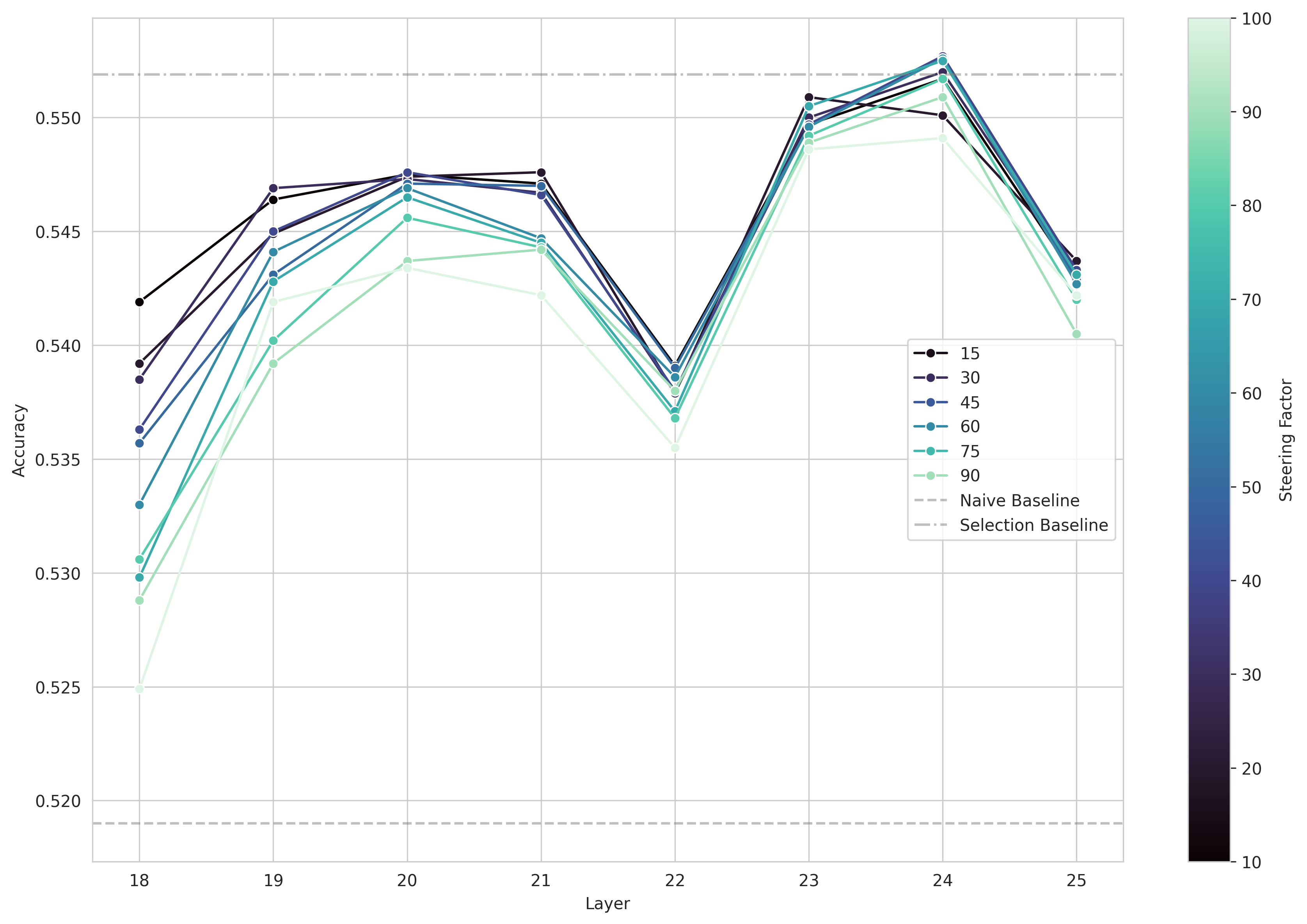}
\includegraphics[width=0.49\textwidth]{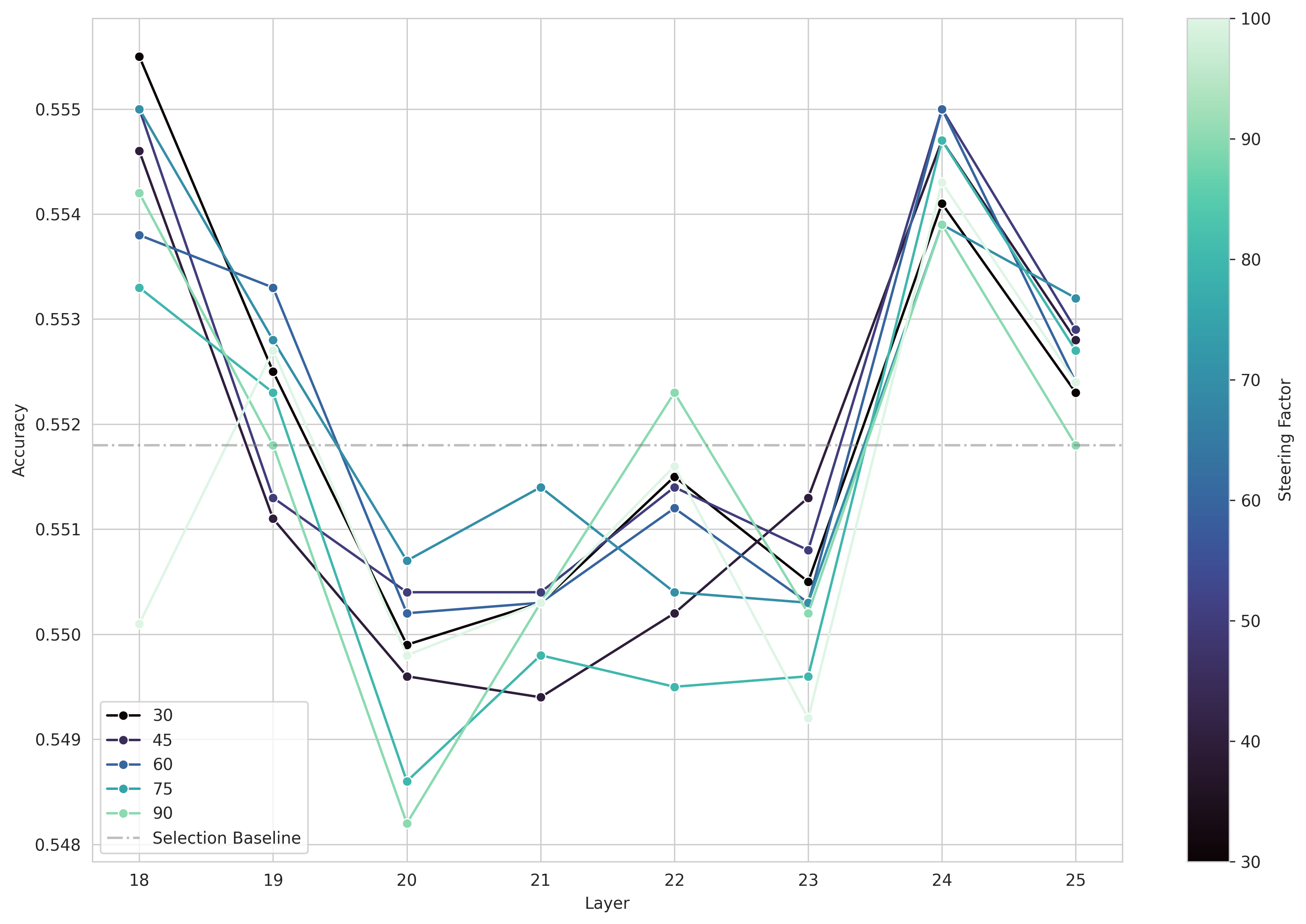}
\caption{MMLU accuracy per fixed coefficient across layers $\ell \geq 18$ for unconstrained (left) and constrained (right) decoding. Unconstrained and constrained settings show different optimal layers, indicating distinct feature requirements for format correction vs factual answering.}
\label{fig:gemma_mmlu_layers_18}
\end{figure}

\begin{figure}[h]
  \centering
  \includegraphics[width=0.49\textwidth]{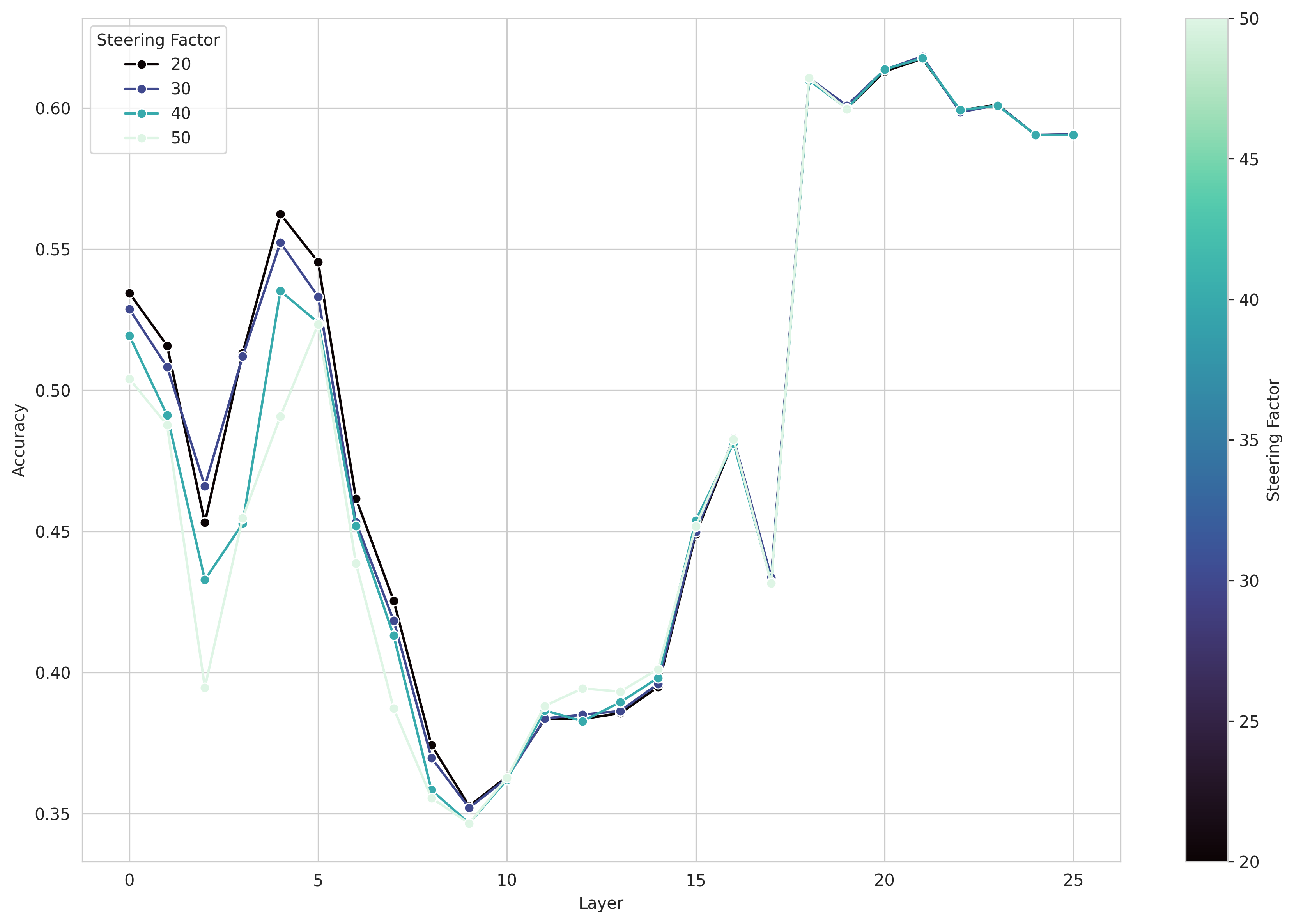}
  \includegraphics[width=0.49\textwidth]{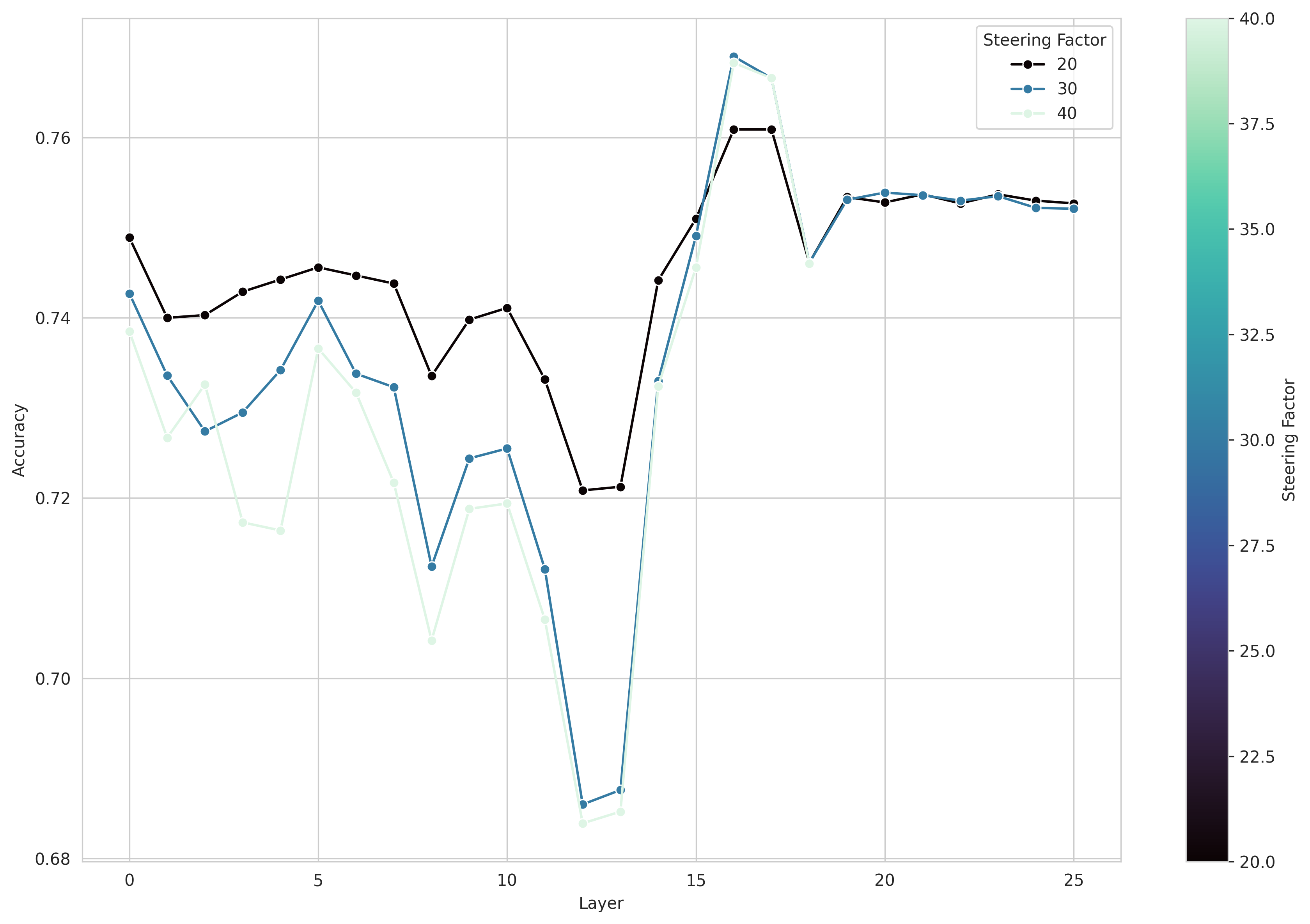}
  \caption{BBQ accuracy per fixed coefficient across layers for ambiguous (left) and disambiguated (right) contexts. Different optimal layers indicate context-dependent bias mitigation strategies.}
  \label{fig:gemma_bbq_layers}
\end{figure}

\begin{figure}[h]
\centering
\includegraphics[width=0.49\textwidth]{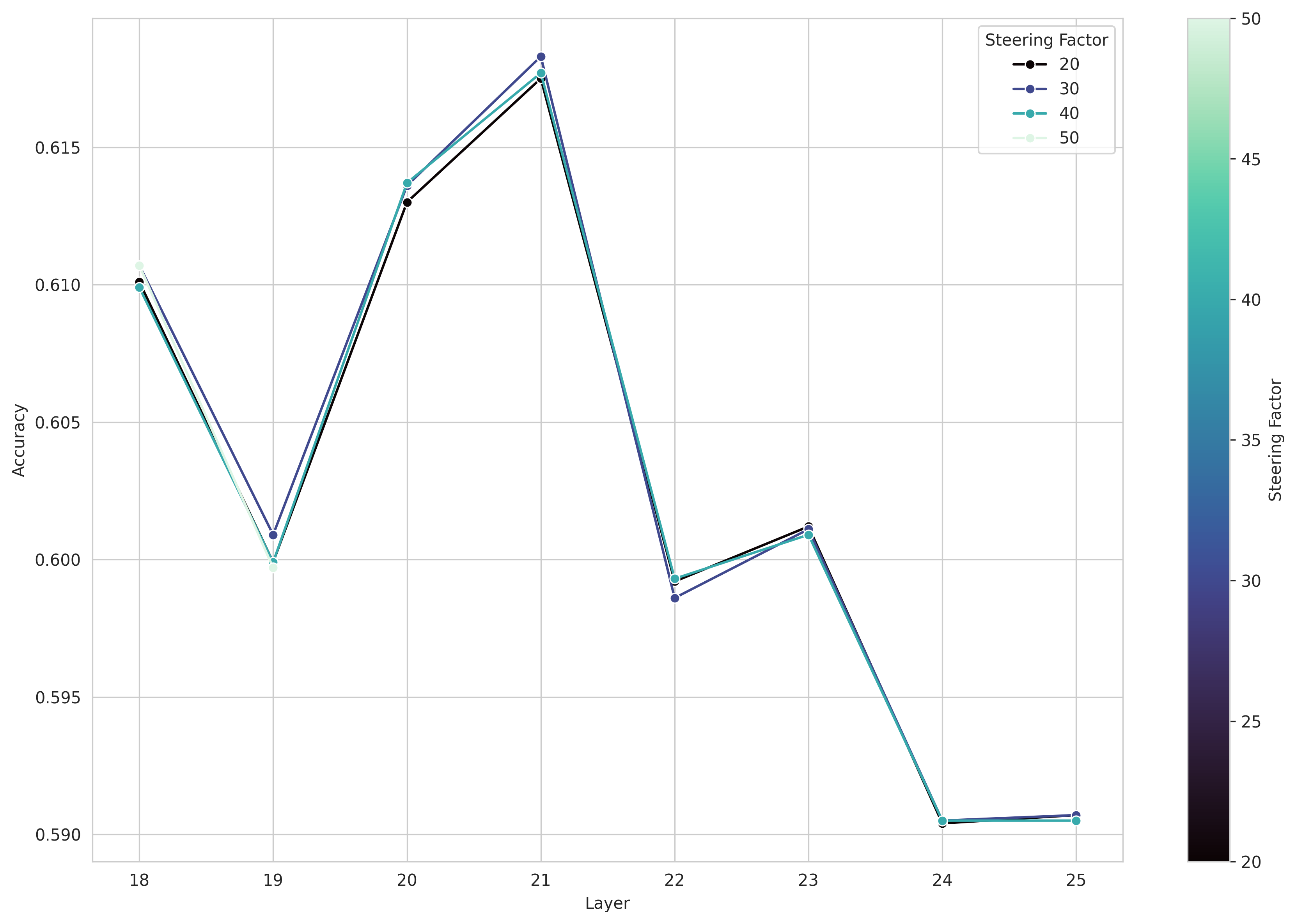}
\includegraphics[width=0.49\textwidth]{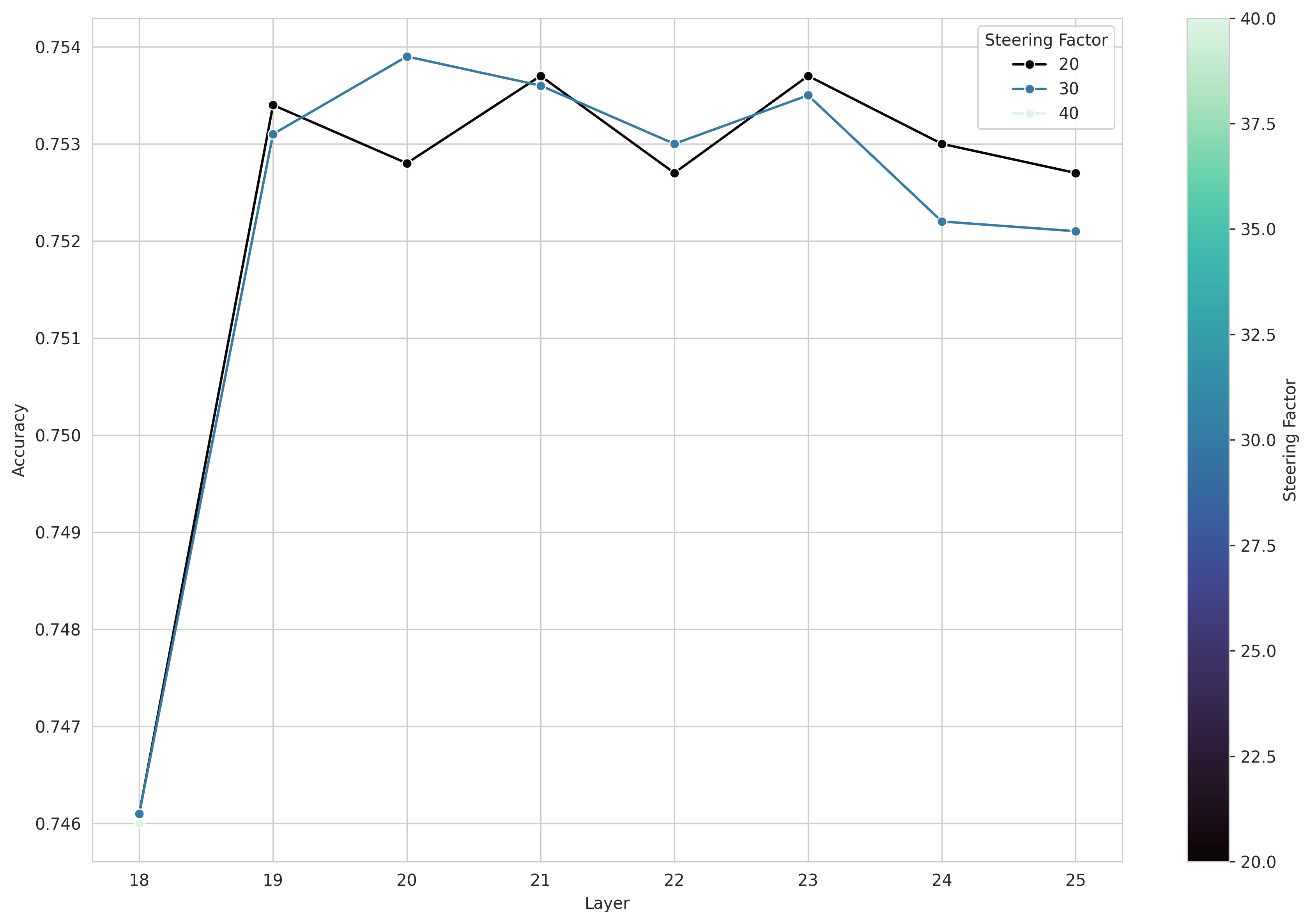}
\caption{BBQ accuracy per fixed coefficient across layers $\ell \geq 18$ for ambiguous (left) and disambiguated (right) contexts. Later layers show distinct optimal coefficients for each context type.}
\label{fig:gemma_bbq_layers_18}
\end{figure}

The analysis for layers $\ell \geq 18$ for BBQ tasks (\hyperref[fig:gemma_bbq_layers_18]{\textcolor{linkblue}{Figure~\ref*{fig:gemma_bbq_layers_18}}}) confirms that ambiguous and disambiguated contexts require different optimal intervention layers.
This suggests that bias mitigation strategies must be adapted to the specific type of ambiguity present in the input.
The layer-wise analysis shows similar patterns across tasks: later layers generally provide more effective intervention points, while early layers show degraded performance with large coefficients.
This pattern holds across different coefficient values and task types, consistent with observations that residual stream norms increase across layers~\citep{stefanhex2023residual}.

\section{Feature Analysis Details}
\label{app:feature_details}

\subsection{Corrected Features for GSM8K Task}

The following features demonstrate positive steering effects, correcting model outputs from incorrect to correct responses:

\begin{itemize}[itemsep=0pt, parsep=0pt, topsep=0pt]
\item \texttt{\href{https://neuronpedia.org/gemma-2-2b/20-gemmascope-res-16k/4504}{4504}} diagrams and schematic representations related to workflows, communication systems, and protocols (correct/incorrect: 321/170, corrected/misguided: 15/4)
\item \texttt{\href{https://neuronpedia.org/gemma-2-2b/20-gemmascope-res-16k/406}{406}} specific mentions of individuals, organizations, and their respective roles or activities (correct/incorrect: 116/113, corrected/misguided: 12/4)
\item \texttt{\href{https://neuronpedia.org/gemma-2-2b/20-gemmascope-res-16k/1440}{1440}} instances of the word "vice" and related terms (correct/incorrect: 161/119, corrected/misguided: 11/4)
\item \texttt{\href{https://neuronpedia.org/gemma-2-2b/20-gemmascope-res-16k/14204}{14204}} indicators of group dynamics and power relations (correct/incorrect: 248/179, corrected/misguided: 16/6)
\item \texttt{\href{https://neuronpedia.org/gemma-2-2b/20-gemmascope-res-16k/10961}{10961}} terms related to statistical methods and implementation details (correct/incorrect: 223/182, corrected/misguided: 20/8)
\item \texttt{\href{https://neuronpedia.org/gemma-2-2b/20-gemmascope-res-16k/2699}{2699}} significant concepts related to problems and solutions within a defined framework (correct/incorrect: 234/187, corrected/misguided: 17/7)
\item \texttt{\href{https://neuronpedia.org/gemma-2-2b/20-gemmascope-res-16k/2317}{2317}} phrases indicating comparisons or relationships between two entities or elements (correct/incorrect: 120/86, corrected/misguided: 12/5)
\item \texttt{\href{https://neuronpedia.org/gemma-2-2b/20-gemmascope-res-16k/2720}{2720}} mathematical expressions and equations related to functions and inequalities (correct/incorrect: 151/150, corrected/misguided: 11/5)
\item \texttt{\href{https://neuronpedia.org/gemma-2-2b/20-gemmascope-res-16k/347}{347}} phrases and concepts related to accountability and compliance (correct/incorrect: 148/129, corrected/misguided: 15/7)
\item \texttt{\href{https://neuronpedia.org/gemma-2-2b/20-gemmascope-res-16k/7708}{7708}} mathematical operations and expressions in various forms (correct/incorrect: 188/217, corrected/misguided: 19/9)
\end{itemize}

\subsection{Additional Corrected and Misguided Examples}
\label{subsec:additional_corrected_example}

\begin{figure}[h]
\centering
\includegraphics[width=\textwidth]{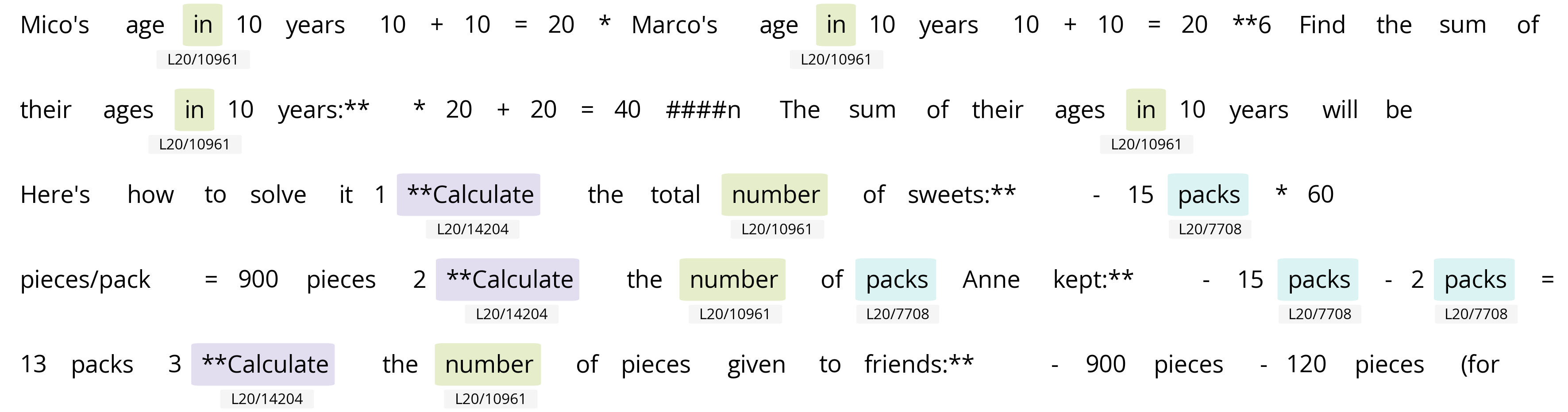}
\caption{GSM8K corrected case 1: Contextually appropriate feature steering activates relevant numerical tokens, strengthening reasoning.}
\label{fig:corrected_example_1}
\end{figure}

\begin{figure}[h]
\centering
\includegraphics[width=\textwidth]{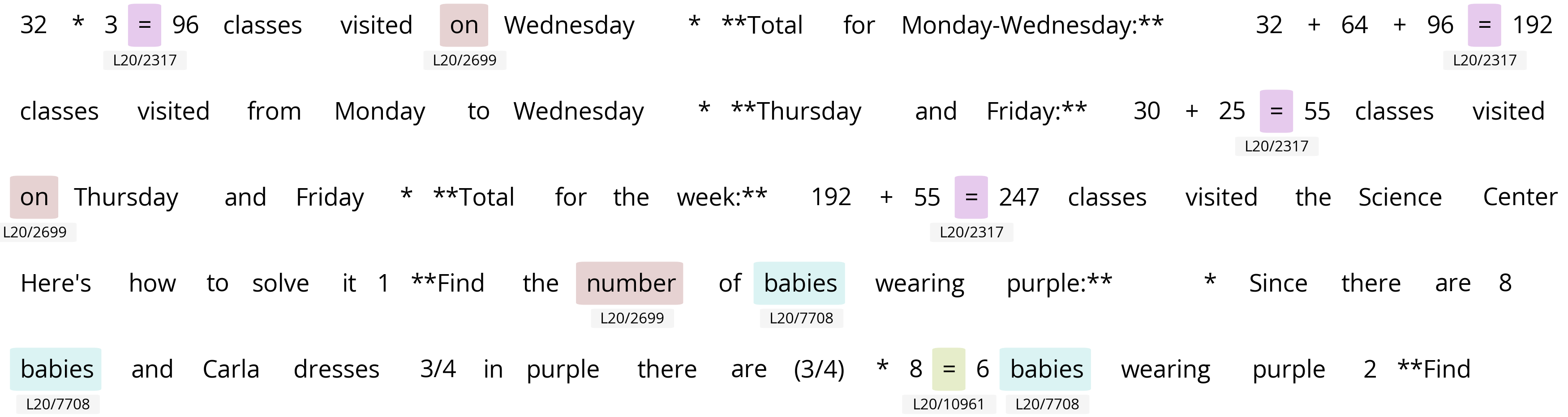}
\caption{GSM8K corrected case 2: Semantically coherent feature activations align with equality tokens, guiding correct computation.}
\label{fig:corrected_example_2}
\end{figure}

\begin{figure}[h]
\centering
\includegraphics[width=\textwidth]{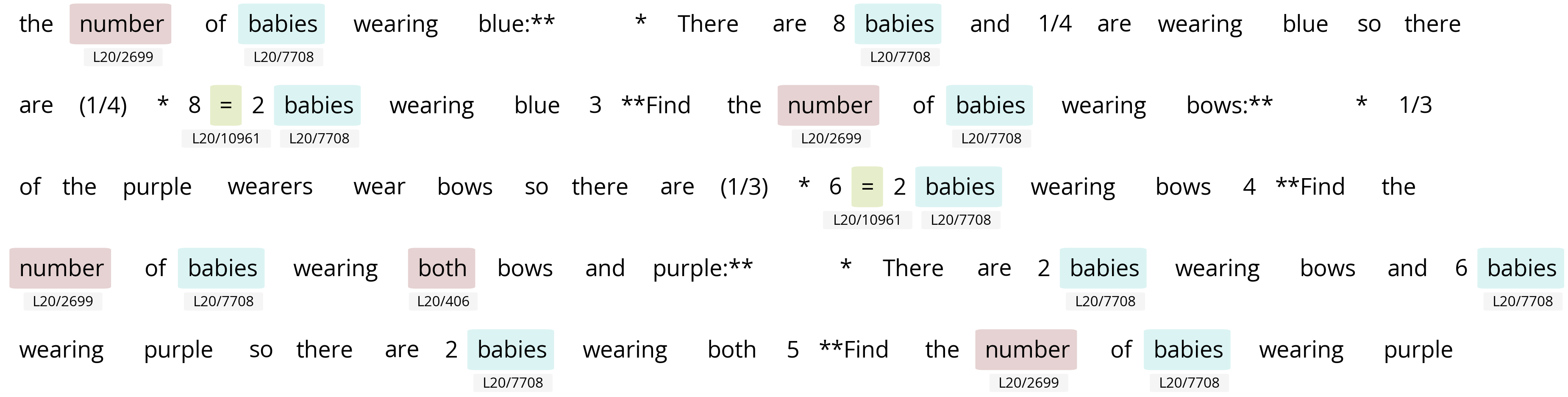}
\caption{GSM8K corrected case 3: Feature 7708, described as \textit{mathematical operations and expressions in various forms}, activates on quantitative units like "babies" and "packs", demonstrating lexical generalization beyond surface forms.}
\label{fig:additional_corrected_example_3}
\end{figure}

\begin{figure}[h]
\centering
\includegraphics[width=\textwidth]{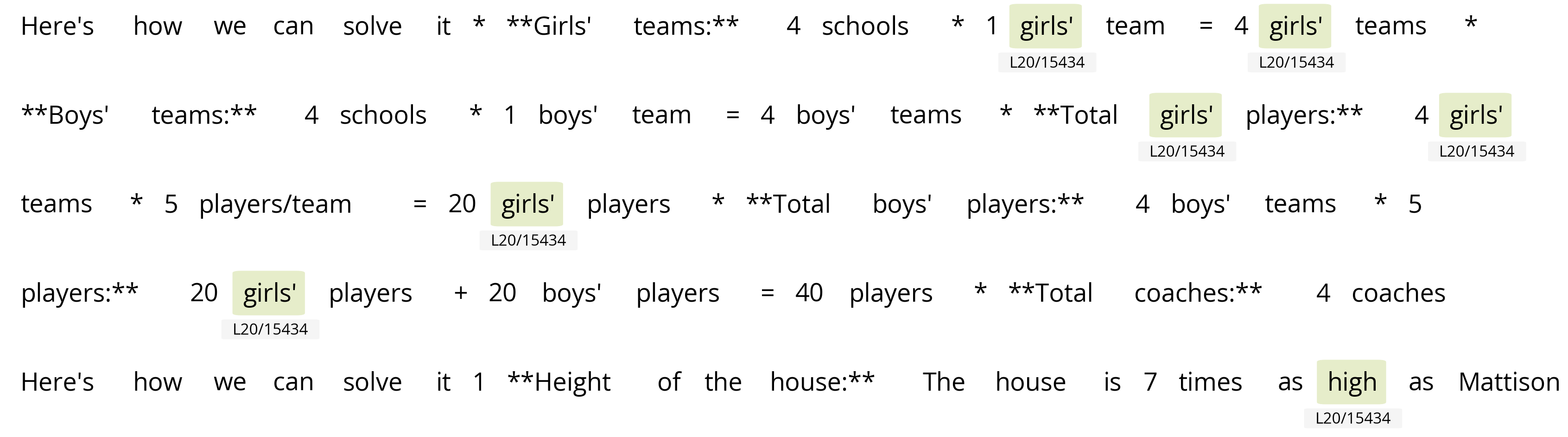}
\caption{GSM8K misguided case 1: Selection of task-irrelevant features interferes with mathematical reasoning and diverts the solution process.}
\label{fig:misguided_example_1}
\end{figure}

\begin{figure}[h]
\centering
\includegraphics[width=\textwidth]{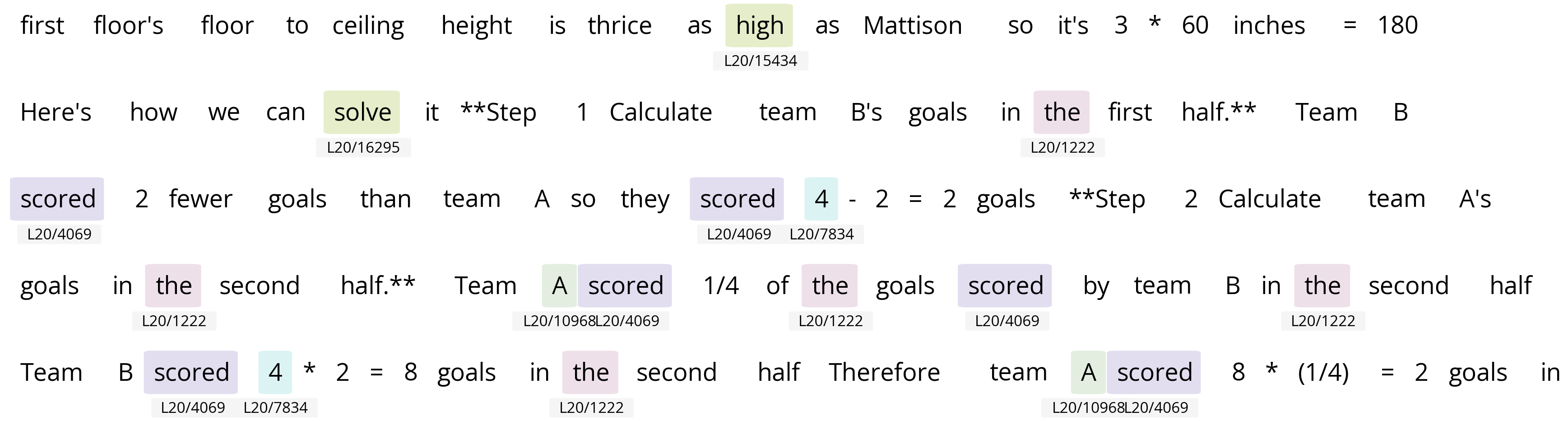}
\caption{GSM8K misguided case 2: Token-feature misalignment produces semantically incoherent activations, degrading reasoning performance.}
\label{fig:misguided_example_2}
\end{figure}

\subsection{Misguided Features for GSM8K Task}

Conversely, these features demonstrate negative steering effects, inadvertently changing correct responses to incorrect ones:

\begin{itemize}[itemsep=0pt, parsep=0pt, topsep=0pt]
\item \texttt{\href{https://neuronpedia.org/gemma-2-2b/20-gemmascope-res-16k/7999}{7999}} mathematical expressions or equations (correct/incorrect: 40/42, corrected/misguided: 1/10)
\item \texttt{\href{https://neuronpedia.org/gemma-2-2b/20-gemmascope-res-16k/4069}{4069}} mathematical notation and geometric properties related to circles and angles (correct/incorrect: 121/100, corrected/misguided: 7/16)
\item \texttt{\href{https://neuronpedia.org/gemma-2-2b/20-gemmascope-res-16k/13752}{13752}} elements related to job creation and economic context (correct/incorrect: 109/78, corrected/misguided: 5/10)
\item \texttt{\href{https://neuronpedia.org/gemma-2-2b/20-gemmascope-res-16k/10968}{10968}} specific coding or mathematical terms and phrases related to programming or data analysis (correct/incorrect: 99/112, corrected/misguided: 8/16)
\item \texttt{\href{https://neuronpedia.org/gemma-2-2b/20-gemmascope-res-16k/16295}{16295}} significant terms and phrases related to scientific studies, particularly in the context of law, medicine, and research methodologies (correct/incorrect: 99/93, corrected/misguided: 6/11)
\item \texttt{\href{https://neuronpedia.org/gemma-2-2b/20-gemmascope-res-16k/12174}{12174}} references to understanding and problem-solving processes (correct/incorrect: 180/157, corrected/misguided: 10/17)
\item \texttt{\href{https://neuronpedia.org/gemma-2-2b/20-gemmascope-res-16k/15434}{15434}} complex relationships involving socio-legal and psychological themes (correct/incorrect: 113/125, corrected/misguided: 8/13)
\item \texttt{\href{https://neuronpedia.org/gemma-2-2b/20-gemmascope-res-16k/1222}{1222}} sentences expressing emotional vulnerability and complex interpersonal dynamics (correct/incorrect: 74/61, corrected/misguided: 7/11)
\item \texttt{\href{https://neuronpedia.org/gemma-2-2b/20-gemmascope-res-16k/7834}{7834}} indicative symbols and formatting used in programming or mathematical expressions (correct/incorrect: 97/101, corrected/misguided: 9/14)
\item \texttt{\href{https://neuronpedia.org/gemma-2-2b/20-gemmascope-res-16k/3415}{3415}} patterns of conditional phrases and expressions of uncertainty (correct/incorrect: 92/108, corrected/misguided: 8/12)
\end{itemize}

\subsection{Additional Misguided Example}

\begin{figure}[h]
\centering
\includegraphics[width=\textwidth]{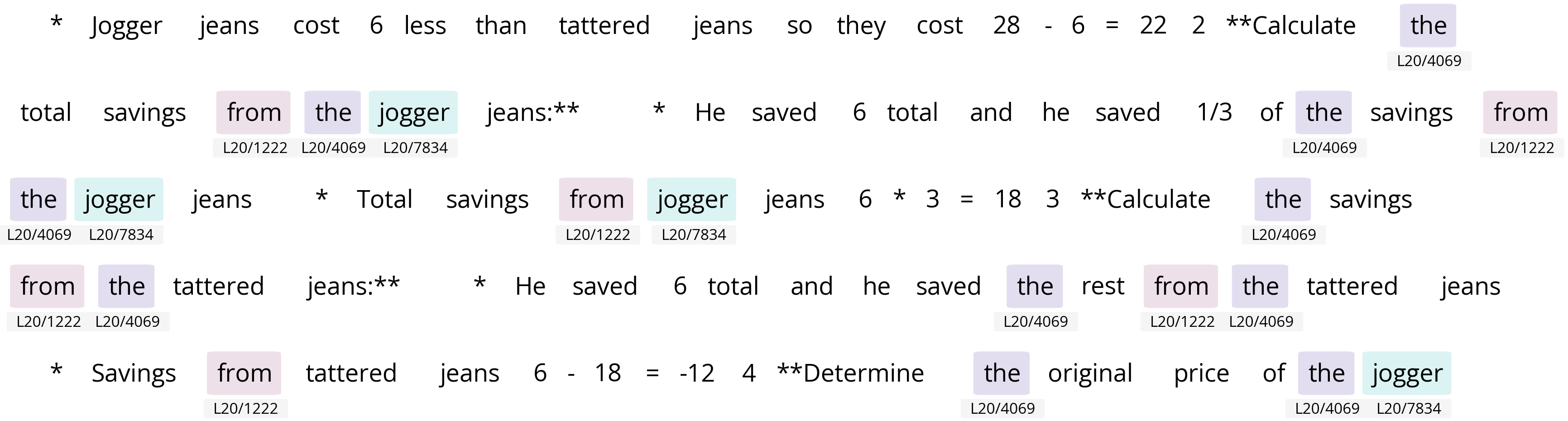}
\caption{GSM8K misguided example 3: Feature interference causes abandonment of correct solution path.}
\label{fig:misguided_example_3}
\end{figure}

Figure~\ref{fig:misguided_example_3} illustrates a case where misguided feature steering causes the model to abandon a correct solution path. The steered model selects features associated with task-irrelevant semantics (e.g., socio-legal framing or programming constructs from the misguided feature list above), which interfere with the arithmetic reasoning required for the problem. This token-level interference is visible in CRL's intervention logs: the description of the selected feature does not align with the mathematical operation at that position, which provides a concrete diagnostic signal for identifying when steering goes wrong.

\subsection{Additional Branch Analysis}
\label{subsec:additional_branch_analysis}

We analyze decision bifurcations at critical tokens, comparing layer 10 and layer 20 feature selections across GSM8K tasks.
The following cases show different steering outcomes depending on which layer's features are selected.

\textbf{Layer-Specific Semantic Patterns.}
Layer 10 features exhibit concrete, syntactic properties while layer 20 features demonstrate abstract, contextual semantics.
Balloon pricing is the exception: it activates layer 10's ``mathematical proofs'' feature recognizing logical derivation structure, whereas layer 20's ``summaries'' feature treats it as tutorial explanation.
Rice consumption with ``kg/member'' notation activates layer 10's ``clinical examinations'' feature due to surface similarity with medical dosage, while layer 20's ``assessments and strategic plans'' is consistent with planning semantics.
Restaurant problems show layer 10's ``mathematical notation'' feature outperforming layer 20's ``food and dining'' by prioritizing numerical structure.
Layer 10 thus performs surface-level pattern matching (notation $\to$ similar-looking domains), while layer 20 captures task-relevant abstract structure, consistent with the residual stream hierarchy where early layers encode syntactic/structural information and later layers encode abstract task semantics.
Feature labels alone provide limited predictive value for steering outcomes: ``emotional relationships'' solves speed calculations while ``mathematical notation'' fails on some arithmetic, indicating that learned steering effects diverge from semantic interpretations.

\begin{figure}[h]
\centering
\includegraphics[width=1\textwidth]{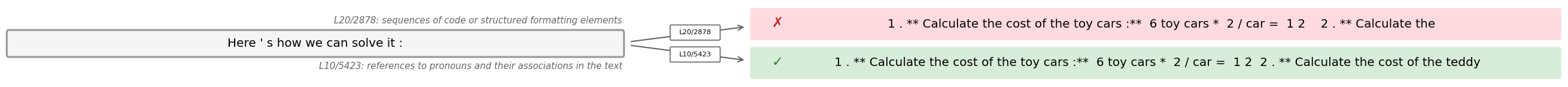}
\includegraphics[width=1\textwidth]{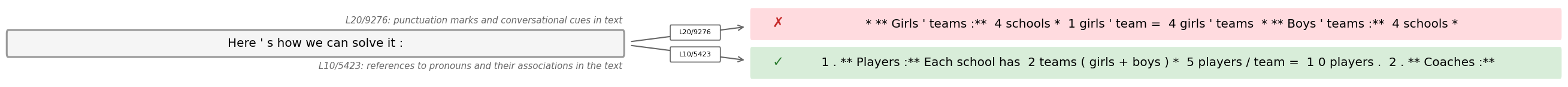}
\includegraphics[width=1\textwidth]{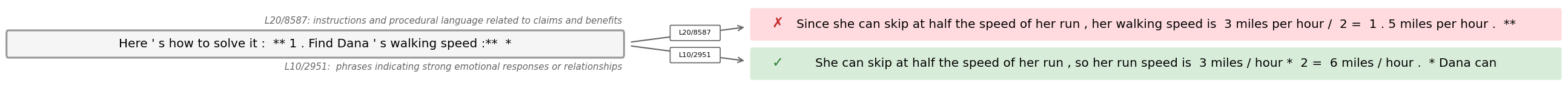}
\caption{Cases where layer 10 features succeed. Top: Toy calculation where L10's "pronoun references" captures concrete items while L20's "code formatting" misapplies structure. Middle: Tournament problem where L10's "pronoun references" handles entity tracking while L20's "punctuation cues" misidentifies structure. Bottom: Speed calculation where L10's "emotional relationships" correctly solves the problem while L20's "claims and benefits" misapplies procedural framing.}
\label{fig:branch_l10_wins}
\end{figure}

In these three cases, L10 features produce correct outputs while L20 features fail. Both the toy calculation and tournament problems use L10's "pronoun references" feature, which generates accurate enumerated calculations, whereas L20's "code formatting" and "punctuation cues" features produce incorrect results. The speed conversion case shows L10's "emotional relationships" feature generating the correct formula ("3 miles/hour * 2 = 6"), while L20's "claims and benefits" produces wrong logic ("3 miles per hour / 2 = 1.5"). Feature labels alone do not predict which layer succeeds: the "emotional relationships" label bears no obvious relation to speed calculation.

\begin{figure}[h]
\centering
\includegraphics[width=1\textwidth]{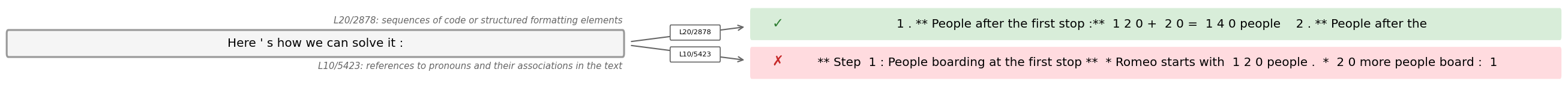}
\includegraphics[width=1\textwidth]{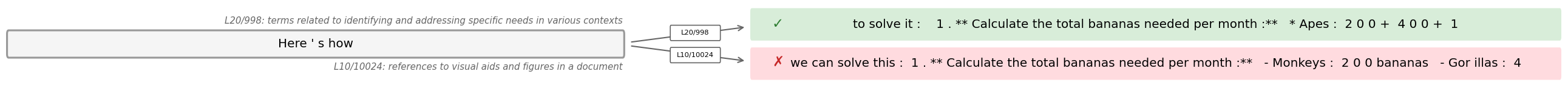}
\includegraphics[width=1\textwidth]{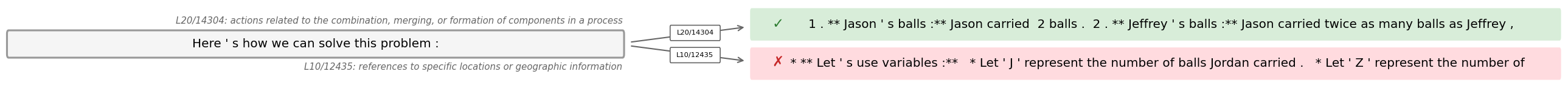}
\includegraphics[width=1\textwidth]{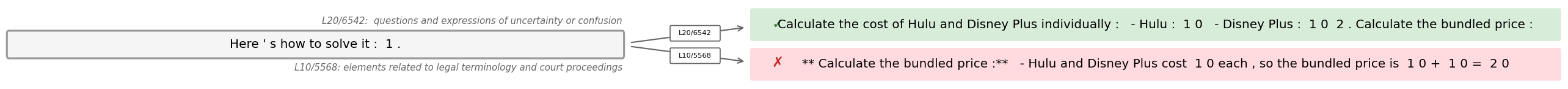}
\includegraphics[width=1\textwidth]{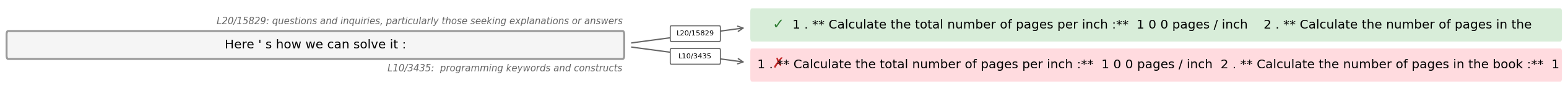}
\caption{Cases where layer 20 features succeed. Top: Bus problem where L20's "code formatting" correctly structures sequential steps while L10's "pronoun references" fails. Second: Banana calculation where L20's "needs identification" captures requirement semantics while L10's "visual aids" produces incorrect output. Third: Variable abstraction where L20's "combination and formation" proceeds with direct calculation while L10's "geographic information" chooses variable-based approach. Fourth: Subscription pricing where L20's "uncertainty expressions" captures choice scenarios while L10's "legal terminology" produces incorrect output. Bottom: Pages calculation where L20's "questions and inquiries" captures exploratory reasoning while L10's "programming constructs" treats it as code.}
\label{fig:branch_l20_wins}
\end{figure}

In these five cases, L20 features produce correct outputs while L10 features fail. The bus passenger problem shows L20's "code formatting" generating correct sequential tracking ("1. **People after the first stop:** 120 + 20 = 140"), while L10's "pronoun references" produces verbose, incorrect narrative. The variable abstraction case shows L10 choosing a variable-based approach ("Let 'J' represent...") while L20 proceeds directly with numerical calculation ("Jason carried 2 balls"). In the subscription pricing problem, L10's "legal terminology" generates incorrect output, while L20's "uncertainty expressions" handles the choice scenario. The banana and pages problems similarly show L10 features producing incorrect outputs, whereas L20 features generate correct multi-step solutions.

\textbf{Observations.}
Across these eight examples, we observe a pattern: L10 succeeds on three cases involving discrete item counting and direct arithmetic, while L20 succeeds on five cases involving accumulation over multiple steps, choice scenarios, and variable reasoning.
L10 failures show patterns of surface matching (e.g., financial terminology leading to "legal terminology", notation leading to "programming constructs"), whereas L20 failures show structural misapplication to simple problems (e.g., "code formatting" on toy calculation).
Feature labels provide limited predictive value: "emotional relationships" solves speed calculations, "code formatting" performs both correctly and incorrectly on different tasks.
These divergent outcomes demonstrate that layer choice affects steering effectiveness, which CRL's layer-specific policy exploits to adapt to different problem structures.

\begin{figure}[h]
\centering
\includegraphics[width=\textwidth]{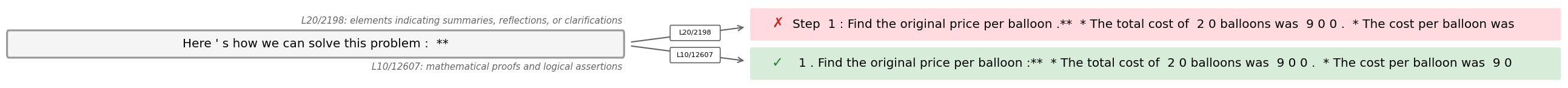}
\includegraphics[width=\textwidth]{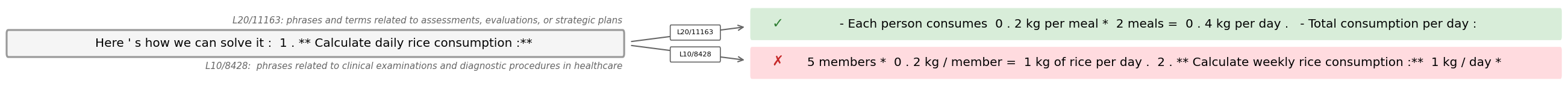}
\includegraphics[width=\textwidth]{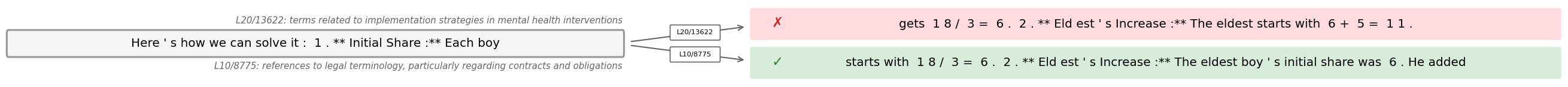}
\includegraphics[width=\textwidth]{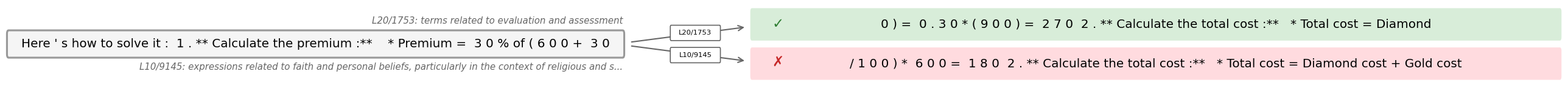}
\includegraphics[width=\textwidth]{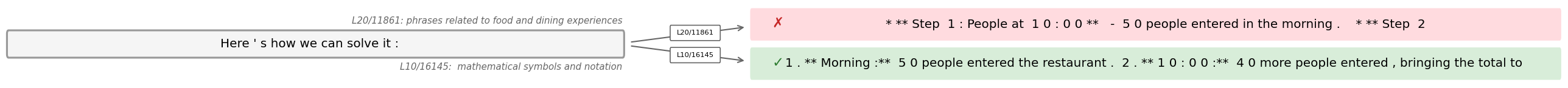}
\includegraphics[width=\textwidth]{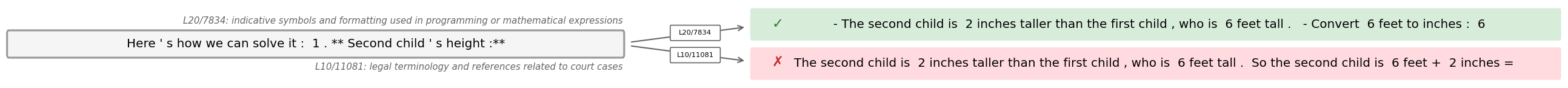}
\caption{Additional branch comparison cases between layer 10 and layer 20 feature semantics. Each panel shows a problem context, competing features from both layers, and the correct steering selection.}
\label{fig:branch_comparison}
\end{figure}

\subsection{Feature Behavior and Bottleneck Diagnostics}
\label{subsec:feature_behavior}

\textbf{Feature Controllability.}
Our experiments reveal insights into SAE feature controllability.
Corrected cases demonstrate clear semantic coherence between feature activations and their documented descriptions, as shown in GSM8K examples.
Misguided cases provide additional diagnostic value: they reveal where alignments between features and tokens break down, enabling targeted improvements.
These patterns suggest that feature interactions extend beyond simple linear combinations, opening directions for multi-feature coordination strategies.

\textbf{Bottleneck Diagnostics.}
CRL's critic trajectories diagnose where improvements originate.
BBQ demonstrates effective policy-critic coordination with clear distinction between sample categories.
For MMLU, CRL learns to eliminate invalid responses that the baseline model generates instead of selecting from given options (\hyperref[fig:hallucination_answers]{\textcolor{linkblue}{Figure~\ref*{fig:hallucination_answers}}}).
This task-dependent pattern enables targeted optimization strategies.

\subsection{Cross-Model Effectiveness Axes}
\label{subsec:effectiveness_axes}

Across both Gemma 2B and LLaMA 8B, two axes predict CRL's effectiveness.
First, \textit{signal clarity}: CRL shows strongest gains where there is room for improvement (BBQ Ambig: +5.69 Gemma/+1.07 LLaMA, HarmBench: +7.66/+5.36) and modest gains where baselines are strong (BBQ Disambig: +0.47/+0.09, MMLU-Pro: +0.19/+1.03).
Second, \textit{objective alignment}: CRL struggles when the desired direction opposes the model's default tendency, as in LLaMA's XSTest decline ($-$1.90), where reducing over-refusal conflicts with shared safety representations.

\subsection{Extended Limitations and Future Work}
\label{subsec:extended_limitations}

\textbf{Multi-layer steering vs.\ attribution.}
The multi-layer variant achieves stronger steering (87.14\% on HarmBench vs.\ 49.12\% single-layer) but trades attribution clarity, and top-$k$ selection with interaction analysis is future work.

\textbf{Scalability.}
Each task/layer combination requires separate PPO training, with policy/critic parameters representing 3.3--3.8\% of base model size and 8--9\% inference overhead.
Scaling to 100K+ dictionaries may require efficient architectures.

\textbf{Reward design and HarmBench evaluator coupling.}
Safety tasks use classifier-based rewards (a RoBERTa-based refusal classifier; \hyperref[subsec:task_rewards]{\textcolor{linkblue}{Appendix~\ref*{subsec:task_rewards}}}) and SimpleQA uses STS similarity; CRL's intervention logs make learned behaviors auditable but do not formally prevent reward-metric coupling.
On HarmBench in particular, the same refusal classifier serves as both the PPO reward signal and the test-time evaluator: the +7.66\,pp gain therefore partly reflects optimization against the evaluator rather than purely held-out refusal robustness, and the absolute number is not directly comparable to attack-success-rate leaderboards that use automated jailbreak attacks (GCG, PAIR). As partial evidence that the gain is not merely reward-metric optimization, random-feature steering, which never optimizes against the reward, gains far less than the learned policy (45.35 vs 49.12 refusal, \hyperref[tab:steering_baselines]{\textcolor{linkblue}{Table~\ref*{tab:steering_baselines}}}).
A held-out evaluator independent of the training signal, together with a standard attack protocol, would be required to fully disentangle these factors.

\textbf{Complementarity to other methods.}
CRL is complementary to fine-tuning, providing additive gains on top of supervised fine-tuned models, and complements activation-based SAE analysis with dynamic intervention-based views of model behavior.

\subsection{Computational Cost and Coefficient Sensitivity}
\label{subsec:runtime_overhead}

CRL introduces minimal computational overhead during inference (8--9\% latency on A100).
The overhead consists of SAE encoding, policy forward pass, and steering vector computation.
\hyperref[tab:compute_cost]{\textcolor{linkblue}{Table~\ref*{tab:compute_cost}}} reports per-task training times and resource requirements.

\begin{table}[ht!]
\centering
\small
\captionof{table}{Training cost per task on a single A100. Policy+critic parameters are 84M (3.3\% of Gemma 2B) and 302M (3.8\% of LLaMA 8B). All runs use batch size 8, 4000 samples, 1000 PPO iterations.}
\label{tab:compute_cost}
\begin{tabular}{llccc}
\toprule
\textbf{Model} & \textbf{Task} & \textbf{Gen Length} & \textbf{Training Time} & \textbf{Peak Memory} \\
\midrule
Gemma 2 2B & MMLU & 1 token & $\sim$37 min & $\sim$6.8 GB \\
Gemma 2 2B & BBQ & 1 token & $\sim$55 min & $\sim$6.8 GB \\
Gemma 2 2B & GSM8K & 1024 tokens & $\sim$5.5 hrs & $\sim$6.8 GB \\
Gemma 2 2B & HarmBench & 32 tokens & $\sim$3 hrs & $\sim$6.8 GB \\
LLaMA 3.1 8B & MMLU & 1 token & $\sim$2.5 hrs & $\sim$15.7 GB \\
\bottomrule
\end{tabular}
\end{table}

For steering coefficients, our adaptive strategy averages the magnitudes of activations from correctly predicted samples: $c = \frac{1}{|D^+|} \sum_{x \in D^+} \|\mathbf{a}_t \mathbf{W}_{dec}\|$, where $D^+$ denotes correct samples.
This eliminates manual tuning across 26 layers.
\hyperref[tab:coefficient_sensitivity]{\textcolor{linkblue}{Table~\ref*{tab:coefficient_sensitivity}}} shows that moderate coefficients (25-50) achieve optimal performance, with analysis for each layer in \hyperref[fig:gemma_mmlu_layers]{\textcolor{linkblue}{Figure~\ref*{fig:gemma_mmlu_layers}}}.

\begin{table}[ht!]
\centering
\small
\begin{minipage}{0.48\textwidth}
  \centering
  \captionof{table}{Runtime overhead. CRL adds 8--9\% latency.}
  \label{tab:runtime_overhead}
  \begin{tabular}{lcc}
  \toprule
  \textbf{Task} & \textbf{Type} & \textbf{Overhead} \\
  \midrule
  MMLU & Single & +8.0\% \\
  BBQ & Single & +7.8\% \\
  GSM8K & Multi & +8.8\% \\
  HarmBench & Multi & +9.1\% \\
  \bottomrule
  \end{tabular}
\end{minipage}
\hfill
\begin{minipage}{0.48\textwidth}
  \centering
  \captionof{table}{Coefficient sensitivity on MMLU (layer 24).}
  \label{tab:coefficient_sensitivity}
  \begin{tabular}{lccccc}
  \toprule
  \textbf{Coef.} & 10 & 25 & 50 & 75 & 100 \\
  \midrule
  Acc. (\%) & 53.2 & 54.8 & 55.4 & 54.1 & 52.3 \\
  \bottomrule
  \end{tabular}
\end{minipage}
\end{table}

\textbf{Complementarity with Fine-tuning.}
CRL provides additive gains when applied on top of supervised fine-tuned models.
On MMLU, accuracy improves from 55.2\% (base) to 55.7\% (SFT) to 56.1\% (SFT + CRL), consistent with learned feature steering providing signals complementary to fine-tuning.
CRL applies on top of an SFT model without further weight updates, and the Gemma Scope SAEs transfer across model variants without retraining the autoencoder.

\subsection{Cross-Task Transfer Analysis}
\label{subsec:cross_task_transfer_appendix}

To evaluate whether CRL captures task-specific mechanisms or dataset-specific shortcuts, we apply policies trained on one task to other tasks, using the training task's optimal layer. \hyperref[tab:cross_task_full]{\textcolor{linkblue}{Table~\ref*{tab:cross_task_full}}} reports the full results alongside the performance of native CRL for reference.

\begin{center}
\small
\captionof{table}{Cross-task transfer results. ``Native CRL'' shows the target task's own trained policy for reference. All experiments use the source task's optimal layer.}
\label{tab:cross_task_full}
\begin{tabular}{lcccc}
\toprule
\textbf{Train $\to$ Eval} & \textbf{Layer} & \textbf{Acc.\ (\%)} & \textbf{vs.\ Base} & \textbf{Native CRL} \\
\midrule
MMLU $\to$ GSM8K & 24 & 31.01 & $-$10.31 & 55.65 (L20) \\
GSM8K $\to$ MMLU & 20 & 52.24 & +0.18 & 55.37 (L24) \\
HarmBench $\to$ XSTest & 21 & 86.35 & +1.27 & 87.62 (L12) \\
BBQ $\to$ MMLU & 5 & 53.55 & +1.49 & 55.37 (L24) \\
\bottomrule
\end{tabular}
\end{center}

The transfer pattern is asymmetric: MMLU features actively harm GSM8K ($-$10.31\%), while GSM8K features remain inert on MMLU (+0.18\%).
If the policy captured dataset-specific shortcuts (e.g., bias toward numeric outputs), GSM8K features should also disrupt MMLU, but they do not.
Features from HarmBench slightly improve XSTest (+1.27\%), consistent with shared safety representations across related tasks.
This asymmetry is not explained by layer distance alone. MMLU$\to$GSM8K and GSM8K$\to$MMLU have identical layer distance ($|\Delta|=4$) yet opposite effects ($-10.31$ vs $+0.18$), while the pairs with larger distance, HarmBench$\to$XSTest ($|\Delta|=9$, $+1.27$) and BBQ$\to$MMLU ($|\Delta|=19$, $+1.49$), are the least harmful. A distance-only account, on which damage is a function of $|\Delta\text{layer}|$, predicts neither the sign reversal at fixed distance nor harm that fails to grow with distance. We note what this does not exclude: a \emph{directed} depth account, since applying layer-24 features at layer 20 is not the same operation as the reverse. The interference is therefore structured by the task rather than a function of intervention depth alone. With four pairs this is evidence against a pure depth account rather than a demonstration of circuit overlap; matched-versus-mismatched-depth pairs are the control we flag as next.

\subsection{Dynamic Coefficient Ablation}
\label{subsec:dynamic_coeff_ablation}

The steering coefficient $c$ is fixed during training, and the policy learns to select features calibrated to this static scale.
We tested three per-token dynamic alternatives (\hyperref[tab:dynamic_coeff]{\textcolor{linkblue}{Table~\ref*{tab:dynamic_coeff}}}): norm-scaled ($c \propto \|\mathbf{x}_t\|$), activation-scaled ($c$ set to the natural SAE activation of the selected feature), and critic-gated ($c \propto V_\phi(\mathbf{s}_t)$).

\begin{center}
\small
\captionof{table}{Variants with dynamic coefficients vs.\ the static one (dataset-averaged). All dynamic variants degrade performance due to mismatch between training and inference in the learned feature-coefficient coupling.}
\label{tab:dynamic_coeff}
\begin{tabular}{lcc}
\toprule
\textbf{Method} & \textbf{HarmBench} & \textbf{GSM8K} \\
\midrule
Static (dataset-averaged) & 49.75 & 55.65 \\
Norm-scaled ($c \propto \|\mathbf{x}_t\|$) & 45.00 & 41.39 \\
Activation-scaled ($c = $ natural SAE activation) & 41.79 & 40.33 \\
Critic-gated ($c \propto V_\phi(\mathbf{s}_t)$) & 43.93 & 41.47 \\
\bottomrule
\end{tabular}
\end{center}

The policy is trained with a fixed coefficient and learns a joint feature-coefficient coupling where selected features are calibrated to the static intervention scale.
Changing the coefficient at inference time disrupts this coupling.
The static coefficient is itself robust, as varying $c$ from 10 to 100 on MMLU changes accuracy by only $\sim$3pp (\hyperref[tab:coefficient_sensitivity]{\textcolor{linkblue}{Table~\ref*{tab:coefficient_sensitivity}}}), confirming that the dataset-averaged approach finds a stable operating point.

\subsection{Component Ablation: Isolating CRL and AFM}
\label{subsec:component_ablation}

To isolate the contributions of learned feature selection and Adaptive Feature Masking, we compare CRL variants against heuristic baselines (\hyperref[tab:ablation_full]{\textcolor{linkblue}{Table~\ref*{tab:ablation_full}}}).
All methods share an identical decoding and evaluation setup, and only feature selection differs.

\begin{center}
\small
\captionof{table}{Component ablation. Numbers for CRL without AFM are from runs with matched hyperparameters. HarmBench CRL+AFM: mean over 3 seeds (paper reports 49.12, within run-to-run variance, $\pm$1.59).}
\label{tab:ablation_full}
\begin{tabular}{lcc}
\toprule
\textbf{Method} & \textbf{BBQ Ambig} & \textbf{HarmBench} \\
\midrule
No steering (Base) & 60.17 & 41.46 \\
Random feature & 58.36 & 45.35 \\
Random + AFM & 60.16 & 46.96 \\
Most-active feature & 59.94 & 48.03 \\
CRL without AFM & 62.55 & 48.67$_{\pm 0.56}$ \\
CRL + AFM (full) & \textbf{65.86} & \textbf{49.12}$_{\pm 1.59}$ \\
\bottomrule
\end{tabular}
\end{center}

CRL without AFM achieves +2.39 on BBQ Ambig and +7.21 on HarmBench over the baseline, confirming that the learned policy provides most of the gain.
AFM contributes an additional +3.31 on BBQ Ambig and +0.45 on HarmBench by encouraging exploration of diverse features.
Random+AFM improves over random alone (+1.80 BBQ, +1.61 HarmBench), isolating the independent contribution of AFM from learned selection.

\subsection{Testing the Prescribed Remedy}
\label{subsec:remedy_test}

\hyperref[subsec:sae_gaps]{\textcolor{linkblue}{Section~\ref*{subsec:sae_gaps}}} prescribes which remedy goes with which bottleneck: policy bottlenecks (HarmBench) call for selection diversity rather than critic redesign.
We apply that prescribed intervention to the matched task and two mismatched controls, training each from scratch over 3 seeds.
The intervention broadens feature selection: feature masking is relaxed from the single-feature default to all positions, top-$k$ is widened, and exploration is raised.
Deltas are measured against default-configuration CRL runs matched on seed.

\begin{center}
\small
\captionof{table}{Policy-broadening intervention on the matched policy-bottleneck task and two mismatched critic-bottleneck controls. $\Delta$ is accuracy in percentage points against CRL with the default configuration, matched on seed.}
\label{tab:remedy_test}
\begin{tabular}{lccc}
\toprule
\textbf{Task (diagnosis)} & \textbf{Tokens} & \textbf{Mean $\Delta$} & \textbf{Per seed} \\
\midrule
HarmBench (policy) & multi & +4.2 & +4.3, +5.4, +2.9 \\
MMLU (critic) & single & $-$3.2 & $-$4.5, $-$4.2, $-$0.8 \\
XSTest (critic) & multi & $-$0.7 & $-$2.5, +1.0, $-$0.6 \\
\bottomrule
\end{tabular}
\end{center}

Among the three tasks, the intervention yields a positive mean effect only on the task diagnosed as policy-bottlenecked, with consistent sign across all three seeds. XSTest is multi-token like HarmBench but diagnosed as a critic bottleneck (\hyperref[subsec:critic_behavior_analysis]{\textcolor{linkblue}{Section~\ref*{subsec:critic_behavior_analysis}}}), so it is a partial control against the simplest alternative based only on token count.
Under that narrow account both multi-token tasks should benefit, but XSTest shows no consistent benefit across seeds.
This weighs against a token-count-only explanation and strengthens, though it does not isolate, the diagnosis-based account.
The experiment does not establish that increasing critic capacity repairs the critic-bottleneck tasks, which remains open.

\end{document}